%% file: aistats.tex
\pdfoutput=1
\documentclass[twoside]{article}
 
\usepackage[accepted]{aistats2019}
\usepackage[round]{natbib}

\usepackage{amssymb,amsmath}
\usepackage{empheq}
\usepackage{mathtools}
\mathtoolsset{showonlyrefs}
\usepackage{mathbbol}
\usepackage{bm}
\usepackage{nicefrac}
\usepackage{booktabs}

\usepackage{enumitem}

\usepackage{parskip}

\usepackage{xcolor}
\usepackage{xspace}
\usepackage{floatrow}
\newfloatcommand{capbtabbox}{table}[][\FBwidth]
\usepackage{algorithm}
\usepackage{algorithmic}
\usepackage{pgf}

\usepackage{mdframed}
\definecolor{theoremcolor}{rgb}{0.94, 0.97, 1.0}  % lightblue
\mdfsetup{
    backgroundcolor=theoremcolor,
    linewidth=0pt,
}

\newmdtheoremenv{definition}{Definition} 
\newmdtheoremenv{proposition}{Proposition}
\newmdtheoremenv{corollary}{Corollary} 
\newmdtheoremenv{theorem}{Theorem} 
\newmdtheoremenv{lemma}{Lemma} 

\usepackage{marvosym}

\usepackage{hyperref}
\ifdefined\nohyperref\else\ifdefined\hypersetup
  \definecolor{mydarkblue}{rgb}{0,0.08,0.45}
  \hypersetup{ %
    pdftitle={},
    pdfauthor={},
    pdfsubject={},
    pdfkeywords={},
    pdfborder=0 0 0,
    pdfpagemode=UseNone,
    colorlinks=true,
    linkcolor=mydarkblue,
    citecolor=mydarkblue,
    filecolor=mydarkblue,
    urlcolor=mydarkblue,
    pdfview=FitH}

  \ifdefined\isaccepted \else
    \hypersetup{pdfauthor={Anonymous Submission}}
  \fi
\fi\fi

\DeclareMathOperator*{\argmin}{\mathsf{argmin}}
\DeclareMathOperator*{\argmax}{\mathsf{argmax}}

\DeclareMathOperator*{\dom}{\mathsf{dom}}
\DeclareMathOperator*{\interior}{\mathsf{int}}
\DeclareMathOperator*{\relint}{\mathsf{relint}}

\DeclareMathOperator*{\sigmoid}{\mathsf{sigmoid}}
\DeclareMathOperator*{\softmax}{\mathsf{softmax}}
\DeclareMathOperator*{\sparsemax}{\mathsf{sparsemax}}

\def\RR{{\mathbb{R}}}

\def\cC{{\mathcal{C}}}
\def\cD{{\mathcal{D}}}
\def\cX{{\mathcal{X}}}
\def\cY{{\mathcal{Y}}}

\def\ones{\bm{1}}
\def\zeros{\bm{0}}
\def\e{\bm{e}}

\def\p{{\bm{p}}}

\def\x{{\bm{x}}}
\def\y{{\bm{y}}}

\def\bpsi{\bm{\psi}}

\def\s{\bm{\theta}}

\def\pp{p}

\def\ss{\theta}

\def\yHat{\widehat{\y}}
\def\yHatOmega{\widehat{\y}_{\Omega}}

\newcommand{\DP}[2]{{\langle #1, #2\rangle}}
\newcommand\logls{logistic\xspace}
\newcommand\wrt{w.r.t.\ }
\newcommand\fy{\mbox{F-Y}\xspace}

\def\thresh{\tau}

\def\KL{\mathsf{KL}}
\def\HH{\mathsf{H}}
\def\HHs{\HH^{\textsc{s}}}
\def\HHt{\HH^{\textsc{t}}}
\def\HHr{\HH^{\textsc{r}}}
\def\HHn{\HH^{\textsc{n}}}
\def\HHsq{\HH^{\textsc{sq}}}

\begin{document}

% For long titles
\runningtitle{Learning Classifiers with Fenchel-Young Losses: Generalized
Entropies, Margins, and Algorithms}

% For long list of authors

\twocolumn[

\aistatstitle{Learning Classifiers with Fenchel-Young Losses:\\Generalized Entropies,
Margins, and Algorithms}

\aistatsauthor{ Mathieu Blondel \And Andr\'e F.T. Martins \And  Vlad Niculae }
\aistatsaddress{ NTT CS laboratories \\ Kyoto, Japan \And  
       Unbabel, Instituto de Telecomunica\c{c}\~oes \\ Lisbon, Portugal \And
       Instituto de Telecomunica\c{c}\~oes \\ Lisbon, Portugal}

%\aistatsauthor{ Author Name 1 \And Author Name 2 \And  Author Name 3 }
%\aistatsaddress{ Institution 1 \And  Institution 2 \And Instituton 3}

]

\begin{abstract}
This paper studies Fenchel-Young losses, a generic way to \textit{construct}
convex loss functions from a regularization function.  We analyze their
properties in depth, showing that they unify many well-known loss functions and
allow to create useful new ones easily.
Fenchel-Young losses constructed from a generalized
entropy, including the Shannon and Tsallis entropies, induce predictive probability
distributions.  We formulate conditions for a
generalized entropy to yield losses with a separation margin, and probability
distributions with sparse support.
Finally, we derive efficient algorithms, making
Fenchel-Young losses appealing both in theory and practice.
\end{abstract}

\vspace{-1.05em}
\section{Introduction}

Loss functions are a cornerstone of statistics and machine learning: They
measure the difference, or ``loss,'' between a ground-truth label and a
prediction. Some loss functions, such as the hinge loss of support vector
machines, are intimately connected to the notion of separation margin---a
prevalent concept in statistical learning theory, which has been used to prove
the famous perceptron mistake bound \citep{perceptron} and many other
generalization bounds \citep{Vapnik1998,Schoelkopf2002}.  For probabilistic
classification, the most popular loss is arguably the (multinomial) logistic
loss. It is smooth, enabling fast convergence rates, and the
softmax operator provides a consistent mapping to probability distributions.
However, the logistic loss does not enjoy a margin, and the generated
probability distributions have \textbf{dense} support, which is undesirable in
some applications for interpretability or computational efficiency reasons.

To address these shortcomings, \citet{sparsemax} proposed a new loss based on the
projection onto the simplex.  Unlike the logistic loss, this ``sparsemax'' loss has
a natural separation margin and induces a \textbf{sparse} probability
distribution.  However, the sparsemax loss was derived in a relatively ad-hoc
manner and it is still relatively poorly understood.
Thorough understanding of the
core principles underpinning these losses, enabling the creation of new losses
combining their strengths, is still lacking. 

This paper studies and extends Fenchel-Young (\fy) losses, recently proposed for
structured prediction \citep{sparsemap}. We show that \fy losses
provide a generic and principled way to \textbf{construct} a loss with
an associated probability distribution. 
We uncover a fundamental connection between
generalized entropies, margins, and sparse probability distributions. 
%We further derive sufficient conditions ensuring twice differentiability
%everywhere of a loss.
In sum, we make the following contributions.
\begin{itemize}[topsep=0pt,itemsep=2pt,parsep=2pt,leftmargin=10pt]

\item We introduce regularized prediction functions to generalize the softmax
    and sparsemax transformations, possibly beyond the probability simplex
    (\S\ref{sec:reg_pred}).

\item We study \fy losses and their properties,
    showing that they unify many existing losses, including the
    %hinge, multinomial \logls, one-vs-all \logls and sparsemax losses
    %(\S\ref{sec:fy_losses}).
    hinge, \logls, and sparsemax losses
    (\S\ref{sec:fy_losses}).

\item We then show how to seamlessly create entire new families of losses
    from generalized entropies.
We derive efficient algorithms to compute the associated probability
distributions,
making such losses appealing both in theory and in practice 
    (\S\ref{sec:proba_clf}).  

\item We characterize which entropies yield
    sparse distributions and losses with a separation margin,
    notions we prove to be intimately connected
    (\S\ref{sec:margin}).

\item Finally, we demonstrate \fy losses on the task of sparse label
    proportion estimation (\S\ref{sec:experiments}).
\end{itemize}

\textbf{Notation.}
We denote the probability simplex by $\triangle^d \coloneqq
\{\p \in \RR_+^d \colon \|\p\|_1 = 1\}$,
the domain of  
$\Omega \colon \RR^d \rightarrow \RR\cup\{\infty\}$ 
by $\dom(\Omega) \coloneqq \{\p \in \RR^d \colon
\Omega(\p) < \infty\}$,
the Fenchel conjugate of $\Omega$ by 
$\Omega^*(\s) \coloneqq \displaystyle{\sup_{\p \in \dom(\Omega)}} \DP{\s}{\p} -
\Omega(\p)$,
the indicator function of a set $\cC$ by $I_\cC$.
%the
%interior and relative interior of $\cC$ by $\interior(\cC)$ and $\relint(\cC)$.

\section{Regularized prediction functions}
\label{sec:reg_pred}

%In this section, we introduce the concept of regularized prediction function,
%which will be central to this paper. 
We consider a general predictive setting with input $\x \in \cX$, and
a parametrized model $\bm{f}_{W}:\cX \rightarrow \RR^d$, producing a score
vector $\s \coloneqq \bm{f}_{W}(\x)$. 
To map $\s$ to predictions, we introduce regularized prediction functions.
\vspace{0.5em}
\begin{definition}{Regularized prediction function}\label{def:regularized_pred}

Let $\Omega \colon \RR^d \to \RR \cup \{\infty\}$ be a regularization function,
with $\dom(\Omega) \subseteq \RR^d$.  The prediction function regularized by
$\Omega$ is defined by
\begin{equation}
\widehat{\y}_{\Omega}(\s) 
\in \argmax_{\p \in
\dom(\Omega)} {\DP{\s}{\p}} - \Omega(\p).
\label{eq:prediction}
\end{equation}
\end{definition}
We emphasize that the regularization is \wrt the output and not \wrt the
model parameters $W$, as is usually the case in the
literature.
The optimization problem in \eqref{eq:prediction}
balances between two terms: an ``affinity'' term
$\DP{\s}{\p}$, and a ``confidence'' term $\Omega(\p)$
which should be low if $\p$ is ``uncertain''.
Two important classes
of convex $\Omega$ are (squared) norms 
and, when $\dom(\Omega)$ is the probability simplex,
generalized negative entropies.
However, our framework does not require $\Omega$ to be convex in
general.
Allowing extended-real $\Omega$ further permits
general
domain constraints in \eqref{eq:prediction} via indicator functions, as we now
illustrate.

\paragraph{Examples.}

%We now illustrate several possible choices of $\Omega$, leading to
%a \textbf{closed-form solution}.
When $\Omega = I_{\triangle^d}$,
$\yHatOmega(\s)$ is a one-hot representation of the argmax
prediction
\begin{equation}
\yHatOmega(\s) \in
\argmax_{\p \in \triangle^d} ~ \langle \s, \p \rangle =
\argmax_{\y \in \{\e_1,\dots,\e_d\}} \langle \s, \y \rangle.
\end{equation}
We can see that output as a probability distribution that assigns all probability
mass on the same class.
When $\Omega = -\HHs + I_{\triangle^d}$, where
$\HHs(\p) \coloneqq -\sum_i p_i \log p_i$ is Shannon's entropy,
$\yHatOmega(\s)$ is the well-known softmax
\begin{equation}
\yHatOmega(\s)
= \softmax(\s) \coloneqq
\frac{\exp(\s)}{\sum_{j=1}^d \exp(\ss_j)}.
\label{eq:softmax}
\end{equation}
See \citet[Ex.\ 3.25]{boyd_book} for a derivation. The resulting distribution
always has \textbf{dense} support.
When $\Omega=\frac{1}{2} \|\cdot\|^2 + I_{\triangle^d}$, 
$\yHatOmega$ is the
Euclidean projection onto the probability simplex
\begin{equation}
\yHatOmega(\s) = \sparsemax(\s) \coloneqq
\argmin_{\p \in \triangle^d} \|\p - \s\|^2,
\label{eq:sparsemax}
\end{equation}
a.k.a.\ the sparsemax transformation \citep{sparsemax}.
The distribution has \textbf{sparse} support (it may assign exactly
zero probability to low-scoring classes) and can
be computed \textbf{exactly} in $O(d)$ time
\citep{Brucker1984,duchi,Condat2016}.
This paradigm is not limited to the probability simplex:
When $\Omega(\p) = -\sum_i \HHs([p_i,1-p_i]) + I_{[0,1]^d}(\p)$,
we get
\begin{equation}
\yHatOmega(\s) = \sigmoid(\s)
\coloneqq \frac{\ones}{\ones + \exp(-\s)},
\label{eq:sigmoid}
\end{equation}
i.e., the sigmoid function evaluated coordinate-wise. We can think of its
output as a positive measure (unnormalized probability distribution).

\paragraph{Properties.}

We now discuss simple properties of regularized prediction functions.
The first two assume that $\Omega$ is a symmetric function, i.e.,
that it satisfies
\begin{equation}
\Omega(\p)=\Omega(\bm{P}\p)
\quad \forall \p \in \dom(\Omega), \forall \bm{P} \in \mathcal{P},
\end{equation}
where $\mathcal{P}$ is the set of $d \times d$ permutation matrices.

\vspace{0.5em}
\begin{proposition}{Properties of $\yHatOmega(\s)$}

\begin{enumerate}[topsep=0pt,itemsep=3pt,parsep=3pt,leftmargin=15pt]

\item {\bf Effect of a permutation.} If $\Omega$ is symmetric, then
$\forall \bm{P} \in \mathcal{P}$:
    $\yHatOmega(\bm{P} \s) = \bm{P} \yHatOmega(\s)$.

\item {\bf Order preservation.}
 Let $\p = \yHatOmega(\s)$. If $\Omega$ is symmetric, then the coordinates of
    $\p$ and $\s$ are sorted the same way, i.e., $\ss_i > \ss_j \Rightarrow
    p_i \ge p_j$ and $p_i > p_j \Rightarrow \ss_i > \ss_j$. 

\item {\bf Gradient mapping.}
$\yHatOmega(\s)$ is a subgradient of $\Omega^*$ at $\s$, i.e.,
$\yHatOmega(\s) \in \partial \Omega^*(\s)$.
If $\Omega$ is strictly convex,
$\yHatOmega(\s)$ is the gradient of $\Omega^*$, i.e.,
$\yHatOmega(\s) = \nabla \Omega^*(\s)$.

\item {\bf Temperature scaling.} For any constant $t > 0$,
$\widehat{\y}_{t \Omega}(\s) \in \partial \Omega^*(\nicefrac{\s}{t})$.
If $\Omega$ is strictly convex,
$\widehat{\y}_{t \Omega}(\s) = \yHatOmega(\nicefrac{\s}{t}) = \nabla
\Omega^*(\nicefrac{\s}{t})$.

\end{enumerate}

\label{prop:prediction_func}
\end{proposition}
The proof is given in \S\ref{appendix:proof_prediction_func}.
For classification, the order-preservation property ensures that the
highest-scoring class according to $\s$ and $\yHatOmega(\s)$ agree 
with each other:
\begin{equation}
\argmax_{i \in [d]} \ss_i = \argmax_{i \in [d]} \left(\yHatOmega(\s)\right)_i.
\end{equation}
Temperature scaling is useful to control how close we are to unregularized
prediction functions. 

\section{Fenchel-Young losses}
\label{sec:fy_losses}

\begin{table*}
    \caption{{\bf Examples of regularized prediction functions and their
            associated
        Fenchel-Young losses.} For multi-class classification, we denote the 
ground-truth by $\y=\e_k$, where $\e_i$
    denotes a standard basis (``one-hot'') vector.  
We denote by $\HHs(\p) \coloneqq -\sum_i \pp_i\log
\pp_i$ the Shannon entropy of a distribution $\p \in \triangle^d$.}
\begin{center}
\begin{small}
\begin{tabular}{@{\hskip 0pt}l@{\hskip 0pt}c@{\hskip 0pt}c@{\hskip 5pt}c@{\hskip 10pt}c@{\hskip 0pt}}
\toprule
Loss & $\dom(\Omega)$ & $\Omega(\p)$ & $\widehat{\y}_{\Omega}(\s)$ & $L_{\Omega}(\s; \y)$ \\
\midrule
Squared %& $\RR^d$ 
& $\RR^d$ & $\frac{1}{2}\|\p\|^2$ & $\s$ & $\frac{1}{2}\|\y-\s\|^2$ \smallskip
\\[0.5em]
Perceptron \citep{perceptron} %& $\{\e_i\}_{i=1}^d$ 
& $\triangle^d$ & $0$ & $\argmax(\s)$ 
& $\max_i ~ \ss_i -\ss_k$ 
\smallskip
\\
Hinge \citep{multiclass_svm} %& $\{\e_i\}_{i=1}^d$ 
& $\triangle^d$ & $\DP{\p}{\e_k - \ones}$ & 
$\argmax(\ones\!-\!\e_k\!+\!\s)$
& $\max_i ~ [[i \neq k]] + \ss_i -\ss_k$ 
\smallskip
\\
Sparsemax \citep{sparsemax} %& $\{\e_i\}_{i=1}^d$ 
& $\triangle^d$ & $\frac{1}{2}\|\p\|^2$ & $\sparsemax(\s)$ & 
%$\frac{1}{2}\|\y-\s\|^2 - \frac{1}{2}\|\widehat{\y}_{\Omega}(\s) - \s\|^2$ 
$\frac{1}{2}\|\y-\s\|^2- \frac{1}{2}\|\widehat{\y}_{\Omega}(\s) - \s\|^2$ 
\smallskip
\\
Logistic (multinomial) %& $\{\e_i\}_{i=1}^d$ 
& $\triangle^d$ & $-\HHs(\p)$ & $\softmax(\s)$ & 
%$\KL(\y\|\widehat{\y}_{\Omega}(\s))$ 
$\log\sum_i\exp \ss_i -\ss_k$ 
\smallskip
\\[0.5em]
%% VN: I see what you mean with these slight spaces, but I fear they are
%%     too slow and look more like formatting mistakes in the paper currently.
%%     In the journal version we can add small headers clarifying the groups of
%%     losses, but here it may be better to use uniform row spacing..
Logistic (one-vs-all) %& $\{\e_i\}_{i=1}^d$ 
& $[0,1]^d$ 
&
$-\hspace{-.6ex}\sum_i\!\HHs\hspace{-.2ex}\bigl(\hspace{-.3ex}[p_i,\!1\!-\!p_i]\hspace{-.3ex}\bigr)$ & $\sigmoid(\s)$ &
$\sum_i\log\bigl(1\hspace{-.4ex}+\hspace{-.4ex}\exp(-(2
y_i\hspace{-.4ex}-\hspace{-.4ex}1) \ss_i)\bigr)$
\smallskip
\\
\bottomrule
\end{tabular}
\end{small}
\end{center}
\label{tab:fy_losses_examples}
\end{table*}

%In the previous section, we have introduced regularized prediction functions
%over arbitrary domains, as
%a generalization of classical (unregularized) decision functions.
In this section, we introduce Fenchel-Young losses as a natural way to learn
models whose output layer is a regularized prediction function.
%
%Given a function $\Omega$, we define the 
%Given a regularized prediction function $\yHatOmega$, we define its
%associated Fenchel-Young loss as follows.
\vspace{0.5em}
\begin{definition}{Fenchel-Young loss generated by $\Omega$}\label{def:FY_loss}

Let $\Omega \colon \RR^d \to \RR \cup \{\infty\}$ be a regularization
function such that the maximum in
\eqref{eq:prediction} is achieved for all $\s \in \RR^d$.
Let $\y \in \cY
\subseteq \dom(\Omega)$ be a ground-truth label and $\s \in \dom(\Omega^*)
= \RR^d$ be a vector of prediction scores. 
The {\bf Fenchel-Young loss} $L_\Omega \colon \dom(\Omega^*)
\times \dom(\Omega) \to \RR_+$ generated by $\Omega$ is
\begin{equation}
L_{\Omega}(\s; \y) 
\coloneqq \Omega^*(\s) + \Omega(\y) - \DP{\s}{\y}.
\label{eq:fy_losses}
\end{equation}
\end{definition}
Fenchel-Young losses can also be written as
$L_{\Omega}(\s; \y) = f_{\s}(\y) - f_{\s}(\yHatOmega(\s))$,
where $f_{\s}(\p) \coloneqq \Omega(\p) - \DP{\s}{\p}$, 
highlighting the relation with the regularized prediction function
$\yHatOmega(\s)$.
Therefore, as long as we can compute 
$\yHatOmega(\s)$, we can evaluate the associated Fenchel-Young loss
$L_\Omega(\s; \y)$.
Examples of Fenchel-Young losses are given
in Table~\ref{tab:fy_losses_examples}. 
In addition to the aforementioned multinomial logistic and sparsemax losses, we
recover the squared, hinge and one-vs-all logistic losses, for suitable choices
of $\Omega$ and $\dom(\Omega)$.

\vspace{-0.5em}
\paragraph{Properties.}

As the name indicates, this family of loss functions is grounded in the
Fenchel-Young inequality
\citep[Proposition 3.3.4]{borwein_2010}
\begin{equation}
\Omega^*(\s) + \Omega(\p) \ge \DP{\s}{\p} 
~\forall \s \in \dom(\Omega^*), \p \in \dom(\Omega).
\label{eq:fenchel_young_inequality}
\end{equation}
The inequality, together with well-known properties of
convex conjugates,
imply the following results.
\vspace{0.5em}
\begin{proposition}{Properties of \fy losses}
\label{prop:fy_losses}

\begin{enumerate}[topsep=0pt,itemsep=3pt,parsep=3pt,leftmargin=15pt]

\item {\bf Non-negativity.} $L_{\Omega}(\s; \y) \ge 0$ for any $\s \in
    \dom(\Omega^*)=\RR^d$
and $\y \in \cY \subseteq \dom(\Omega)$. 

%\item {\bf Zero loss.} If $\Omega$ is a lower semi-continuous proper convex
    %function, then the loss is zero iff  $\y \in \partial \Omega^*(\s)$.
    %If $\Omega$ is strictly convex, the loss is zero iff $\y = \yHatOmega(\s) =
    %\nabla \Omega^*(\s)$.

\item {\bf Zero loss.} If $\Omega$ is a lower semi-continuous proper convex
    function, then
    $\min_{\s} L_\Omega(\s; \y) = 0$ and $L_\Omega(\s; \y) = 0$ iff $\y
    \in \partial \Omega^*(\s)$.  
If $\Omega$ is strictly convex, then $L_\Omega(\s; \y) = 0$ iff  $\y
= \yHatOmega(\s) = \nabla \Omega^*(\s)$.

\item {\bf Convexity \& subgradients.} $L_{\Omega}$ is convex in $\s$ and the
    residual vectors are its subgradients: $\widehat{\y}_{\Omega}(\s) - \y \in
    \partial L_{\Omega}(\s; \y)$. 
    
\item {\bf Differentiability \& smoothness.}
    If $\Omega$ is strictly convex, then $L_{\Omega}$
    is differentiable and $\nabla L_{\Omega}(\s; \y) = \widehat{\y}_{\Omega}(\s) - \y$.
    If $\Omega$ is strongly convex, then $L_\Omega$ is smooth, i.e., $\nabla
    L_\Omega(\s; \y)$ is Lipschitz continuous.

\item {\bf Temperature scaling.} For any constant $t > 0$, 
    $L_{t \Omega}(\s; \y) = t L_{\Omega}(\nicefrac{\s}{t}; \y)$.

\end{enumerate}

\end{proposition}
Remarkably, the non-negativity and convexity properties
hold even if $\Omega$ is not convex.  The zero loss property 
follows from the fact that, if $\Omega$ is l.s.c.\ proper convex, then
\eqref{eq:fenchel_young_inequality} becomes an equality (i.e., the duality gap
is zero) if and only if $\s \in \partial \Omega(\p)$.  
It suggests that
minimizing a Fenchel-Young loss requires adjusting $\s$ to produce
predictions $\widehat{\y}_\Omega(\s)$ that are close to the target $\y$,
reducing the duality gap. 

\paragraph{Relation with Bregman divergences.}

Fenchel-Young losses seamlessly work when ground-truth vectors are label
proportions, i.e., $\y \in \triangle^d$ instead of $\y \in \{\e_i\}_{i=1}^d$.
For instance, setting
$\Omega$ to the Shannon negative entropy restricted to $\triangle^d$
yields the cross-entropy loss, 
$L_{\Omega}(\s; \y) = \KL(\y \| \softmax(\s))$, where $\KL$ denotes the
Kullback-Leibler divergence.
From this example, it is tempting to conjecture that a similar result holds for
more general Bregman divergences \citep{bregman_1967}. 
Recall that the Bregman divergence $B_\Omega \colon
\dom(\Omega) \times \relint(\dom(\Omega)) \to \RR_+$ generated by a strictly
convex and differentiable $\Omega$ is
\begin{equation}
B_\Omega(\y||\p) 
\coloneqq \Omega(\y) - \Omega(\p) - 
\DP{\nabla \Omega(\p)}{\y - \p},
\label{eq:Bregman_div}
\end{equation}
the difference at $\y$ between $\Omega$ and its
linearization around $\p$. It turns out that $L_\Omega(\s; \y)$ is not in
general equal to $B_\Omega(\y || \yHatOmega(\s))$. 
However, when $\Omega = \Psi + I_\cC$, where $\Psi$ is a
Legendre-type function \citep{Rockafellar1970,wainwright_2008}, meaning
that it is strictly convex,
differentiable and its gradient explodes at the boundary of its domain,
we have the following proposition,
proved in \S\ref{appendix:proof_Bregman_div}.
\vspace{0.5em}
\begin{proposition}\label{prop:Bregman_div}
Let $\Omega \coloneqq \Psi + I_\cC$, where $\Psi$ is of Legendre type
and $\cC \subseteq \dom(\Psi)$ is a
convex set.  Then, for all $\s \in \RR^d$ and $\y \in \cC$, we have:
\begin{equation}
    0 \le \underbrace{B_\Omega(\y || \yHatOmega(\s))}_{\text{possibly
    non-convex in } \s} \le \underbrace{L_{\Omega}(\s; \y)}_{\text{convex in } \s}
\label{eq:fy_bound}
\end{equation}
with equality when the loss is $0$.
If $\cC = \dom(\Psi)$, i.e., $\Omega = \Psi$, then
$L_\Omega(\s; \y) = B_\Omega(\y || \yHatOmega(\s))$.
\end{proposition}
As an example, applying \eqref{eq:fy_bound} with $\Psi = \frac{1}{2}
\|\cdot\|^2$ and $\cC = \triangle^d$, we get that the sparsemax loss is a
convex upper-bound for the non-convex $\frac{1}{2} \|\y -
\sparsemax(\cdot)\|^2$. This suggests that the sparsemax loss can be useful for
sparse label proportion estimation, as we confirm in \S\ref{sec:experiments}.

The relation between Fenchel-Young losses and Bregman divergences can be further
clarified using duality.
Letting $\s = \nabla \Omega(\p)$ (i.e., $(\s, \p)$ is a \textbf{dual pair}), we have 
$\Omega^*(\s) = \DP{\s}{\p} - \Omega(\p)$. Substituting in 
\eqref{eq:Bregman_div}, we get 
$B_\Omega(\y||\p) = L_{\Omega}(\s; \y).$ 
In other words, 
Fenchel-Young losses can be
viewed as a ``mixed-form Bregman divergence'' 
\cite[Theorem 1.1]{amari_2016}
where the
argument $\p$ in \eqref{eq:Bregman_div} is \textbf{replaced by its dual point} $\s$.
This difference is best seen by comparing
the function signatures, $L_\Omega \colon \dom(\Omega^*) \times \dom(\Omega) \to
\RR_+$ vs. $B_\Omega \colon \dom(\Omega) \times \relint(\dom(\Omega)) \to
\RR_+$. An important consequence is that 
Fenchel-Young losses do not impose any restriction on their left argument $\s$:
Our assumption that the maximum in the prediction
function \eqref{eq:prediction} is achieved for all $\s \in \RR^d$ implies
$\dom(\Omega^*) = \RR^d$. 
%This contrasts with proper loss functions, which are defined over the
%probability simplex,
%as discussed in \S\ref{sec:related_work}.

\begin{figure*}
\centering\input{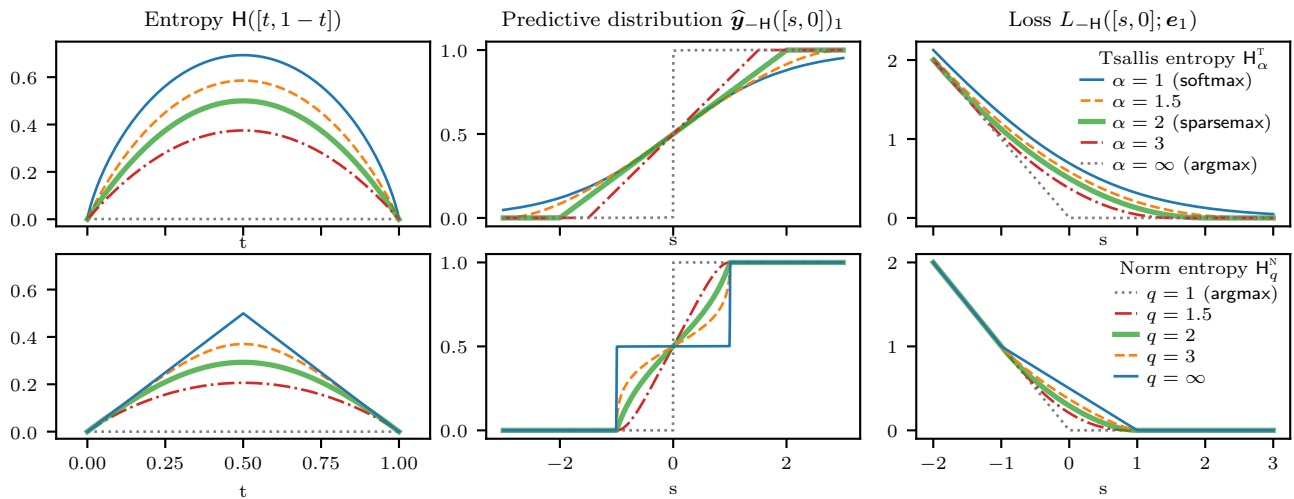}
\caption{{\bf New families of losses made possible by our framework.}
    Left: Tsallis and norm entropies. Center: regularized prediction functions.
    Right: Fenchel-Young loss.  Except for softmax, which never exactly reaches
    0, all distributions shown in the center can have
    \textbf{sparse support}. As can be checked visually,
    $\widehat{\y}_{-\HH}$ is differentiable everywhere when $\alpha, q \in (1,2)$.
    Hence, $L_{-\HH}$ is \textbf{twice differentiable} everywhere for these values.
}
\label{fig:entropy_max_argmax}
\end{figure*}

\section{New loss functions for sparse probabilistic classification}
\label{sec:proba_clf}

In the previous section, we presented Fenchel-Young losses in a broad setting.
We now restrict to classification over the probability simplex and show how
to easily create several entire new families of losses.

\paragraph{Generalized entropies.}
A natural choice of regularization function $\Omega$ over the probability simplex is
$\Omega=-\HH$, where $\HH$ is a generalized entropy
\citep{degroot_1962,grunwald_2004}: a concave function over
$\triangle^d$, used
to measure the ``uncertainty'' in a distribution $\p \in
\triangle^d$.

\textbf{Assumptions:}
We will make the following assumptions about $\HH$.

\begin{enumerate}[topsep=0pt,itemsep=3pt,parsep=3pt,leftmargin=25pt]
    \item[\textbf{A.1.}] 
    %Non-negativity: $H(\p) \ge 0$ for all $\p \in \triangleY$.
    Zero entropy: $\HH(\p) = 0$ if $\p$ is a delta distribution,
    i.e., $\p \in \{\e_i\}_{i=1}^d$.
    \item[\textbf{A.2.}] 
        Strict concavity: $\HH\big((1-\alpha)\p + \alpha \p'\big) > {(1-\alpha)} \HH(\p) +
        \alpha \HH(\p')$, for $\p \neq \p'$, $\alpha \in (0,1)$.
    \item[\textbf{A.3.}] 
        Symmetry: $\HH(\p) =
        \HH(\bm{P}\p)$ for any $\bm{P} \in \mathcal{P}$.

\end{enumerate}

Assumptions A.2 and A.3 imply that $\HH$ is Schur-concave \citep{Bauschke2017},
a common requirement in generalized entropies. This in turn implies assumption
A.1, up to a constant (that constant can easily be subtracted so as to satisfy
assumption A.1).
As suggested by the next result, 
proved in \S\ref{appendix:proof_generalized_entropy},
together, these assumptions
imply that $\HH$ can be used as a sensible
uncertainty measure.
\vspace{0.4em}
%\begin{proposition}{Non-negativity of $\HH$ and maximum entropy}
\begin{proposition}
If $\HH$ satisfies assumptions A.1-A.3, then it is non-negative and uniquely
maximized by the uniform distribution $\p = \mathbf{1}/d$.
\label{prop:generalized_entropy}
\end{proposition}
A particular case of generalized entropies satisfying assumptions A.1--A.3 are
uniformly separable functions of the form $\HH(\p) = \sum_{j=1}^d h(p_j)$,
where $h:[0,1]\rightarrow\RR_+$ is a non-negative strictly concave function such
that $h(0)=h(1)=0$. However, our framework is not restricted to this form.

\paragraph{Induced Fenchel-Young loss.}

If the ground truth is $\y = \e_k$ and assumption
A.1.\ holds, \eqref{eq:fy_losses} becomes
\begin{equation}
L_{-\HH}(\s; \e_k) 
%= (-\HH)^*(\s) - \HH(\e_k) - \DP{\s}{\e_k}
= (-\HH)^*(\s) - \ss_k.
\label{eq:fy_loss_multiclass2}
\end{equation}
By using the fact that $\Omega^*(\s + c \ones) = \Omega^*(\s) + c$ for all $c
\in \RR$ if $\dom(\Omega) \subseteq \triangle^d$,
we can further rewrite it as
\begin{equation}
L_{-\HH}(\s; \e_k) 
= (-\HH)^*(\s - \ss_k \ones).
\label{eq:fy_loss_multiclass}
\end{equation}
This expression shows that Fenchel-Young losses over $\triangle^d$ can be
written solely in terms of the generalized ``cumulant function'' $(-\HH)^*$. 
Indeed, when $\HH$ is Shannon's entropy, we recover the cumulant (a.k.a.
log-partition) function
$(-\HHs)^*(\s) = \log \sum_{i=1}^d \exp(\theta_i)$.
When $\HH$ is strongly concave over $\triangle^d$,
we can also see $(-\HH)^*$ as a smoothed max operator
\citep{Niculae2017,differentiable_dp} and hence $L_{-\HH}(\s;
\e_k)$ can be seen as a smoothed upper-bound of the perceptron loss
$(\s; \e_k) \mapsto \max_{i \in [d]} \theta_i - \theta_k$.

We now give two examples of generalized entropies.
The resulting families of prediction and loss functions, new to our knowledge,
are illustrated in Figure \ref{fig:entropy_max_argmax}.
We provide more examples in \S\ref{appendix:more_entropies}.
%(we
%omit the indicator function $I_{\triangle^d}$ from the definitions since there
%is no ambiguity).  

\paragraph{Tsallis $\alpha$-entropies \citep{Tsallis1988}.} 
Defined as
%\begin{equation}
$\HHt_{\alpha}(\p) \coloneqq k(\alpha-1)^{-1} \big(1 - \|\cdot\|^\alpha_\alpha\big)$, 
%\end{equation}
% = k(\alpha-1)^{-1}(1 - \|\p\|_{\alpha}^{\alpha})$,
where $\alpha \ge 1$ and $k$ is an arbitrary positive constant, 
these entropies arise as a generalization of the Shannon-Khinchin axioms to
non-extensive systems \citep{Suyari2004} and have numerous scientific
applications \citep{GellMannTsallis2004,Martins2009JMLR}. 
For convenience, we set $k = \alpha^{-1}$ for the rest of this paper. 
Tsallis entropies satisfy assumptions A.1--A.3 and can also be written in
uniformly separable form:
\begin{equation}
\HHt_{\alpha}(\p) \coloneqq \sum_{j=1}^d h_{\alpha}(p_j)
\quad \text{with} \quad
h_{\alpha}(t) \coloneqq \frac{t-t^\alpha}{\alpha(\alpha - 1)}.
\label{eq:tsallis_separable}
\end{equation}
The limit case $\alpha \rightarrow 1$ corresponds to the Shannon entropy.
When $\alpha=2$, we recover the Gini index \citep{gini_index}, a 
popular ``impurity measure'' for decision trees:
\begin{equation}
\HHt_2(\p) = \frac{1}{2}\sum_{j=1}^d p_j (1-p_j) =
\frac{1}{2}(1-\|\p\|_2^2)
\quad \forall \p \in \triangle^d.
\label{eq:gini_entropy}
\end{equation}
It is easy to check that $L_{-\HHt_2}$ recovers
the sparsemax loss \citep{sparsemax} (cf.\ Table~\ref{tab:fy_losses_examples}). 
Another interesting case is $\alpha\rightarrow +\infty$, which gives
$\HHt_\infty(\p) = 0$, hence $L_{-\HHt_\infty}$ is the perceptron loss in
Table~\ref{tab:fy_losses_examples}. The resulting ``$\argmax$'' distribution puts
all probability mass on the top-scoring classes.
In summary, $\widehat{\y}_{-\HHt_\alpha}$ for $\alpha \in \{1, 2, \infty\}$ is
$\softmax$, $\sparsemax$, and $\argmax$, and $L_{-\HHt_\alpha}$ is the logistic,
sparsemax and perceptron loss, respectively. Tsallis entropies induce a
\textbf{continuous parametric family} subsuming these important cases. 
Since the best surrogate loss often depends on the data \citep{nock_2009},
tuning $\alpha$ typically improves accuracy, as we confirm in
\S\ref{sec:experiments}.

\paragraph{Norm entropies.}

An interesting class of non-separable entropies are entropies generated by a
$q$-norm, defined as
$\HHn_q(\p) \coloneqq 1 - \|\p\|_q$.
We call them norm entropies. 
From the Minkowski inequality, $q$-norms with $q>1$ are strictly convex on the
simplex, so $\HHn_q$ satisfies assumptions A.1--A.3 for $q > 1$.
The limit case $q
\rightarrow \infty$ is particularly interesting: in this case, we obtain
$\HHn_{\infty} = 1-\|\cdot\|_{\infty}$, recovering the Berger-Parker dominance
index  \citep{Berger1970}, widely used in ecology to measure species diversity.  
We surprisingly encounter $\HHn_\infty$ again in \S\ref{sec:margin}, 
as a limit case for the existence of separation margins.

\paragraph{Computing $\widehat{\y}_{-\HH}(\s)$.}

For non-separable entropies $\HH$, the regularized prediction function
$\widehat{\y}_{-\HH}(\s)$ does not generally enjoy a closed-form expression and
one must resort to projected gradient methods to compute it.  Fortunately, for
uniformly separable entropies, which we saw to be the case of Tsallis entropies,
we now show that $\widehat{\y}_{-\HH}(\s)$ can be computed in linear time.
\vspace{0.5em}
\begin{proposition}%
\label{prop:root_finding}%
Reduction to root finding

Let $\HH(\p) = \sum_i h(p_i) + I_{\triangle^d}(\p)$ where $h \colon [0, 1]
\rightarrow \RR_+$ is strictly concave and differentiable. Then,
\begin{equation}
\widehat{\y}_{-\HH}(\s) = 
\p(\thresh) \coloneqq (-h')^{-1}(\max\{\s - \thresh, -h'(0)\})
\end{equation}
where $\thresh$ is a root of
$\phi(t) \coloneqq \DP{\p(t)}{\ones} - 1$, 
in the tight search interval
$[\thresh_{\min},\thresh_{\max}]$, where 
$\thresh_{\min} \coloneqq \max(\s) +h'(1)$
and
$\thresh_{\max} \coloneqq \max(\s) +h'\left(\nicefrac{1}{d}\right)$.
\end{proposition}
An approximate $\thresh$ such that $|\phi(\thresh)| \le \epsilon$ can be found in
$O(\nicefrac{1}{\log \epsilon})$ time by, e.g., bisection.
The related problem of Bregman projection
onto the probability simplex was recently studied by \citet{bregmanproj} but our
derivation is different and more direct (cf.\
\S\ref{appendix:proof_root_finding}).

\section{Separation margin of \fy losses}
\label{sec:margin}

In this section, we are going to see that the simple assumptions A.1--A.3 about
a generalized entropy $\HH$ are enough to obtain results about the {\bf separation margin} associated with $L_{-\HH}$. 
The notion of margin is well-known in machine learning, lying at the heart of support vector machines and leading to generalization error bounds \citep{Vapnik1998,Schoelkopf2002,guermeur_2007}. 
We provide a definition and will see that many other Fenchel-Young losses also have a ``margin,'' for suitable conditions on $\HH$.  
Then, we take a step further, and connect the existence of a margin with the {\bf sparsity} of the regularized prediction function, 
providing necessary and sufficient conditions for Fenchel-Young losses to have a margin.  Finally, we show how this margin can be computed analytically.
\vspace{0.5em}
\begin{definition}{Separation margin}

Let $L(\s; \e_k)$ be a loss function over $\RR^d \times \{\e_i\}_{i=1}^d$.
%where $\e_k$ is the ground-truth label.
We say that $L$ has
the \emph{separation margin property} if there exists $m>0$ such that:
\begin{equation}
\ss_k \ge m + \max_{j \ne k} \ss_j \quad \Rightarrow \quad L(\s; \e_k) = 0.
\label{eq:margin_definition}
\end{equation}
The smallest possible $m$ that satisfies \eqref{eq:margin_definition} is called 
the \emph{margin} of $L$, denoted $\mathrm{margin}(L)$.
\label{def:margin}
\end{definition}

\paragraph{Examples.}  
The most famous example of a loss with a separation margin is the  
{\bf multi-class hinge loss}, 
$L(\s; \e_k) = \max\{0, \max_{j \ne k} 1 + \ss_j - \ss_k\}$, 
which we saw in Table \ref{tab:fy_losses_examples} to be 
a Fenchel-Young loss: it is immediate from the definition that its margin is $1$. 
Less trivially, \citet[Prop.~3.5]{sparsemax} showed that the {\bf sparsemax loss} also 
has the separation margin property. 
On the negative side, the logistic loss does not have a margin, as it is
strictly positive.
Characterizing which Fenchel-Young losses have a margin is an open question
which we address next.

\paragraph{Conditions for existence of margin.} 
To accomplish our goal, we need to characterize the gradient mappings $\partial(-\HH)$ and $\nabla(-\HH)^{*}$ associated with generalized entropies
(note that $\partial(-\HH)$
is never single-valued:  
if $\s$ is in $\partial (-\HH)(\bm{p})$, then so is $\s+c\mathbf{1} $, for any
constant $c \in \mathbb{R}$).
Of particular importance is the subdifferential set 
$\partial(-\HH)(\e_k)$. 
The next proposition, whose proof we defer to 
\S\ref{appendix:proof_margin}, uses this set to provide a necessary and sufficient condition for the existence of a separation margin, along with a formula for computing it. 
\vspace{0.5em}
\begin{proposition}\label{prop:margin}
Let $\HH$ satisfy A.1--A.3. Then:
\begin{enumerate}
\item The loss $L_{-\HH}$ has a separation margin iff there is  a $m>0$ such that $m\e_k \in \partial (-\HH)(\e_k)$. 
\item If the above holds, then the margin of $L_{-\HH}$ is given by the smallest
    such $m$ or, equivalently,
\begin{equation}\label{eq:margin}
\mathrm{margin}(L_{-\HH}) = \sup_{\p \in \triangle^d} \frac{\HH(\p)}{1 - \|\p\|_{\infty}}. 
\end{equation}
\end{enumerate}
\end{proposition}
Reassuringly, the first part confirms that
the logistic loss does not have a margin,
since $\partial(-\HHs)(\e_k)=\varnothing$.
% VN: trying to de-emphasize and make less awkward. Weird to remark on a
%     proposition by actually referring to a lemma that is only in supp.
%
A second interesting fact is that the denominator of \eqref{eq:margin} is the generalized
entropy $\HHn_{\infty}(\p)$ introduced in \S\ref{sec:proba_clf}: the {\bf $\infty$-norm entropy}. 
As Figure~\ref{fig:entropy_max_argmax} suggests, this entropy provides an upper bound for convex losses with unit margin. 
This provides some intuition to the formula \eqref{eq:margin}, which seeks a
distribution $\p$ maximizing the {\bf entropy ratio} between $\HH(\p)$ and
$\HHn_{\infty}(\p)$.

\paragraph{Equivalence between sparsity and margin.} 
The next result, proved in \S\ref{appendix:proof_full_simplex},
characterizes more precisely the image of $\nabla (-\HH)^*$. 
In doing so, it establishes a key result in this paper: 
{\bf a sufficient condition for the existence of a separation margin in $L_{-\HH}$ is the sparsity of the regularized prediction function $\widehat{\y}_{-\HH} \equiv \nabla (-\HH)^*$},   
i.e., its ability to reach the entire simplex, including the boundary points. If $\HH$ is uniformly separable, this is also a necessary condition.  
\vspace{0.5em}
\begin{proposition}{Equivalence between sparse probability distribution and loss
    enjoying a margin}
\label{prop:full_simplex}

Let $\HH$ satisfy A.1--A.3 and be 
uniformly separable, i.e., $\HH(\p) = \sum_{i=1}^d h(p_i)$. Then the following statements are all equivalent:
\begin{enumerate}
\item $\partial (-\HH)(\p) \ne \varnothing$ %is non-empty 
for any $\p \in \triangle^d$;
\item The mapping $\nabla (-\HH)^*$ covers the full simplex, i.e., $\nabla
    (-\HH)^*(\RR^d) = \triangle^d$;
\item $L_{-\HH}$ has the separation margin property.
\end{enumerate}
For a general $\HH$ (not necessarily separable) satisfying A.1--A.3,
we have (1) $\Leftrightarrow$ (2) $\Rightarrow$ (3). 
\end{proposition}

Let us reflect for a moment on the three conditions stated in  Proposition~\ref{prop:full_simplex}. 
The first two conditions involve the subdifferential and gradient of $-\HH$ and its conjugate; the third condition is the margin property of $L_{-\HH}$. To provide some intuition, consider the case where $\HH$ is separable with $\HH(\p) = \sum_i h(\pp_i)$ and $h$ is differentiable in $(0,1)$. Then, from the concavity of $h$, its derivative $h'$ is decreasing, hence the first condition  is met if $\lim_{t=0^+} h'(t) < \infty$ and $\lim_{t=1^-} h'(t) > -\infty$. 
This is the case with Tsallis entropies for $\alpha > 1$, but not Shannon
entropy, since
$h'(t) = -1-\log t$ explodes at $0$.
Functions whose gradient ``explodes'' in the boundary of their domain (hence
failing to meet the first condition in Proposition~\ref{prop:full_simplex}) are
called ``essentially smooth'' \citep{Rockafellar1970}.  
For those functions, $\nabla (-\HH)^*$ maps only to the relative interior of
$\triangle^d$, never attaining boundary points \citep{wainwright_2008}; this is expressed in the second condition.  
This prevents essentially smooth functions from generating a sparse $\y_{-\HH} \equiv \nabla (-\HH)^*$ or (if they are separable) a
loss $L_{-\HH}$ with a margin, as asserted by the third condition. 
Since Legendre-type functions (\S\ref{sec:fy_losses})
are strictly convex \textit{and} essentially smooth,
by Proposition \ref{prop:Bregman_div}, loss functions for which
the composite form $L_{-\HH}(\s; \y) = B_{-\HH}(\y || \widehat{\y}_{-\HH}(\s))$ holds, which is the
case of the logistic loss but not of the sparsemax loss, do not enjoy a margin
and cannot induce a sparse probability distribution.
This is geometrically visible in Figure~\ref{fig:entropy_max_argmax}.  

\paragraph{Margin computation.} 

For Fenchel-Young losses that have the separation margin property, Proposition~\ref{prop:margin} provided a formula for determining the margin. 
While informative, formula \eqref{eq:margin} is not very practical, 
as it involves a generally non-convex optimization problem.
The next proposition, proved in
\S\ref{appendix:proof_margin_separable}, takes a step  further and
provides a remarkably simple closed-form expression for  generalized
entropies that are {\bf twice-differentiable}. 
To simplify notation, we denote by $\nabla_j \HH(\p) \equiv (\nabla \HH(\p))_j$
the $j^{\text{th}}$ component of $\nabla \HH(\p)$. 
\vspace{0.5em}
\begin{proposition}\label{prop:margin_separable}
Assume $\HH$ satisfies the conditions in Proposition~\ref{prop:full_simplex} 
and is twice-differ\-entiable on the simplex. Then, for arbitrary $j \ne k$:
\begin{equation}\label{eq:margin_general}
\mathrm{margin}(L_{-\HH}) = {\nabla_j{\HH}(\e_k)} - {\nabla_k{\HH}(\e_k)}.
\end{equation}
In particular, if $\HH$ is separable, i.e., $\HH(\p) = \sum_{i=1}^{|\cY|}
h(\pp_i)$, where $h:[0,1] \rightarrow \RR_{+}$ is concave, twice differentiable,
with $h(0) = h(1) = 0$, then
\begin{equation}\label{eq:margin_separable}
    \mathrm{margin}(L_{-\HH}) = h'(0) - h'(1) =-\hspace{-.2ex}\int_{0}^1\hspace{-.4ex}h''(t) dt.
\end{equation}
\end{proposition}

The compact formula \eqref{eq:margin_general} provides a geometric characterization of separable entropies and their margins: 
\eqref{eq:margin_separable} tells us that only the slopes of $h$ at the two extremities of %the interval
$[0,1]$ are relevant in determining the margin.
% VN: this is a nice thought but not so important.
%     and, as Andre admitted, you can design such an entropy anyway
%     by scaling sparsemax.
% It is also a \textbf{constructive result}, since it allows to design an entropy that leads to a loss with a prescribed margin.

\paragraph{Example: case of Tsallis and norm entropies.} 
%Let us apply directly formula \eqref{eq:margin_separable} to obtain margin values of Tsallis and norm entropies. 
As seen in \S\ref{sec:proba_clf}, Tsallis entropies are separable with $h(t) = (t - t^\alpha)/(\alpha(\alpha-1))$. 
For $\alpha > 1$, $h'(t) = (1 - \alpha t^{\alpha -
1})/(\alpha(\alpha-1))$, hence $h'(0)=1/(\alpha(\alpha-1))$ and
$h'(1)=-1/\alpha$. Proposition~\ref{prop:margin_separable} then yields 
\begin{equation}
    \mathrm{margin}(L_{-\HHt_\alpha}) = h'(0) - h'(1) = (\alpha-1)^{-1}. %\frac{1}{\alpha - 1}.
\end{equation}
Norm entropies, while not separable, have gradient
$\nabla \HHn_q(\p) = -( \nicefrac{\p}{\|\p\|_q} )^{q-1}$, 
giving $\nabla \HHn_q(\e_k) = -\e_k$, so
\begin{equation}
\mathrm{margin}(\HHn_q) =
\nabla_j \HHn_q(\e_k) - \nabla_k \HHn_q(\e_k) = 1, 
\end{equation}
as confirmed visually in Figure~\ref{fig:entropy_max_argmax},
in the binary case. 

%\begin{proposition}\label{prop:margin_known_entropies}
%\begin{enumerate}[topsep=0pt,itemsep=3pt,parsep=3pt,leftmargin=15pt]
%\item Let ${\HHt_\alpha}$ be the Tsallis $\alpha$-entropy. Then, for $\alpha \in
%    [0,1]$, $L_{-{\HHt_{\alpha}}}$ does not
%    have the separation margin property. For $\alpha>1$, we have
%    $\mathrm{margin}(L_{-\HHt_\alpha}) = 1/(\alpha-1)$. 
%\item Let ${\HHn_q}$ be the $q$-norm entropy. Then, for any $q \ge 1$, we have $\mathrm{margin}(L_{-{\HHn_{q}}}) = 1$.
%\end{enumerate}
%\end{proposition}
%\begin{proof}
%As seen in \S\ref{sec:categorical_distrib}, Tsallis entropies are separable with $h(t) = (t - t^\alpha)/(\alpha(\alpha-1))$. 
%For $\alpha > 1$, the derivative is $h'(t) = (1 - \alpha t^{\alpha -
%1})/(\alpha(\alpha-1))$, so we have $h'(0)=1/(\alpha(\alpha-1))$ and
%$h'(1)=-1/\alpha$. Proposition~\ref{prop:margin_separable} then yields
%$\mathrm{margin}(L_{-\HHt}) = h'(0) - h'(1) = 1/(\alpha - 1)$. 
%Norm entropies are not separable, but their gradient can be expressed as
%$\nabla \HHn_q(\p) = -( {\p}/{\|\p\|_q} )^{q-1}$. 
%We thus have $\nabla \HHn_q(\e_k) = -\e_k$, so $\mathrm{margin}(\HHn_q) =
%\nabla_j \HHn_q(\e_k) - \nabla_k \HHn_q(\e_k) = 1$. 
%\end{proof}

\section{Experimental results}
\label{sec:experiments}

As we saw, $\alpha$-Tsallis entropies
generate a family of losses, with the logistic ($\alpha \to 1$) and sparsemax losses
($\alpha=2$) as important special cases.
In addition, they are
twice differentiable for $\alpha \in
[1,2)$, produce sparse probability distributions for $\alpha > 1$ and are
computationally efficient for any $\alpha \ge 1$, thanks to Proposition~
\ref{prop:root_finding}. In this section, we demonstrate their usefulness on
the task of label proportion estimation and compare different solvers for
computing $\widehat{\y}_{-\HHt_\alpha}$.

\begin{figure*}
\begin{floatrow}
%\begin{table*}[t]
\capbtabbox{%
\label{tab:label_proportion}
\centering
\small
    \begin{tabular}{r c c c c}
        \toprule
        & $\alpha=1$ & $\alpha=1.5$ & $\alpha=2$ & tuned
        $\alpha$ \\
        & (logistic) & & (sparsemax) &\\
       %$\alpha$ & $1$ & $1.5$ & $2$ & tuned \\
        \midrule
Birds & 0.359 / 0.530 & 0.364 / 0.504 & 0.364 / 0.504 & {\bf 0.358} / {\bf 0.501} \\
Cal500 & 0.454 / {\bf 0.034} & 0.456 / 0.035 & {\bf 0.452} / 0.035 & 0.456 / {\bf 0.034} \\
Emotions & 0.226 / 0.327 & 0.225 / {\bf 0.317} & 0.225 / {\bf 0.317} & {\bf 0.224} / 0.321 \\
Mediamill & 0.375 / 0.208 & 0.363 / 0.193 & {\bf 0.356} / {\bf 0.191} & 0.361 / 0.193 \\
Scene & {\bf 0.175} / {\bf 0.344} & 0.176 / 0.363 & 0.176 / 0.363 & {\bf 0.175} / 0.345 \\
TMC & 0.225 / 0.337 & 0.224 / {\bf 0.327} & 0.224 / {\bf 0.327} & {\bf 0.217} / 0.328 \\
Yeast & {\bf 0.307} / {\bf 0.183} & 0.314 / 0.186 & 0.314 / 0.186 & {\bf 0.307} / {\bf 0.183} \\
\midrule
Avg.\ rank & 2.57 / 2.71 & 2.71 / 2.14 & 2.14 / 2.00 & \textbf{1.43} /
\textbf{1.86} \\
        \bottomrule
    \end{tabular}}{\caption{\textbf{Test-set performance of Tsallis losses for
        various $\alpha$ on the task of sparse label proportion estimation:}
        average Jensen-Shannon divergence (left) and mean squared error (right).
        Lower is better.}}%
\ffigbox[\FBwidth][\FBheight][b]{%
\hspace{-.2cm}%
\centering%
\footnotesize \input{figures/bisection_time_to_acc.pgf}\vspace{-.2cm}}%
{\caption{\label{fig:solver_comparison}Median time until $10^{-5}$ accuracy 
is met for computing $\yHat_{-\HHt_{1.5}}$.}}
\end{floatrow}
\end{figure*}

\paragraph{Label proportion estimation.}

Given an input vector $\x \in \cX \subseteq \RR^p$, where $p$ is the number of
features, our goal is to estimate a vector of label proportions $\y \in
\triangle^d$, where $d$ is the number of classes.  If $\y$ is sparse, we expect
the superiority of Tsallis losses over the conventional logistic loss on this
task.  At training time, given a set of $n$ $(\x_i,\y_i)$ pairs, we estimate a
matrix $W \in \RR^{d \times p}$ by minimizing the convex objective
\begin{equation}
R(W) \coloneqq 
\sum_{i=1}^n L_\Omega(W \x_i; \y_i) + \frac{\lambda}{2} \|W\|_F^2.
\end{equation}
We use L-BFGS \citep{lbfgs} for simplicity.
From Proposition \ref{prop:fy_losses} and using the chain rule, we obtain the
gradient expression
$\nabla R(W) = (\widehat{Y}_\Omega - Y)^\top X + \lambda W$,
where $\widehat{Y}_\Omega$, $Y$ and $X$ are matrices whose rows
gather $\yHatOmega(W \x_i)$, $\y_i$ and $\x_i$, for $i=1,\dots,n$.
At test time, we predict label proportions by 
$\p = \y_{-\HHt_\alpha}(W \x)$.

We ran experiments on $7$ standard multi-label benchmark datasets --- see
\S\ref{appendix:exp} for dataset characteristics.  For all datasets, we removed
samples with no label, normalized samples to have zero mean unit variance, and
normalized labels to lie in the probability simplex.  We chose $\lambda \in
\{10^{-4},10^{-3},\dots,10^4\}$ and $\alpha \in \{1,1.1,\dots,2\}$ against the
validation set.  We report the test set mean Jensen-Shannon divergence,
$\text{JS}(\p, \y) \coloneqq \frac{1}{2} \text{KL}(\p || \frac{\p+\y}{2}) +
\frac{1}{2} \text{KL}(\y || \frac{\p+\y}{2})$, and the mean squared error
$\frac{1}{2} \|\p - \y\|^2$ in Table \ref{tab:label_proportion}. 
As can be seen, the loss with tuned $\alpha$ achieves the best averaged rank
overall.
Tuning $\alpha$ allows to choose the best loss in the family in a
data-driven fashion. Additional experiments confirm these
findings --- see \S\ref{appendix:exp}.

\paragraph{Solver comparison.}

Next, we compared bisection (binary search) and Brent's method for solving
\eqref{eq:prediction} by root finding (Proposition~\ref{prop:root_finding}).
We focus on $\HHt_{1.5}$, i.e.\ the $1.5$-Tsallis entropy, and also
compare against using a generic projected gradient algorithm
(FISTA) to solve \eqref{eq:prediction} naively.
We measure the time needed 
to reach a solution $\p$ with $\| \p - \p^\star \|_2 < 10^{-5}$,
over 200 samples $\s \in \RR^d \sim \mathcal{N}(\bm{0}, \sigma
\bm{I})$ with $\log\sigma\sim\mathcal{U}(-4, 4)$.
Median and 99\% CI times reported in 
Figure~\ref{fig:solver_comparison}
reveal that root finding scales better,
with Brent's method outperforming FISTA by one to two orders of magnitude.

\section{Related work}
\label{sec:related_work}

\textbf{Proper scoring rules (proper losses)} are a well-studied object
in statistics \citep{grunwald_2004,gneiting_2007} and machine
learning \citep{reid_composite_binary,vernet_2016},
that measures the discrepancy between 
a ground-truth $\y \in \triangle^d$ and a probability forecast $\p \in
\triangle^d$ in a \textbf{Fisher-consistent} manner.
From \citet{savage_1971} (see also \citet{gneiting_2007}),
we can construct a proper
scoring rule $S_\Omega \colon \triangle^d \times \triangle^d \to \RR_+$ by
\begin{equation}
    S_\Omega(\p; \y) 
\coloneqq
\langle \nabla \Omega(\p), \y - \p \rangle - \Omega(\p) 
= B_\Omega(\y||\p) - \Omega(\y),
\label{eq:S_Omega}
\end{equation}
recovering the well-known relation between 
Bregman divergences and proper scoring rules.
For example, using the Gini index $\HH(\p) = 1 - \|\p\|^2$
generates the \textbf{Brier score} \citep{brier_1950}
$S_{-\HH}(\p; \e_k) =
\sum_{i=1}^d ([[k = i]] - p_i)^2$,
showing that the sparsemax loss and the Brier score share the same generating
function.
More generally, while a scoring rule $S_\Omega$ is related to a \textbf{primal-space} Bregman
divergence, a Fenchel-Young loss $L_\Omega$ can be seen as a
\textbf{mixed-space} Bregman divergence (\S\ref{sec:fy_losses}). 
This difference has a number of
important consequences. First,
$S_\Omega$ is \textbf{not necessarily convex} in $\p$
(\citet[Proposition 17]{vernet_2016} show that it is in fact quasi-convex).
%This is consistent with the well-known fact that $B_\Omega(\y || \p)$ is convex
%in $\y$ but not necessarily in $\p$. 
In contrast,
$L_\Omega$ \textbf{is always} convex in $\s$. Second, the first argument is
\textbf{constrained} to $\triangle^d$ for $S_\Omega$, while
\textbf{unconstrained} for $L_\Omega$.

In practice, proper scoring rules (losses) are often
composed with an \textbf{invertible} link function $\bpsi^{-1} \colon \RR^d \to
\triangle^d$. This form of a loss, $S_\Omega(\bpsi^{-1}(\s); \y)$, is sometimes
called composite \citep{buja_2005,reid_composite_binary,vernet_2016}.  
Although the decoupling between loss and link has merits
\citep{reid_composite_binary}, the composition of $S_\Omega(\cdot; \y)$ and $\bpsi^{-1}(\s)$ is not
necessarily convex in $\s$.  The \textbf{canonical link function}
\citep{buja_2005} of $S_\Omega$ is a link function that ensures the convexity of
$S_\Omega(\bpsi^{-1}(\s); \y)$ in $\s$. It also plays a key role in
generalized linear models \citep{glm}.  
Following Proposition \ref{prop:Bregman_div},
when $\Omega$ is Legendre type, we obtain
\begin{equation}
L_\Omega(\s; \y) = B_\Omega(\y || \yHatOmega(\s)) = S_\Omega(\yHatOmega(\s);
\y) + \Omega(\y).
\end{equation}
Thus, in this case, Fenchel-Young losses and proper
composite losses coincide up to the constant term $\Omega(\y)$ (which vanishes
if $\y = \e_k$ and $\Omega$ satisfies assumption A.1), with $\bpsi^{-1} =
\yHatOmega$ the canonical inverse link function.  Fenchel-Young losses, however,
require neither invertible link nor Legendre type assumptions, allowing to
express losses (e.g., hinge or sparsemax) that are not expressible in composite
form.  Moreover, as seen in \S\ref{sec:margin}, a Legendre-type $\Omega$
precisely precludes sparse probability distributions and losses enjoying a
margin.

\paragraph{Other losses.} 

\cite{nock_2009} proposed binary classification losses based
on the Legendre transformation but require invertible mappings.
\citet{masnadi_2011} studied the Bayes consistency of related binary
classification loss functions.
%but without noticing the connection with Fenchel conjugacy and generalized
%entropies.
\citet[Proposition 3]{duchi_2016} derived the multi-class loss
\eqref{eq:fy_loss_multiclass2}, a special case of Fenchel-Young loss over the
probability simplex, and showed (Proposition 4) that any strictly concave
generalized entropy generates a classification-calibrated loss. 
\citet{amid_2017} proposed a different family of losses based on the Tsallis
divergence, to interpolate between convex and non-convex losses, for
robustness to label noise. 

\paragraph{Smoothing techniques.}

Fenchel duality also plays a key role in smoothing techniques
\citep{nesterov_smooth,beck_2012},
which have been used extensively to create smoothed losses \citep{accelerated_sdca}.
However, these techniques were applied on a per-loss basis and were not
connected to the induced probability distribution. In contrast, we propose a
generic construction, with clear links between smoothing
and the distribution induced by $\yHatOmega$.

\section{Conclusion}

We showed that regularization and Fenchel duality provide simple core
principles, unifying many
existing loss functions, and allowing to create useful new ones easily.
In particular, we derived a new family of loss functions based on Tsallis
entropies, which includes the logistic, sparsemax and perceptron losses as
special cases. With the unique exception of the logistic loss, losses in this
family induce sparse probability distributions.
%We studied Fenchel-Young losses in a broad setting and demonstrated that they not
%only unify many existing loss functions but also allow to create new useful ones easily.
We also showed a close and fundamental relationship between generalized
entropies, losses enjoying a margin and sparse probability distributions.
Remarkably, Fenchel-Young losses can be defined
over arbitrary domains, allowing to construct loss functions
for a large variety of applications \citep{journal_version}.

\clearpage

\subsubsection*{Acknowledgements}

MB thanks Arthur Mensch and Gabriel Peyr\'{e} for numerous
fruitful discussions and Tim Vieira for introducing him to generalized
exponential families.  AM and VN were partially 
supported by the European
Research Council (ERC StG DeepSPIN 758969) and by the Funda\c{c}\~ao para a
Ci\^encia e Tecnologia through contracts UID/EEA/50008/2013 and
CMUPERI/TIC/0046/2014 ~(GoLocal).

%\bibliography{aistats}
%\bibliographystyle{abbrvnat}
%\bibliographystyle{plainnat}

\input{aistats.bbl}
\clearpage
\onecolumn
\appendix

\begin{center}
\Huge Appendix
\end{center}

\section{More examples of generalized entropies}
\label{appendix:more_entropies}

In this section, we give two more examples of generalized entropies: squared
norm entropies and R\'enyi entropies.

\paragraph{Squared norm entropies.}

Inspired by \citet{Niculae2017}, as a simple extension of the Gini index
\eqref{eq:gini_entropy}, we consider the following generalized entropy based on
squared $q$-norms:
\begin{equation}
\HHsq_{q}(\p) 
\coloneqq \frac{1}{2} (1 - \|\p\|^2_q) 
= \frac{1}{2} - \frac{1}{2} \left(\sum_{j=1}^d p_j^q \right)^{\frac{2}{q}}.
\end{equation}
The constant term $\frac{1}{2}$, omitted by \citet{Niculae2017},
ensures satisfaction of A.1. 
For $q \in (1,2]$, it is known that the squared $q$-norm is strongly convex \wrt
$\|\cdot\|_{q}$ \citep{ball_1994}, implying that $(-\HHsq_{q})^*$, and therefore
$L_{-\HHsq_{q}}$, is smooth.
Although $\widehat{\y}_{-\HHsq_q}(\s)$ cannot be solved in closed form for $q
\in (1,2)$, it can be solved efficiently using projected gradient descent
methods. 

\paragraph{R\'enyi {\boldmath $\beta$}-entropies.} 

R\'enyi entropies \citep{Renyi1960} are defined for any $\beta\ge 0$ as:
\begin{equation}\label{eq:renyi_entropies}
\HHr_{\beta}(\p) \coloneqq \frac{1}{1-\beta} \log \sum_{j=1}^d
p_j^\beta.
\end{equation}
Unlike Shannon and Tsallis entropies, R\'enyi entropies are not separable, with
the exception of $\beta \rightarrow 1$, which also recovers Shannon entropy as a
limit case. The case $\beta \rightarrow +\infty$ gives $\HHr_{\beta}(\p) = -\log \|\p\|_{\infty}$. 
For $\beta \in [0, 1]$, R\'enyi entropies satisfy assumptions
A.1--A.3; for $\beta > 1$, R\'enyi entropies fail to be concave.  They are
however pseudo-concave \citep{Mangasarian1965}, meaning that, for all $\p,
\bm{q} \in \triangle^d$,  $\DP{\nabla \HHr_\beta(\p)}{\bm{q} - \p} \le 0$ implies
$\HHr_\beta(\bm{q}) \le \HHr_\beta(\p)$.  This implies, among other things, that
points $\p \in \triangle^d$ with zero gradient are maximizers of $\DP{\p}{\s} +
\HHr_{\beta}(\p)$, which allows us to compute the predictive distribution
$\widehat{\y}_{-\HHr_{\beta}}$ with gradient-based methods.  

\begin{figure}[h]
\centering
\input{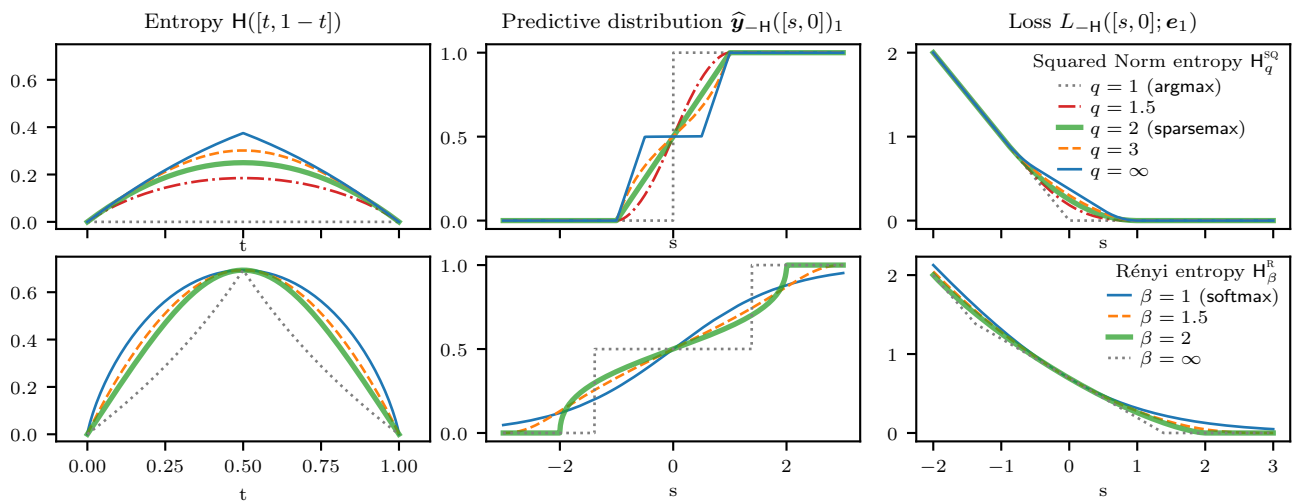}
\caption{Squared norm and R\'enyi entropies, together with the distributions
and losses they generate.}
\end{figure}

\newpage
\section{Experiment details and additional empirical results}
\label{appendix:exp}

\paragraph{Benchmark datasets.}

The datasets we used in \S\ref{sec:experiments} are summarized below.

\begin{table*}[h]
    \caption{Dataset statistics}
    \small
    \centering
    \begin{tabular}{r c c c c c c c}
        \toprule
        Dataset & Type & Train & Dev & Test & Features &
        Classes & Avg. labels \\
        \midrule
        Birds & Audio & 134 & 45 & 172 & 260 & 19 & 2 \\
        Cal500 & Music & 376 & 126 & 101 & 68 & 174 & 25 \\
        Emotions & Music & 293 & 98 & 202 & 72 & 6 & 2 \\
        Mediamill & Video & 22,353 & 7,451 & 12,373 & 120 & 101 & 5 \\
        Scene & Images & 908 & 303 & 1,196 & 294 & 6 & 1\\
        SIAM TMC & Text & 16,139 & 5,380 & 7,077 & 30,438 & 22 & 2\\
        Yeast & Micro-array & 1,125 & 375 & 917 & 103 & 14 & 4\\
        \bottomrule
    \end{tabular}
\end{table*}

The datasets can be downloaded from
\url{http://mulan.sourceforge.net/datasets-mlc.html} and
\url{https://www.csie.ntu.edu.tw/~cjlin/libsvmtools/datasets/}.

\paragraph{Sparse label proportion estimation on synthetic data.}

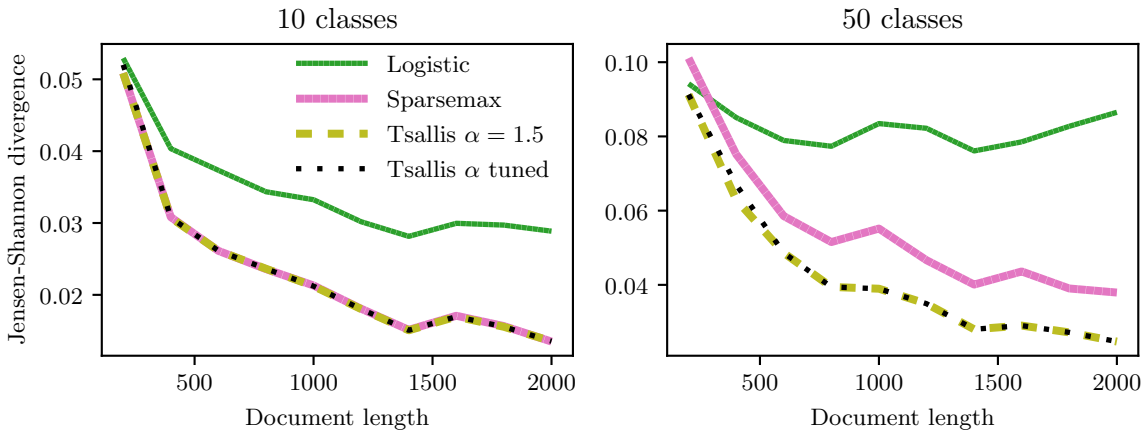
\begin{figure}[h]
\centering
\input{figures/exp_label_proportions.pgf}
\caption{Jensen-Shannon divergence between predicted and true label
    proportions, when varying document length, of various losses generated by a
Tsallis entropy.}
\label{fig:label_proportions_synth}
\end{figure}

We follow \citet{sparsemax} and generate a document
$\x \in \RR^p$ from a mixture of multinomials and label proportions $\y \in
\triangle^d$ from a multinomial. The number of words in $\x$ and labels in $\y$
is sampled from a Poisson distribution --- see \citet{sparsemax} for a precise
description of the generative process.  We use 1200 samples as training set, 200
samples as validation set and 1000 samples as test set.
We tune $\lambda \in \{10^{-6}, 10^{-5}, \dots, 10^0\}$ 
and $\alpha \in \{1.0, 1.1, \dots, 2.0\}$ against the validation set.
We report the Jensen-Shannon divergence in Figure
\ref{fig:label_proportions_synth}.  Results using the mean squared error (MSE)
were entirely similar.
When the number of classes is 10, we see that Tsallis and sparsemax losses
perform almost exactly the same, both outperforming softmax.
When the number of classes is 50, Tsallis losses outperform both sparsemax and
softmax.

\newpage
\section{Proofs}
\label{appendix:proofs}

In this section, we give proofs omitted from the main text.

\subsection{Proof of Proposition~\ref{prop:prediction_func}}
\label{appendix:proof_prediction_func}

\paragraph{Effect of a permutation.}

Let $\Omega$ be symmetric.  We first prove that $\Omega^*$ is
symmetric as well. Indeed, we have
\begin{equation}
\Omega^*(\bm{P}\s) 
= \sup_{\p \in \dom(\Omega)} (\bm{P} \s)^\top \p - \Omega(\p) 
= \sup_{\p \in \dom(\Omega)} \s^\top \bm{P}^\top
\p - \Omega(\bm{P}^\top \p) = \Omega^*(\s).  
\end{equation}
The last equality was obtained by
a change of variable $\p' = \bm{P}^\top \p$, from which $\p$ is
recovered as $\p = \bm{P}\p'$, which proves $\nabla \Omega^*(\bm{P} \p)
= \bm{P} \nabla \Omega^*(\p)$.  

\paragraph{Order preservation.}

Since $\Omega^*$ is convex, the gradient operator $\nabla
\Omega^*$ is monotone, i.e.,
\begin{equation}
(\s' - \s)^{\top} (\p'-\p) \ge 0
\end{equation}
for any
$\s, \s' \in \RR^d$, $\p=\nabla \Omega^*(\s)$ and $\p'=\nabla
\Omega^*(\s')$.  Let $\s'$ be obtained from $\s$ by swapping two coordinates,
i.e., $\ss_j'=\ss_i$, $\ss_i'=\ss_j$, and $\ss_k'=\ss_k$ for any $k \notin
\{i,j\}$. Then, since $\Omega$ is symmetric, we obtain:
\begin{eqnarray}
2(\ss_j - \ss_i)(p_j-p_i)\ge 0,
\end{eqnarray}
which implies $\ss_i > \ss_j \Rightarrow p_i \ge p_j$ and $p_i > p_j
\Rightarrow \ss_i \ge \ss_j$. To fully prove the claim, we need to show that the last
inequality is strict: to do this, we simply invoke
$\nabla \Omega^*(\bm{P} \p) = \bm{P} \nabla \Omega^*(\p)$
with a matrix $\bm{P}$ that permutes $i$ and $j$, from which we must have $\ss_i
= \ss_j \Rightarrow p_i = p_j$. 

\paragraph{Gradient mapping.} This follows directly from Danskin's theorem
\citep{danskin_theorem}. See also \citet[Proposition B.25]{bertsekas_book}.

\paragraph{Temperature scaling.} 

This immediately follows from properties of the $\argmax$ operator.

\subsection{Proof of Proposition \ref{prop:Bregman_div}}
\label{appendix:proof_Bregman_div}

We set $\Omega \coloneqq \Psi + I_\cC$.

\paragraph{Bregman projections.}

If $\Psi$ is Legendre type, then $\nabla \Psi(\nabla \Psi^*(\s)) = \s$ for
all $\s \in \interior(\dom(\Psi^*))$, where $\interior(\cD)$ denotes the
interior of $\cD$. Using this and our assumption that
$\dom(\Psi^*)=\RR^d$, we get for all $\s \in \RR^d$:
\begin{equation}
B_\Psi(\p || \nabla \Psi^*(\s)) 
= \Psi(\p) - \DP{\s}{\p} + \DP{\s}{\nabla \Psi^*(\s)} - \Psi(\nabla
\Psi^*(\s)).
\label{eq:Bregman_div_nabla_Omega_conj}
\end{equation}
The last two terms are independent of $\p$ and therefore
\begin{equation}
\yHatOmega(\s)
= \argmax_{\p \in \cC} \DP{\s}{\p} - \Psi(\p)
= \argmin_{\p \in \cC} B_\Psi(\p||\nabla \Psi^*(\s)),
\end{equation}
where $\cC \subseteq \dom(\Psi)$.
The r.h.s. is the
Bregman projection of $\nabla \Psi^*(\s)=\widehat{\y}_\Psi(\s)$ onto $\cC$.

\paragraph{Difference of Bregman divergences.}

Let $\p = \yHatOmega(\s)$.
Using \eqref{eq:Bregman_div_nabla_Omega_conj}, we obtain
\begin{align}
B_\Psi(\y || \nabla \Psi^*(\s)) - 
B_\Psi(\p || \nabla \Psi^*(\s))
&= \Psi(\y) - \DP{\s}{\y} + \DP{\s}{\p} - \Psi(\p)
\\
&= \Omega(\y) - \DP{\s}{\y} + \Omega^*(\s) \\
&= L_{\Omega}(\s; \y),
\label{eq:Bregman_diff}
\end{align}
where we assumed $\y \in \cC$ and $\cC \subseteq \dom(\Psi)$,
implying $\Psi(\y) = \Omega(\y)$.

If $\cC = \dom(\Psi)$ (i.e., $\Omega = \Psi$), then $\p = \nabla
\Psi^*(\s)$ and $B_\Psi(\p || \nabla \Psi^*(\s)) = 0$. We thus get
the \textbf{composite form} of Fenchel-Young losses
\begin{equation}
B_\Omega(\y || \nabla \Omega^*(\s))    
= B_\Omega(\y || \yHatOmega(\s))    
= L_\Omega(\s; \y).
\end{equation}

\paragraph{Bound.}

Let $\p = \yHatOmega(\s)$. Since $\p$ is the Bregman projection of $\nabla
\Psi^*(\s)$ onto $\cC$,
we can use the well-known Pythagorean theorem for
Bregman divergences (see, e.g., \citet[Appendix A]{bregman_clustering}) to
obtain for all $\y \in \cC \subseteq \dom(\Psi)$:
\begin{equation}
B_\Psi(\y || \p) + B_\Psi(\p || \nabla \Psi^*(\s))
\le B_\Psi(\y || \nabla \Psi^*(\s)).
\end{equation}
Using \eqref{eq:Bregman_diff}, we obtain for all $\y \in \cC \subseteq
\dom(\Psi)$:
\begin{equation}
    0 \le B_\Psi(\y || \p) = B_\Omega(\y || \p) \le L_{\Omega}(\s; \y).
\end{equation}

Since $\Omega$ is a l.s.c.\ proper convex
function, from Proposition \ref{prop:fy_losses}, we immediately get
\begin{equation}
\p = \y 
\Leftrightarrow
L_{\Omega}(\s; \y) = 0
\Leftrightarrow
B_\Omega(\y || \p) = 0.
\end{equation}

\subsection{Proof of Proposition \ref{prop:generalized_entropy}}
\label{appendix:proof_generalized_entropy}

The two facts stated in Proposition \ref{prop:generalized_entropy} ($\HH$ is
always non-negative and maximized by the uniform distribution) follow directly
from Jensen's inequality.  Indeed, for all $\p \in \triangle^d$:
\begin{itemize}

\item $\HH(\p) \ge \sum_{j=1}^d p_j \HH(\e_j) = 0$; 

\item $\HH(\mathbf{1}/ d) = \HH\left(\sum_{\bm{P} \in \mathcal{P}} \frac{1}{d!}
\bm{P}\p \right) \ge \sum_{\bm{P} \in \mathcal{P}} \frac{1}{d!}
\HH(\bm{P} \p) = \HH(\p)$,

\end{itemize}
where $\mathcal{P}$ is the set of $d \times d$ permutation matrices. 
Strict concavity ensures that $\p = \ones / d$ is the unique maximizer.

\subsection{Proof of Proposition~\ref{prop:root_finding}}
\label{appendix:proof_root_finding}

Let $\Omega(\p) = \sum_{j=1}^d g(p_j) + I_{\triangle^d}(\p)$, where
$g \colon [0,1]\rightarrow\RR_+$ is a
non-negative, strictly convex, differentiable function.
Therefore, $g'$ is strictly monotonic on $[0,1]$, thus invertible.
We show how computing $\nabla(\Omega)^*$ reduces to finding the root of
a monotonic scalar function, for which efficient algorithms 
are available.

From strict convexity and the definition of the convex conjugate,
\begin{equation}
    \nabla \Omega^*(\s) = \argmax_{\p\in\triangle^d}
    \DP{\p}{\s} - \sum_j g(\pp_j).
\end{equation}
The constrained optimization problem above has Lagrangian
\begin{equation}
\mathcal{L}(\p, \bm{\nu}, \thresh) \coloneqq 
\sum_{j=1}^d g(\pp_j) - \DP{\s + \bm{\nu}}{\p} + \thresh (\mathbf{1}^\top \p - 1).
\end{equation}
A solution
$(\p^\star, \bm{\nu}^\star, \thresh^\star)$
must satisfy the KKT conditions
\begin{empheq}[left=\empheqlbrace]{align}
    g'(\pp_j) - \ss_j - \nu_j + \thresh &=0 \qquad \forall j \in [d]
        \label{eqn:stationarity_separable}\\
    \DP{\p}{\bm{\nu}}&= 0
        \label{eqn:complementary_slack_separable}\\
    \p \in \triangle^d, \,\, \bm{\nu}&\ge 0.
\end{empheq}
Let us define
\[
    \thresh_{\min} \coloneqq \max(\s) - g'(1)
    \quad \text{and} \quad
    \thresh_{\max} \coloneqq \max(\s) - g'\left(\frac{1}{d}\right).
\]
Since $g$ is strictly convex, $g'$ is increasing and so
$\thresh_{\min}<\thresh_{\max}$. For any $\thresh \in
[\thresh_{\min}, \thresh_{\max}]$,
we construct $\bm{\nu}$ as
\[
    \nu_j \coloneqq \begin{cases}
        0, & \ss_j - \thresh \geq g'(0) \\
        g'(0) - \ss_j + \thresh, & \ss_j - \thresh < g'(0) \\
    \end{cases}
\]
By construction, $\nu_j \geq 0$, satisfying dual feasability.
Injecting $\nu$ into \eqref{eqn:stationarity_separable}
and combining the two cases, we obtain
\begin{equation}\label{eq:bisect-station-each}
    g'(\pp_j) = \max\{\ss_j - \thresh,~g'(0)\}.
\end{equation}

We show that i) the stationarity conditions have a unique solution given
$\thresh$, and ii) $[\thresh_{\min}, \thresh_{\max}]$ forms a
sign-changing bracketing interval, and thus contains $\tau^\star$, which can
then be found by one-dimensional search. The solution verifies all KKT conditions,
thus is globally optimal.

\paragraph{Solving the stationarity conditions.}
Since $g$ is strictly convex, its derivative $g'$ is continuous and strictly
increasing, and is thus a one-to-one mapping between $[0, 1]$ and
$[g'(0), g'(1)]$. Denote by $(g')^{-1} \colon [g'(0), g'(1)] \rightarrow [0, 1]$
its inverse. If $\ss_j - \thresh \ge g'(0)$,
we have 
\begin{equation}
\begin{aligned}
    g'(0) \leq 
g'(\pp_j) = \ss_j - \thresh 
                                    &\leq \max(\s) - \thresh_{\min} \\
                                    &= \max(\s) - \max(\s) + g'(1) \\
                                    &= g'(1). \\
\end{aligned}
\end{equation}
Otherwise,
$g'(\pp_j) = g'(0)$. This verifies that the r.h.s.\ of \eqref{eq:bisect-station-each}
is always within the domain of $(g')^{-1}$. We can thus apply the inverse to both
sides to solve for $\pp_j$, obtaining
\begin{equation}\label{eq:bisectprimaldual}
    \pp_j(\thresh) = (g')^{-1}(\max\{\ss_j - \thresh,~g'(0)\}).
\end{equation}
Strict convexity implies the optimal $\p^\star$ is unique; it can be seen
that $\tau^\star$ is also unique. Indeed, assume optimal $\tau^\star_1,
\tau^\star_2$. Then, $\p(\tau^\star_1) = \p(\tau^\star_2)$, so
$\max(\s-\tau^\star_1, g'(0)) =  \max(\s-\tau^\star_2, g'(0))$.
This implies either $\tau^\star_1 = \tau^\star_2$, or $\s -
\tau^\star_{\{1,2\}} \leq g'(0)$, in which case $\p=\zeros \notin \triangle^d$,
which is a contradiction.

\paragraph{Validating the bracketing interval.}

Consider the primal infeasability function
$\phi(\thresh) \coloneqq \DP{\p(\thresh)}{\ones} - 1$;
$\p(\thresh)$ is primal feasible iff $\phi(\thresh)=0.$
We show that $\phi$ is decreasing on $[\thresh_{\min}, \thresh_{\max}]$, and that
it has opposite signs at the two extremities.
From the intermediate value theorem, the unique
root $\thresh^\star$ must satisfy $\thresh^\star\in[\thresh_{\min}, \thresh_{\max}]$.

Since $g'$ is increasing, so is $(g')^{-1}$.
Therefore, for all $j$, $\pp_j(\thresh)$ is decreasing, and so is the
sum $\phi(\thresh) = \sum_j \pp_j(\thresh) - 1$. It remains to check the signs
at the boundaries.
\begin{equation}
\begin{aligned}
    \sum_i \pp_i(\thresh_{\max})
    &= \sum_i (g')^{-1}(\max\{\ss_i - \max(\s) +
    g'\left(\nicefrac{1}{d}\right),~g'(0)\})\\
    &\leq d ~ (g')^{-1}(\max\{g'\left(\nicefrac{1}{d}\right),~g'(0)\})\\
    &= d ~ (g')^{-1}\left(g'\left(\nicefrac{1}{d}\right)\right) = 1,\\
\end{aligned}
\end{equation}
where we upper-bounded each term of the sum by the largest one. At the other end,
\begin{equation}
\begin{aligned}
    \sum_i \pp_i(\thresh_{\min})
    &= \sum_i (g')^{-1}(\max\{\ss_i - \max(\s) + g'(1),~g'(0)\})\\
    &\geq (g')^{-1}(\max\{g'(1),~g'(0)\})\\
    &= (g')^{-1}(g'(1)) = 1,\\
\end{aligned}
\end{equation}
using that a sum of non-negative terms is no less than its largest term.
Therefore, $\phi(\thresh_{\min}) \ge 0$ and $\phi(\thresh_{\max}) \le 0$.  This
implies that there must exist $\thresh^\star$ in $[\thresh_{\min}, \thresh_{\max}]$
satisfying $\phi(\thresh^\star)=0$.
The corresponding triplet $\left(\p(\thresh^\star), \bm{\nu}(\thresh^\star),
\thresh^\star\right)$ thus satisfies all of the KKT conditions, confirming that
it is the global solution.

Algorithm~\ref{algo:bisect} is an example of a bisection algorithm for finding
an approximate solution; more advanced root finding methods can also be used. We
note that the resulting algorithm resembles the method provided
in \cite{bregmanproj}, with a non-trivial difference being the order of the 
thresholding and $(-g)^{-1}$ in Eq.~\eqref{eq:bisectprimaldual}. 

\begin{center}
\begin{minipage}[t]{0.5\linewidth}
\begin{algorithm}[H]
    \caption{Bisection for $\yHatOmega(\s) = \nabla\Omega^*(\s)$}
\begin{algorithmic}
    %\scriptsize
    \STATE \textbf{Input:} $\s \in \RR^d$,
    $\Omega(\p)= I_{\triangle^d} + \sum_i g(\pp_i)$
\STATE $\p(\thresh) \coloneqq (g')^{-1}(\max\{\s - \thresh,g'(0)\})$
\STATE $\phi(\thresh) \coloneqq \DP{\p(\thresh)}{\ones} - 1$
\STATE $\thresh_{\min} \leftarrow \max(\s) - g'(1)$;
\STATE $\thresh_{\max} \leftarrow \max(\s) - g'\left(\nicefrac{1}{d}\right)$
\STATE  $\thresh \leftarrow (\thresh_{\min} + \thresh_{\max}) / 2$
\STATE \textbf{while} $|\phi(\thresh)| > \epsilon$
\STATE \hspace{0.4cm}\textbf{if} $\phi(\thresh) < 0$\quad$\thresh_{\max}\leftarrow\thresh$ 
\STATE \hspace{0.4cm}\textbf{else}\hspace{45pt}$\thresh_{\min}\leftarrow\thresh$ 
\STATE \hspace{0.35cm} $\thresh \leftarrow (\thresh_{\min} + \thresh_{\max}) / 2$
\STATE \textbf{Output:} $\nabla \yHatOmega(\s) \approx \p(\thresh)$
\end{algorithmic}
\label{algo:bisect}
\end{algorithm}
\end{minipage}
\end{center}

\subsection{Proof of Proposition~\ref{prop:margin}}\label{appendix:proof_margin}

We start by proving the following lemma. 
\vspace{0.5em}
\begin{lemma}\label{prop:inverse_map}
Let $\HH$ satisfy assumptions A.1--A.3. 
Then:
\begin{enumerate}
\item We have $\s \in \partial (-\HH)(\e_k)$ iff $\ss_k = (-\HH)^*(\s)$. That is:
\begin{equation}
\partial (-\HH)(\e_k) = \{\s \in \RR^d \colon \ss_k \ge \DP{\s}{\p} + \HH(\p), \,\,\forall \p \in \triangle^d\}.
\end{equation}
\item If $\s \in \partial (-\HH)(\e_k)$, 
then, we also have $\s' \in \partial (-\HH)(\e_k)$ 
for any $\s'$ such that $\ss_k' = \ss_k$ and 
$\ss_i' \le \ss_i$, for all $i \ne k$.
\end{enumerate}
\end{lemma}

\paragraph{Proof of the lemma:} 
Let $\Omega = -\HH$. 
From Proposition~\ref{prop:prediction_func} (order preservation), we can consider $\partial \Omega(\e_1)$ without loss of generality, in which case any $\s \in \partial \Omega(\e_1)$ satisfies $\ss_1 = \max_{j} \ss_j$. 
We have $\s \in \partial \Omega(\e_1)$ iff 
$\Omega(\e_1) = \DP{\s}{\e_1} - \Omega^*(\s) = \ss_1 - \Omega^*(\s)$. 
Since  $\Omega(\e_1)=0$, 
we must have $\ss_1 = \Omega^*(\s) \ge \sup_{\p\in \triangle^d} \DP{\s}{\p} - \Omega(\p)$, which proves part 1. 
To see 2, note that we have $\ss_k'=\ss_k \ge \DP{\s}{\p} - \Omega(\p) \ge \DP{\s'}{\p} - \Omega(\p)$, for all $\p \in \triangle^d$, from which the result follows. \hfill $\blacksquare$

We now proceed to the proof of Proposition~\ref{prop:margin}. 
Let $\Omega = -\HH$, and suppose that $L_{\Omega}$ has the separation margin property. Then, $\s = m\e_1$ satisfies the margin condition $\ss_1 \ge m + \max_{j\ne 1} \ss_j$, hence $L_{\Omega}(m\e_1, \e_1) = 0$. 
From the first part of Proposition~\ref{prop:fy_losses}, this implies $m\e_1 \in \partial \Omega(\e_1)$. 

Conversely, 
let us assume that $m\e_1 \in \partial \Omega(\e_1)$.
From the second part of Lemma~\ref{prop:inverse_map}, 
this implies that 
$\s \in \partial \Omega(\e_1)$ for 
any $\s$ such that $\ss_1 = m$ and $\ss_i \le 0$ for all $i\ge 2$; and more generally we have $\s + c\mathbf{1} \in \partial \Omega(\e_1)$. 
That is, any $\s$ with $\ss_1 \ge m + \max_{i\ne 1} \ss_i$ satisfies 
$\s \in \partial \Omega(\e_1)$. From Proposition~\ref{prop:fy_losses}, 
this is equivalent to $L_{\Omega}(\s;\e_1) = 0$. 

Let us now determine the margin of $L_{\Omega}$, i.e., the 
smallest $m$ such that 
$m\e_1 \in \partial \Omega(\e_1)$.  
From Lemma~\ref{prop:inverse_map}, 
this is equivalent to 
$m \ge m \pp_1 - \Omega(\p)$ for any $\p \in \triangle^d$, i.e., 
${-\Omega(\p)}{(1-\pp_1)} \le m$. 
Note that by Proposition~\ref{prop:prediction_func} the ``most competitive'' $\p$'s are sorted as $\e_1$, so we may write $\pp_1 = \|\p\|_\infty$ without loss of generality. The margin of $L_{\Omega}$ is the smallest possible such margin, given by \eqref{eq:margin}.

\subsection{Proof of Proposition~\ref{prop:full_simplex}}\label{appendix:proof_full_simplex}

Let us start by showing that conditions 1 and 2 are equivalent. 
To show that 2 $\Rightarrow$ 1, take an arbitrary $\p \in \triangle^d$. From Fenchel-Young duality and the Danskin's theorem, we have that $\nabla(-\HH)^*(\s) = \p \,\, \Rightarrow \,\, \s \in \partial(-\HH)(\p)$, which implies the subdifferential set is non-empty everywhere in the simplex. Let us now prove that 1 $\Rightarrow$ 2. Let $\Omega = -{\HH}$, and assume that $\Omega$ has non-empty subdifferential everywhere in $\triangle^d$.  
We need to show that for any $\p \in \triangle^d$, there is some $\s \in \RR^d$ such that 
$\p \in \argmin_{\p' \in \triangle^d} \Omega(\p') - \DP{\s}{\p'}$. 
The Lagrangian associated with this minimization problem is:
\begin{equation}
\mathcal{L}(\p, \bm{\mu}, \lambda) = \Omega(\p) - \DP{\s + \bm{\mu}}{\p} + \lambda (\mathbf{1}^\top \p - 1).
\end{equation}
The KKT conditions are:
\begin{eqnarray}
\left\{
\begin{array}{l}
0 \in \partial_p \mathcal{L}(\p, \bm{\mu}, \lambda) = 
\partial \Omega(\p) - \s - \bm{\mu} + \lambda \mathbf{1}\\
\DP{\p}{\bm{\mu}} = 0\\
\p \in \triangle^d, \,\, \bm{\mu} \ge 0.
\end{array}
\right.
\end{eqnarray}
For a given $\p \in \triangle^d$, we seek $\s$ such that 
$(\p, \bm{\mu}, \lambda)$ are a solution to the KKT conditions 
for some $\bm{\mu} \ge 0$ and $\lambda \in \RR$. 

We will show that such $\s$ exists by simply choosing 
$\bm{\mu}=\mathbf{0}$ and $\lambda=0$. 
Those choices are dual feasible and guarantee that the slackness complementary condition is satisfied. 
In this case, we have from the first condition that 
$\s \in \partial \Omega(\p)$. 
From the assumption that $\Omega$ has non-empty subdifferential in all the simplex, we have that for any $\p \in \triangle^d$ we can find a 
$\s \in \RR^d$ such that $(\p, \s)$ are a dual pair, i.e., $\p = \nabla \Omega^*(\s)$, which proves that $\nabla \Omega^*(\RR^d) = \triangle^d$.  

Next, we show that condition $1 \Rightarrow 3$. Since $\partial (-\HH)(\p) \ne \varnothing$ everywhere in the simplex, we can take an arbitrary $\s \in \partial (-\HH)(\e_k)$. From  Lemma~\ref{prop:inverse_map}, item 2, we have that $\s' \in \partial (-\HH)(\e_k)$ for $\ss'_k = \ss_k$ and $\ss'_j = \min_\ell \ss_\ell$; since $(-\HH)^*$ is shift invariant, we can without loss of generality have $\ss' = m\e_k$ for some $m>0$, which implies from Proposition~\ref{prop:margin} that $L_\Omega$ has a margin. 

Let us show that, if $-\HH$ is separable, then $3 \Rightarrow 1$, which establishes equivalence between all conditions 1, 2, and 3. 
From Proposition~\ref{prop:margin}, the existing of a separation margin implies that there is some $m$ such that $m\e_k \in \partial (-\HH)(\e_k)$. 
Let $\HH(\p) = \sum_{i=1}^d h(p_i)$, with $h:[0,1]\rightarrow \mathbb{R}_+$ concave. Due to assumption A.1, $h$ must satisfy $h(0)=h(1)=0$. 
Without loss of generality, suppose $\p = [\tilde{\p}; \mathbf{0}_k]$, where  $\tilde{\p} \in \relint(\triangle^{d-k})$ and $\mathbf{0}_k$ is a vector with $k$ zeros. 
We will see that there is a vector $\bm{g} \in \RR^d$ such that 
$\bm{g} \in \partial (-\HH)(\p)$, i.e., satisfying 
\begin{equation}
-\HH(\p') \ge -\HH(\p) + \langle \bm{g}, \p'-\p \rangle, \quad \forall \p' \in \triangle^d. 
\label{eq:subgrad_ineq}
\end{equation}
Since $\tilde{\p} \in \relint(\triangle^{d-k})$, we have $\tilde{p}_i \in ]0,1[$
for $i \in \{1,\ldots,d-k\}$, hence $\partial (-h) (\tilde{p}_i)$ must be
nonempty, since $-h$ is convex and $]0,1[$ is an open set. We show that the following $\bm{g} = (g_1, \ldots, g_d) \in \RR^d$ is a subgradient of $-\HH$ at $\p$:
\begin{equation*}
g_i = \left\{
\begin{array}{ll}
\partial (-h) (\tilde{p}_i), & \text{$i=1,\ldots,d-k$}\\
m, &  \text{$i=d-k+1,\ldots,d$}.
\end{array}
\right.
\end{equation*}
By definition of subgradient, we have 
\begin{equation}
-\psi(p_i') \ge -\psi(\tilde{p}_i) + \partial (-h) (\tilde{p}_i)(p_i'-\tilde{p}_i), \quad \text{for $i=1,\ldots,d-k$}.
\label{eq:subgrad_ineq1}
\end{equation}
Furthermore, since $m$ upper bounds the separation margin of $\HH$, we have from Proposition~\ref{prop:margin} that 
$m \ge \frac{\HH([1-p_i', p_i', 0, \ldots, 0])}{1 - \max\{1-p_i', p_i'\}} =
\frac{h(1-p_i') + h(p_i')}{\min\{p_i', 1-p_i'\}} \ge \frac{h(p_i')}{p_i'}$ for any $p_i' \in ]0,1]$. Hence, we have 
\begin{equation}
-\psi(p_i') \ge -\psi(0) - m(p_i' - 0), \quad \text{for $i=d-k+1,\ldots,d$}.\label{eq:subgrad_ineq2}
\end{equation}
Summing all inequalities in Eqs.~\eqref{eq:subgrad_ineq1}--\eqref{eq:subgrad_ineq2}, we obtain the expression in Eq.~\eqref{eq:subgrad_ineq}, which finishes the proof.

\subsection{Proof of Proposition~\ref{prop:margin_separable}}\label{appendix:proof_margin_separable}

Define $\Omega = -\HH$.
Let us start by writing the margin expression \eqref{eq:margin} as a unidimensional optimization problem. This is done by noticing that the max-generalized entropy problem constrained to $\max(\p)=1-t$ gives $\p=\left[1-t, \frac{t}{d-1}, \ldots, \frac{t}{d-1}\right]$, for $t \in \left[0, 1-\frac{1}{d}\right]$ by a similar argument as the one used in Proposition~\ref{prop:generalized_entropy}. We obtain:
\begin{equation}
\mathrm{margin}(L_{\Omega}) = \sup_{t \in \left[0, 1-\frac{1}{d}\right]} \frac{-\Omega\left(\left[1-t, \frac{t}{d-1}, \ldots, \frac{t}{d-1}\right]\right)}{t}.
\end{equation}

We write the argument above as $A(t) = \frac{-\Omega(\e_1 + t\bm{v})}{t}$, where $\bm{v} := [-1, \frac{1}{d-1}, \ldots, \frac{1}{d-1}]$. 
We will first prove that $A$ is decreasing in $[0, 1-\frac{1}{d}]$, which implies that the supremum (and the margin) equals $A(0)$.  
Note that we have the following expression for the derivative of any function $f(\e_1 + t\bm{v})$:
\begin{equation}
(f(\e_1 + t\bm{v}))' = \bm{v}^\top \nabla f(\e_1 + t\bm{v}).
\end{equation}
Using this fact, we can write the derivative $A'(t)$ as:
\begin{equation}
A'(t) = \frac{-t\bm{v}^\top \nabla \Omega(\e_1 + t\bm{v}) + \Omega(\e_1 + t\bm{v})}{t^2} := \frac{B(t)}{t^2}.
\end{equation}
In turn, the derivative $B'(t)$ is:
\begin{eqnarray}
B'(t) &=& -\bm{v}^\top \nabla \Omega(\e_1 + t\bm{v}) -t(\bm{v}^\top\nabla \Omega(\e_1 + t\bm{v}))' + \bm{v}^\top \nabla \Omega(\e_1 + t\bm{v}) \nonumber\\
&=& -t(\bm{v}^\top\nabla \Omega(\e_1 + t\bm{v}))' \nonumber\\
&=& -t\bm{v}^\top\nabla\nabla \Omega(\e_1 + t\bm{v})\bm{v} \nonumber\\
&\le& 0,
\end{eqnarray}
where we denote by $\nabla\nabla \Omega$ the Hessian of $\Omega$, 
and used the fact that it is positive semi-definite, due to the convexity of $\Omega$. 
This implies that $B$ is decreasing, hence for any $t \in [0,1]$, $B(t) \le B(0) = \Omega(\e_1) = 0$, where we used the fact 
$\|\nabla \Omega(\e_1)\| < \infty$, assumed as a condition of Proposition~\ref{prop:full_simplex}. 
Therefore, we must also have $A'(t) = \frac{B(t)}{t^2} \le 0$ for any $t \in [0,1]$, 
hence $A$ is decreasing, and $\sup_{t \in [0, 1-1/d]} A(t) = \lim_{t\rightarrow 0+} A(t)$. 
By L'H\^opital's rule:
\begin{eqnarray}
\lim_{t\rightarrow 0+} A(t) &=& \lim_{t\rightarrow 0+} (-\Omega(\e_1 + t\bm{v}))' \nonumber\\
&=& -\bm{v}^\top \nabla \Omega(\e_1) \nonumber\\
&=& {\nabla_1 \Omega(\e_1)} - \frac{1}{d-1}\sum_{j\ge 2}{\nabla_j \Omega(\e_1)}\nonumber\\ 
&=& {\nabla_1 \Omega(\e_1)} - {\nabla_2 \Omega(\e_1)},
\end{eqnarray} 
which proves the first part. 

If $\Omega$ is separable, then 
$\nabla_j \Omega(\p) = -h'(\pp_j)$, in particular
${\nabla_1\Omega(\e_1)} = -h'(1)$ and 
${\nabla_2\Omega(\e_1)} = -h'(0)$, yielding 
$\mathrm{margin}(L_{\Omega}) = h'(0) - h'(1)$. 
Since $h$ is twice differentiable, this equals $-\int_{0}^{1} h''(t)dt$, completing the proof. 

\end{document}

%% file: figures/bisection_time_to_acc.pgf
%% Creator: Matplotlib, PGF backend
%%
%% To include the figure in your LaTeX document, write
%%   \input{<filename>.pgf}
%%
%% Make sure the required packages are loaded in your preamble
%%   \usepackage{pgf}
%%
%% Figures using additional raster images can only be included by \input if
%% they are in the same directory as the main LaTeX file. For loading figures
%% from other directories you can use the `import` package
%% and then include the figures with
%%   \import{<path to file>}{<filename>.pgf}
%%
%% Matplotlib used the following preamble
%%   \usepackage{fontspec}
%%   \setmainfont{DejaVuSerif.ttf}[Path=/home/vlad/conda/envs/main/lib/python3.7/site-packages/matplotlib/mpl-data/fonts/ttf/]
%%   \setsansfont{DejaVuSans.ttf}[Path=/home/vlad/conda/envs/main/lib/python3.7/site-packages/matplotlib/mpl-data/fonts/ttf/]
%%   \setmonofont{DejaVuSansMono.ttf}[Path=/home/vlad/conda/envs/main/lib/python3.7/site-packages/matplotlib/mpl-data/fonts/ttf/]
%%
\begingroup%
\makeatletter%
\begin{pgfpicture}%
\pgfpathrectangle{\pgfpointorigin}{\pgfqpoint{1.800000in}{1.600000in}}%
\pgfusepath{use as bounding box, clip}%
\begin{pgfscope}%
\pgfsetbuttcap%
\pgfsetmiterjoin%
\definecolor{currentfill}{rgb}{1.000000,1.000000,1.000000}%
\pgfsetfillcolor{currentfill}%
\pgfsetlinewidth{0.000000pt}%
\definecolor{currentstroke}{rgb}{1.000000,1.000000,1.000000}%
\pgfsetstrokecolor{currentstroke}%
\pgfsetdash{}{0pt}%
\pgfpathmoveto{\pgfqpoint{0.000000in}{0.000000in}}%
\pgfpathlineto{\pgfqpoint{1.800000in}{0.000000in}}%
\pgfpathlineto{\pgfqpoint{1.800000in}{1.600000in}}%
\pgfpathlineto{\pgfqpoint{0.000000in}{1.600000in}}%
\pgfpathclose%
\pgfusepath{fill}%
\end{pgfscope}%
\begin{pgfscope}%
\pgfsetbuttcap%
\pgfsetmiterjoin%
\definecolor{currentfill}{rgb}{1.000000,1.000000,1.000000}%
\pgfsetfillcolor{currentfill}%
\pgfsetlinewidth{0.000000pt}%
\definecolor{currentstroke}{rgb}{0.000000,0.000000,0.000000}%
\pgfsetstrokecolor{currentstroke}%
\pgfsetstrokeopacity{0.000000}%
\pgfsetdash{}{0pt}%
\pgfpathmoveto{\pgfqpoint{0.422284in}{0.259864in}}%
\pgfpathlineto{\pgfqpoint{1.723330in}{0.259864in}}%
\pgfpathlineto{\pgfqpoint{1.723330in}{1.523330in}}%
\pgfpathlineto{\pgfqpoint{0.422284in}{1.523330in}}%
\pgfpathclose%
\pgfusepath{fill}%
\end{pgfscope}%
\begin{pgfscope}%
\pgfpathrectangle{\pgfqpoint{0.422284in}{0.259864in}}{\pgfqpoint{1.301046in}{1.263466in}}%
\pgfusepath{clip}%
\pgfsetbuttcap%
\pgfsetmiterjoin%
\definecolor{currentfill}{rgb}{0.121569,0.466667,0.705882}%
\pgfsetfillcolor{currentfill}%
\pgfsetfillopacity{0.500000}%
\pgfsetlinewidth{0.501875pt}%
\definecolor{currentstroke}{rgb}{0.376471,0.376471,0.376471}%
\pgfsetstrokecolor{currentstroke}%
\pgfsetdash{}{0pt}%
\pgfpathmoveto{\pgfqpoint{0.481423in}{0in}}%
\pgfpathlineto{\pgfqpoint{0.588947in}{0in}}%
\pgfpathlineto{\pgfqpoint{0.588947in}{0.388524in}}%
\pgfpathlineto{\pgfqpoint{0.481423in}{0.388524in}}%
\pgfpathclose%
\pgfusepath{stroke,fill}%
\end{pgfscope}%
\begin{pgfscope}%
\pgfpathrectangle{\pgfqpoint{0.422284in}{0.259864in}}{\pgfqpoint{1.301046in}{1.263466in}}%
\pgfusepath{clip}%
\pgfsetbuttcap%
\pgfsetmiterjoin%
\definecolor{currentfill}{rgb}{0.121569,0.466667,0.705882}%
\pgfsetfillcolor{currentfill}%
\pgfsetfillopacity{0.500000}%
\pgfsetlinewidth{0.501875pt}%
\definecolor{currentstroke}{rgb}{0.376471,0.376471,0.376471}%
\pgfsetstrokecolor{currentstroke}%
\pgfsetdash{}{0pt}%
\pgfpathmoveto{\pgfqpoint{0.911520in}{0in}}%
\pgfpathlineto{\pgfqpoint{1.019045in}{0in}}%
\pgfpathlineto{\pgfqpoint{1.019045in}{0.494933in}}%
\pgfpathlineto{\pgfqpoint{0.911520in}{0.494933in}}%
\pgfpathclose%
\pgfusepath{stroke,fill}%
\end{pgfscope}%
\begin{pgfscope}%
\pgfpathrectangle{\pgfqpoint{0.422284in}{0.259864in}}{\pgfqpoint{1.301046in}{1.263466in}}%
\pgfusepath{clip}%
\pgfsetbuttcap%
\pgfsetmiterjoin%
\definecolor{currentfill}{rgb}{0.121569,0.466667,0.705882}%
\pgfsetfillcolor{currentfill}%
\pgfsetfillopacity{0.500000}%
\pgfsetlinewidth{0.501875pt}%
\definecolor{currentstroke}{rgb}{0.376471,0.376471,0.376471}%
\pgfsetstrokecolor{currentstroke}%
\pgfsetdash{}{0pt}%
\pgfpathmoveto{\pgfqpoint{1.341618in}{0in}}%
\pgfpathlineto{\pgfqpoint{1.449143in}{0in}}%
\pgfpathlineto{\pgfqpoint{1.449143in}{0.698057in}}%
\pgfpathlineto{\pgfqpoint{1.341618in}{0.698057in}}%
\pgfpathclose%
\pgfusepath{stroke,fill}%
\end{pgfscope}%
\begin{pgfscope}%
\pgfpathrectangle{\pgfqpoint{0.422284in}{0.259864in}}{\pgfqpoint{1.301046in}{1.263466in}}%
\pgfusepath{clip}%
\pgfsetbuttcap%
\pgfsetmiterjoin%
\definecolor{currentfill}{rgb}{1.000000,0.498039,0.054902}%
\pgfsetfillcolor{currentfill}%
\pgfsetfillopacity{0.500000}%
\pgfsetlinewidth{0.501875pt}%
\definecolor{currentstroke}{rgb}{0.376471,0.376471,0.376471}%
\pgfsetstrokecolor{currentstroke}%
\pgfsetdash{}{0pt}%
\pgfpathmoveto{\pgfqpoint{0.588947in}{0in}}%
\pgfpathlineto{\pgfqpoint{0.696472in}{0in}}%
\pgfpathlineto{\pgfqpoint{0.696472in}{0.401500in}}%
\pgfpathlineto{\pgfqpoint{0.588947in}{0.401500in}}%
\pgfpathclose%
\pgfusepath{stroke,fill}%
\end{pgfscope}%
\begin{pgfscope}%
\pgfpathrectangle{\pgfqpoint{0.422284in}{0.259864in}}{\pgfqpoint{1.301046in}{1.263466in}}%
\pgfusepath{clip}%
\pgfsetbuttcap%
\pgfsetmiterjoin%
\definecolor{currentfill}{rgb}{1.000000,0.498039,0.054902}%
\pgfsetfillcolor{currentfill}%
\pgfsetfillopacity{0.500000}%
\pgfsetlinewidth{0.501875pt}%
\definecolor{currentstroke}{rgb}{0.376471,0.376471,0.376471}%
\pgfsetstrokecolor{currentstroke}%
\pgfsetdash{}{0pt}%
\pgfpathmoveto{\pgfqpoint{1.019045in}{0in}}%
\pgfpathlineto{\pgfqpoint{1.126569in}{0in}}%
\pgfpathlineto{\pgfqpoint{1.126569in}{0.501535in}}%
\pgfpathlineto{\pgfqpoint{1.019045in}{0.501535in}}%
\pgfpathclose%
\pgfusepath{stroke,fill}%
\end{pgfscope}%
\begin{pgfscope}%
\pgfpathrectangle{\pgfqpoint{0.422284in}{0.259864in}}{\pgfqpoint{1.301046in}{1.263466in}}%
\pgfusepath{clip}%
\pgfsetbuttcap%
\pgfsetmiterjoin%
\definecolor{currentfill}{rgb}{1.000000,0.498039,0.054902}%
\pgfsetfillcolor{currentfill}%
\pgfsetfillopacity{0.500000}%
\pgfsetlinewidth{0.501875pt}%
\definecolor{currentstroke}{rgb}{0.376471,0.376471,0.376471}%
\pgfsetstrokecolor{currentstroke}%
\pgfsetdash{}{0pt}%
\pgfpathmoveto{\pgfqpoint{1.449143in}{0in}}%
\pgfpathlineto{\pgfqpoint{1.556667in}{0in}}%
\pgfpathlineto{\pgfqpoint{1.556667in}{0.856666in}}%
\pgfpathlineto{\pgfqpoint{1.449143in}{0.856666in}}%
\pgfpathclose%
\pgfusepath{stroke,fill}%
\end{pgfscope}%
\begin{pgfscope}%
\pgfpathrectangle{\pgfqpoint{0.422284in}{0.259864in}}{\pgfqpoint{1.301046in}{1.263466in}}%
\pgfusepath{clip}%
\pgfsetbuttcap%
\pgfsetmiterjoin%
\definecolor{currentfill}{rgb}{0.839216,0.152941,0.156863}%
\pgfsetfillcolor{currentfill}%
\pgfsetfillopacity{0.500000}%
\pgfsetlinewidth{0.501875pt}%
\definecolor{currentstroke}{rgb}{0.376471,0.376471,0.376471}%
\pgfsetstrokecolor{currentstroke}%
\pgfsetdash{}{0pt}%
\pgfpathmoveto{\pgfqpoint{0.696472in}{0in}}%
\pgfpathlineto{\pgfqpoint{0.803996in}{0in}}%
\pgfpathlineto{\pgfqpoint{0.803996in}{0.464107in}}%
\pgfpathlineto{\pgfqpoint{0.696472in}{0.464107in}}%
\pgfpathclose%
\pgfusepath{stroke,fill}%
\end{pgfscope}%
\begin{pgfscope}%
\pgfpathrectangle{\pgfqpoint{0.422284in}{0.259864in}}{\pgfqpoint{1.301046in}{1.263466in}}%
\pgfusepath{clip}%
\pgfsetbuttcap%
\pgfsetmiterjoin%
\definecolor{currentfill}{rgb}{0.839216,0.152941,0.156863}%
\pgfsetfillcolor{currentfill}%
\pgfsetfillopacity{0.500000}%
\pgfsetlinewidth{0.501875pt}%
\definecolor{currentstroke}{rgb}{0.376471,0.376471,0.376471}%
\pgfsetstrokecolor{currentstroke}%
\pgfsetdash{}{0pt}%
\pgfpathmoveto{\pgfqpoint{1.126569in}{0in}}%
\pgfpathlineto{\pgfqpoint{1.234094in}{0in}}%
\pgfpathlineto{\pgfqpoint{1.234094in}{0.558369in}}%
\pgfpathlineto{\pgfqpoint{1.126569in}{0.558369in}}%
\pgfpathclose%
\pgfusepath{stroke,fill}%
\end{pgfscope}%
\begin{pgfscope}%
\pgfpathrectangle{\pgfqpoint{0.422284in}{0.259864in}}{\pgfqpoint{1.301046in}{1.263466in}}%
\pgfusepath{clip}%
\pgfsetbuttcap%
\pgfsetmiterjoin%
\definecolor{currentfill}{rgb}{0.839216,0.152941,0.156863}%
\pgfsetfillcolor{currentfill}%
\pgfsetfillopacity{0.500000}%
\pgfsetlinewidth{0.501875pt}%
\definecolor{currentstroke}{rgb}{0.376471,0.376471,0.376471}%
\pgfsetstrokecolor{currentstroke}%
\pgfsetdash{}{0pt}%
\pgfpathmoveto{\pgfqpoint{1.556667in}{0in}}%
\pgfpathlineto{\pgfqpoint{1.664192in}{0in}}%
\pgfpathlineto{\pgfqpoint{1.664192in}{1.255610in}}%
\pgfpathlineto{\pgfqpoint{1.556667in}{1.255610in}}%
\pgfpathclose%
\pgfusepath{stroke,fill}%
\end{pgfscope}%
\begin{pgfscope}%
\pgfsetbuttcap%
\pgfsetroundjoin%
\definecolor{currentfill}{rgb}{0.000000,0.000000,0.000000}%
\pgfsetfillcolor{currentfill}%
\pgfsetlinewidth{0.803000pt}%
\definecolor{currentstroke}{rgb}{0.000000,0.000000,0.000000}%
\pgfsetstrokecolor{currentstroke}%
\pgfsetdash{}{0pt}%
\pgfsys@defobject{currentmarker}{\pgfqpoint{0.000000in}{-0.048611in}}{\pgfqpoint{0.000000in}{0.000000in}}{%
\pgfpathmoveto{\pgfqpoint{0.000000in}{0.000000in}}%
\pgfpathlineto{\pgfqpoint{0.000000in}{-0.048611in}}%
\pgfusepath{stroke,fill}%
}%
\begin{pgfscope}%
\pgfsys@transformshift{0.642709in}{0.259864in}%
\pgfsys@useobject{currentmarker}{}%
\end{pgfscope}%
\end{pgfscope}%
\begin{pgfscope}%
\definecolor{textcolor}{rgb}{0.000000,0.000000,0.000000}%
\pgfsetstrokecolor{textcolor}%
\pgfsetfillcolor{textcolor}%
\pgftext[x=0.642709in,y=0.162642in,,top]{\color{textcolor}\rmfamily\fontsize{9.000000}{10.800000}\selectfont Brent}%
\end{pgfscope}%
\begin{pgfscope}%
\pgfsetbuttcap%
\pgfsetroundjoin%
\definecolor{currentfill}{rgb}{0.000000,0.000000,0.000000}%
\pgfsetfillcolor{currentfill}%
\pgfsetlinewidth{0.803000pt}%
\definecolor{currentstroke}{rgb}{0.000000,0.000000,0.000000}%
\pgfsetstrokecolor{currentstroke}%
\pgfsetdash{}{0pt}%
\pgfsys@defobject{currentmarker}{\pgfqpoint{0.000000in}{-0.048611in}}{\pgfqpoint{0.000000in}{0.000000in}}{%
\pgfpathmoveto{\pgfqpoint{0.000000in}{0.000000in}}%
\pgfpathlineto{\pgfqpoint{0.000000in}{-0.048611in}}%
\pgfusepath{stroke,fill}%
}%
\begin{pgfscope}%
\pgfsys@transformshift{1.072807in}{0.259864in}%
\pgfsys@useobject{currentmarker}{}%
\end{pgfscope}%
\end{pgfscope}%
\begin{pgfscope}%
\definecolor{textcolor}{rgb}{0.000000,0.000000,0.000000}%
\pgfsetstrokecolor{textcolor}%
\pgfsetfillcolor{textcolor}%
\pgftext[x=1.072807in,y=0.162642in,,top]{\color{textcolor}\rmfamily\fontsize{9.000000}{10.800000}\selectfont Bisect}%
\end{pgfscope}%
\begin{pgfscope}%
\pgfsetbuttcap%
\pgfsetroundjoin%
\definecolor{currentfill}{rgb}{0.000000,0.000000,0.000000}%
\pgfsetfillcolor{currentfill}%
\pgfsetlinewidth{0.803000pt}%
\definecolor{currentstroke}{rgb}{0.000000,0.000000,0.000000}%
\pgfsetstrokecolor{currentstroke}%
\pgfsetdash{}{0pt}%
\pgfsys@defobject{currentmarker}{\pgfqpoint{0.000000in}{-0.048611in}}{\pgfqpoint{0.000000in}{0.000000in}}{%
\pgfpathmoveto{\pgfqpoint{0.000000in}{0.000000in}}%
\pgfpathlineto{\pgfqpoint{0.000000in}{-0.048611in}}%
\pgfusepath{stroke,fill}%
}%
\begin{pgfscope}%
\pgfsys@transformshift{1.502905in}{0.259864in}%
\pgfsys@useobject{currentmarker}{}%
\end{pgfscope}%
\end{pgfscope}%
\begin{pgfscope}%
\definecolor{textcolor}{rgb}{0.000000,0.000000,0.000000}%
\pgfsetstrokecolor{textcolor}%
\pgfsetfillcolor{textcolor}%
\pgftext[x=1.502905in,y=0.162642in,,top]{\color{textcolor}\rmfamily\fontsize{9.000000}{10.800000}\selectfont FISTA}%
\end{pgfscope}%
\begin{pgfscope}%
\pgfsetbuttcap%
\pgfsetroundjoin%
\definecolor{currentfill}{rgb}{0.000000,0.000000,0.000000}%
\pgfsetfillcolor{currentfill}%
\pgfsetlinewidth{0.803000pt}%
\definecolor{currentstroke}{rgb}{0.000000,0.000000,0.000000}%
\pgfsetstrokecolor{currentstroke}%
\pgfsetdash{}{0pt}%
\pgfsys@defobject{currentmarker}{\pgfqpoint{-0.048611in}{0.000000in}}{\pgfqpoint{0.000000in}{0.000000in}}{%
\pgfpathmoveto{\pgfqpoint{0.000000in}{0.000000in}}%
\pgfpathlineto{\pgfqpoint{-0.048611in}{0.000000in}}%
\pgfusepath{stroke,fill}%
}%
\begin{pgfscope}%
\pgfsys@transformshift{0.422284in}{0.400541in}%
\pgfsys@useobject{currentmarker}{}%
\end{pgfscope}%
\end{pgfscope}%
\begin{pgfscope}%
\definecolor{textcolor}{rgb}{0.000000,0.000000,0.000000}%
\pgfsetstrokecolor{textcolor}%
\pgfsetfillcolor{textcolor}%
\pgftext[x=0.093197in,y=0.353055in,left,base]{\color{textcolor}\rmfamily\fontsize{9.000000}{10.800000}\selectfont \(\displaystyle 10^{-1}\)}%
\end{pgfscope}%
\begin{pgfscope}%
\pgfsetbuttcap%
\pgfsetroundjoin%
\definecolor{currentfill}{rgb}{0.000000,0.000000,0.000000}%
\pgfsetfillcolor{currentfill}%
\pgfsetlinewidth{0.803000pt}%
\definecolor{currentstroke}{rgb}{0.000000,0.000000,0.000000}%
\pgfsetstrokecolor{currentstroke}%
\pgfsetdash{}{0pt}%
\pgfsys@defobject{currentmarker}{\pgfqpoint{-0.048611in}{0.000000in}}{\pgfqpoint{0.000000in}{0.000000in}}{%
\pgfpathmoveto{\pgfqpoint{0.000000in}{0.000000in}}%
\pgfpathlineto{\pgfqpoint{-0.048611in}{0.000000in}}%
\pgfusepath{stroke,fill}%
}%
\begin{pgfscope}%
\pgfsys@transformshift{0.422284in}{0.754054in}%
\pgfsys@useobject{currentmarker}{}%
\end{pgfscope}%
\end{pgfscope}%
\begin{pgfscope}%
\definecolor{textcolor}{rgb}{0.000000,0.000000,0.000000}%
\pgfsetstrokecolor{textcolor}%
\pgfsetfillcolor{textcolor}%
\pgftext[x=0.173443in,y=0.706568in,left,base]{\color{textcolor}\rmfamily\fontsize{9.000000}{10.800000}\selectfont \(\displaystyle 10^{0}\)}%
\end{pgfscope}%
\begin{pgfscope}%
\pgfsetbuttcap%
\pgfsetroundjoin%
\definecolor{currentfill}{rgb}{0.000000,0.000000,0.000000}%
\pgfsetfillcolor{currentfill}%
\pgfsetlinewidth{0.803000pt}%
\definecolor{currentstroke}{rgb}{0.000000,0.000000,0.000000}%
\pgfsetstrokecolor{currentstroke}%
\pgfsetdash{}{0pt}%
\pgfsys@defobject{currentmarker}{\pgfqpoint{-0.048611in}{0.000000in}}{\pgfqpoint{0.000000in}{0.000000in}}{%
\pgfpathmoveto{\pgfqpoint{0.000000in}{0.000000in}}%
\pgfpathlineto{\pgfqpoint{-0.048611in}{0.000000in}}%
\pgfusepath{stroke,fill}%
}%
\begin{pgfscope}%
\pgfsys@transformshift{0.422284in}{1.107567in}%
\pgfsys@useobject{currentmarker}{}%
\end{pgfscope}%
\end{pgfscope}%
\begin{pgfscope}%
\definecolor{textcolor}{rgb}{0.000000,0.000000,0.000000}%
\pgfsetstrokecolor{textcolor}%
\pgfsetfillcolor{textcolor}%
\pgftext[x=0.173443in,y=1.060081in,left,base]{\color{textcolor}\rmfamily\fontsize{9.000000}{10.800000}\selectfont \(\displaystyle 10^{1}\)}%
\end{pgfscope}%
\begin{pgfscope}%
\pgfsetbuttcap%
\pgfsetroundjoin%
\definecolor{currentfill}{rgb}{0.000000,0.000000,0.000000}%
\pgfsetfillcolor{currentfill}%
\pgfsetlinewidth{0.803000pt}%
\definecolor{currentstroke}{rgb}{0.000000,0.000000,0.000000}%
\pgfsetstrokecolor{currentstroke}%
\pgfsetdash{}{0pt}%
\pgfsys@defobject{currentmarker}{\pgfqpoint{-0.048611in}{0.000000in}}{\pgfqpoint{0.000000in}{0.000000in}}{%
\pgfpathmoveto{\pgfqpoint{0.000000in}{0.000000in}}%
\pgfpathlineto{\pgfqpoint{-0.048611in}{0.000000in}}%
\pgfusepath{stroke,fill}%
}%
\begin{pgfscope}%
\pgfsys@transformshift{0.422284in}{1.461079in}%
\pgfsys@useobject{currentmarker}{}%
\end{pgfscope}%
\end{pgfscope}%
\begin{pgfscope}%
\definecolor{textcolor}{rgb}{0.000000,0.000000,0.000000}%
\pgfsetstrokecolor{textcolor}%
\pgfsetfillcolor{textcolor}%
\pgftext[x=0.173443in,y=1.413594in,left,base]{\color{textcolor}\rmfamily\fontsize{9.000000}{10.800000}\selectfont \(\displaystyle 10^{2}\)}%
\end{pgfscope}%
\begin{pgfscope}%
\pgfsetbuttcap%
\pgfsetroundjoin%
\definecolor{currentfill}{rgb}{0.000000,0.000000,0.000000}%
\pgfsetfillcolor{currentfill}%
\pgfsetlinewidth{0.602250pt}%
\definecolor{currentstroke}{rgb}{0.000000,0.000000,0.000000}%
\pgfsetstrokecolor{currentstroke}%
\pgfsetdash{}{0pt}%
\pgfsys@defobject{currentmarker}{\pgfqpoint{-0.027778in}{0.000000in}}{\pgfqpoint{0.000000in}{0.000000in}}{%
\pgfpathmoveto{\pgfqpoint{0.000000in}{0.000000in}}%
\pgfpathlineto{\pgfqpoint{-0.027778in}{0.000000in}}%
\pgfusepath{stroke,fill}%
}%
\begin{pgfscope}%
\pgfsys@transformshift{0.422284in}{0.259864in}%
\pgfsys@useobject{currentmarker}{}%
\end{pgfscope}%
\end{pgfscope}%
\begin{pgfscope}%
\pgfsetbuttcap%
\pgfsetroundjoin%
\definecolor{currentfill}{rgb}{0.000000,0.000000,0.000000}%
\pgfsetfillcolor{currentfill}%
\pgfsetlinewidth{0.602250pt}%
\definecolor{currentstroke}{rgb}{0.000000,0.000000,0.000000}%
\pgfsetstrokecolor{currentstroke}%
\pgfsetdash{}{0pt}%
\pgfsys@defobject{currentmarker}{\pgfqpoint{-0.027778in}{0.000000in}}{\pgfqpoint{0.000000in}{0.000000in}}{%
\pgfpathmoveto{\pgfqpoint{0.000000in}{0.000000in}}%
\pgfpathlineto{\pgfqpoint{-0.027778in}{0.000000in}}%
\pgfusepath{stroke,fill}%
}%
\begin{pgfscope}%
\pgfsys@transformshift{0.422284in}{0.294123in}%
\pgfsys@useobject{currentmarker}{}%
\end{pgfscope}%
\end{pgfscope}%
\begin{pgfscope}%
\pgfsetbuttcap%
\pgfsetroundjoin%
\definecolor{currentfill}{rgb}{0.000000,0.000000,0.000000}%
\pgfsetfillcolor{currentfill}%
\pgfsetlinewidth{0.602250pt}%
\definecolor{currentstroke}{rgb}{0.000000,0.000000,0.000000}%
\pgfsetstrokecolor{currentstroke}%
\pgfsetdash{}{0pt}%
\pgfsys@defobject{currentmarker}{\pgfqpoint{-0.027778in}{0.000000in}}{\pgfqpoint{0.000000in}{0.000000in}}{%
\pgfpathmoveto{\pgfqpoint{0.000000in}{0.000000in}}%
\pgfpathlineto{\pgfqpoint{-0.027778in}{0.000000in}}%
\pgfusepath{stroke,fill}%
}%
\begin{pgfscope}%
\pgfsys@transformshift{0.422284in}{0.322114in}%
\pgfsys@useobject{currentmarker}{}%
\end{pgfscope}%
\end{pgfscope}%
\begin{pgfscope}%
\pgfsetbuttcap%
\pgfsetroundjoin%
\definecolor{currentfill}{rgb}{0.000000,0.000000,0.000000}%
\pgfsetfillcolor{currentfill}%
\pgfsetlinewidth{0.602250pt}%
\definecolor{currentstroke}{rgb}{0.000000,0.000000,0.000000}%
\pgfsetstrokecolor{currentstroke}%
\pgfsetdash{}{0pt}%
\pgfsys@defobject{currentmarker}{\pgfqpoint{-0.027778in}{0.000000in}}{\pgfqpoint{0.000000in}{0.000000in}}{%
\pgfpathmoveto{\pgfqpoint{0.000000in}{0.000000in}}%
\pgfpathlineto{\pgfqpoint{-0.027778in}{0.000000in}}%
\pgfusepath{stroke,fill}%
}%
\begin{pgfscope}%
\pgfsys@transformshift{0.422284in}{0.345781in}%
\pgfsys@useobject{currentmarker}{}%
\end{pgfscope}%
\end{pgfscope}%
\begin{pgfscope}%
\pgfsetbuttcap%
\pgfsetroundjoin%
\definecolor{currentfill}{rgb}{0.000000,0.000000,0.000000}%
\pgfsetfillcolor{currentfill}%
\pgfsetlinewidth{0.602250pt}%
\definecolor{currentstroke}{rgb}{0.000000,0.000000,0.000000}%
\pgfsetstrokecolor{currentstroke}%
\pgfsetdash{}{0pt}%
\pgfsys@defobject{currentmarker}{\pgfqpoint{-0.027778in}{0.000000in}}{\pgfqpoint{0.000000in}{0.000000in}}{%
\pgfpathmoveto{\pgfqpoint{0.000000in}{0.000000in}}%
\pgfpathlineto{\pgfqpoint{-0.027778in}{0.000000in}}%
\pgfusepath{stroke,fill}%
}%
\begin{pgfscope}%
\pgfsys@transformshift{0.422284in}{0.366282in}%
\pgfsys@useobject{currentmarker}{}%
\end{pgfscope}%
\end{pgfscope}%
\begin{pgfscope}%
\pgfsetbuttcap%
\pgfsetroundjoin%
\definecolor{currentfill}{rgb}{0.000000,0.000000,0.000000}%
\pgfsetfillcolor{currentfill}%
\pgfsetlinewidth{0.602250pt}%
\definecolor{currentstroke}{rgb}{0.000000,0.000000,0.000000}%
\pgfsetstrokecolor{currentstroke}%
\pgfsetdash{}{0pt}%
\pgfsys@defobject{currentmarker}{\pgfqpoint{-0.027778in}{0.000000in}}{\pgfqpoint{0.000000in}{0.000000in}}{%
\pgfpathmoveto{\pgfqpoint{0.000000in}{0.000000in}}%
\pgfpathlineto{\pgfqpoint{-0.027778in}{0.000000in}}%
\pgfusepath{stroke,fill}%
}%
\begin{pgfscope}%
\pgfsys@transformshift{0.422284in}{0.384365in}%
\pgfsys@useobject{currentmarker}{}%
\end{pgfscope}%
\end{pgfscope}%
\begin{pgfscope}%
\pgfsetbuttcap%
\pgfsetroundjoin%
\definecolor{currentfill}{rgb}{0.000000,0.000000,0.000000}%
\pgfsetfillcolor{currentfill}%
\pgfsetlinewidth{0.602250pt}%
\definecolor{currentstroke}{rgb}{0.000000,0.000000,0.000000}%
\pgfsetstrokecolor{currentstroke}%
\pgfsetdash{}{0pt}%
\pgfsys@defobject{currentmarker}{\pgfqpoint{-0.027778in}{0.000000in}}{\pgfqpoint{0.000000in}{0.000000in}}{%
\pgfpathmoveto{\pgfqpoint{0.000000in}{0.000000in}}%
\pgfpathlineto{\pgfqpoint{-0.027778in}{0.000000in}}%
\pgfusepath{stroke,fill}%
}%
\begin{pgfscope}%
\pgfsys@transformshift{0.422284in}{0.506959in}%
\pgfsys@useobject{currentmarker}{}%
\end{pgfscope}%
\end{pgfscope}%
\begin{pgfscope}%
\pgfsetbuttcap%
\pgfsetroundjoin%
\definecolor{currentfill}{rgb}{0.000000,0.000000,0.000000}%
\pgfsetfillcolor{currentfill}%
\pgfsetlinewidth{0.602250pt}%
\definecolor{currentstroke}{rgb}{0.000000,0.000000,0.000000}%
\pgfsetstrokecolor{currentstroke}%
\pgfsetdash{}{0pt}%
\pgfsys@defobject{currentmarker}{\pgfqpoint{-0.027778in}{0.000000in}}{\pgfqpoint{0.000000in}{0.000000in}}{%
\pgfpathmoveto{\pgfqpoint{0.000000in}{0.000000in}}%
\pgfpathlineto{\pgfqpoint{-0.027778in}{0.000000in}}%
\pgfusepath{stroke,fill}%
}%
\begin{pgfscope}%
\pgfsys@transformshift{0.422284in}{0.569209in}%
\pgfsys@useobject{currentmarker}{}%
\end{pgfscope}%
\end{pgfscope}%
\begin{pgfscope}%
\pgfsetbuttcap%
\pgfsetroundjoin%
\definecolor{currentfill}{rgb}{0.000000,0.000000,0.000000}%
\pgfsetfillcolor{currentfill}%
\pgfsetlinewidth{0.602250pt}%
\definecolor{currentstroke}{rgb}{0.000000,0.000000,0.000000}%
\pgfsetstrokecolor{currentstroke}%
\pgfsetdash{}{0pt}%
\pgfsys@defobject{currentmarker}{\pgfqpoint{-0.027778in}{0.000000in}}{\pgfqpoint{0.000000in}{0.000000in}}{%
\pgfpathmoveto{\pgfqpoint{0.000000in}{0.000000in}}%
\pgfpathlineto{\pgfqpoint{-0.027778in}{0.000000in}}%
\pgfusepath{stroke,fill}%
}%
\begin{pgfscope}%
\pgfsys@transformshift{0.422284in}{0.613377in}%
\pgfsys@useobject{currentmarker}{}%
\end{pgfscope}%
\end{pgfscope}%
\begin{pgfscope}%
\pgfsetbuttcap%
\pgfsetroundjoin%
\definecolor{currentfill}{rgb}{0.000000,0.000000,0.000000}%
\pgfsetfillcolor{currentfill}%
\pgfsetlinewidth{0.602250pt}%
\definecolor{currentstroke}{rgb}{0.000000,0.000000,0.000000}%
\pgfsetstrokecolor{currentstroke}%
\pgfsetdash{}{0pt}%
\pgfsys@defobject{currentmarker}{\pgfqpoint{-0.027778in}{0.000000in}}{\pgfqpoint{0.000000in}{0.000000in}}{%
\pgfpathmoveto{\pgfqpoint{0.000000in}{0.000000in}}%
\pgfpathlineto{\pgfqpoint{-0.027778in}{0.000000in}}%
\pgfusepath{stroke,fill}%
}%
\begin{pgfscope}%
\pgfsys@transformshift{0.422284in}{0.647636in}%
\pgfsys@useobject{currentmarker}{}%
\end{pgfscope}%
\end{pgfscope}%
\begin{pgfscope}%
\pgfsetbuttcap%
\pgfsetroundjoin%
\definecolor{currentfill}{rgb}{0.000000,0.000000,0.000000}%
\pgfsetfillcolor{currentfill}%
\pgfsetlinewidth{0.602250pt}%
\definecolor{currentstroke}{rgb}{0.000000,0.000000,0.000000}%
\pgfsetstrokecolor{currentstroke}%
\pgfsetdash{}{0pt}%
\pgfsys@defobject{currentmarker}{\pgfqpoint{-0.027778in}{0.000000in}}{\pgfqpoint{0.000000in}{0.000000in}}{%
\pgfpathmoveto{\pgfqpoint{0.000000in}{0.000000in}}%
\pgfpathlineto{\pgfqpoint{-0.027778in}{0.000000in}}%
\pgfusepath{stroke,fill}%
}%
\begin{pgfscope}%
\pgfsys@transformshift{0.422284in}{0.675627in}%
\pgfsys@useobject{currentmarker}{}%
\end{pgfscope}%
\end{pgfscope}%
\begin{pgfscope}%
\pgfsetbuttcap%
\pgfsetroundjoin%
\definecolor{currentfill}{rgb}{0.000000,0.000000,0.000000}%
\pgfsetfillcolor{currentfill}%
\pgfsetlinewidth{0.602250pt}%
\definecolor{currentstroke}{rgb}{0.000000,0.000000,0.000000}%
\pgfsetstrokecolor{currentstroke}%
\pgfsetdash{}{0pt}%
\pgfsys@defobject{currentmarker}{\pgfqpoint{-0.027778in}{0.000000in}}{\pgfqpoint{0.000000in}{0.000000in}}{%
\pgfpathmoveto{\pgfqpoint{0.000000in}{0.000000in}}%
\pgfpathlineto{\pgfqpoint{-0.027778in}{0.000000in}}%
\pgfusepath{stroke,fill}%
}%
\begin{pgfscope}%
\pgfsys@transformshift{0.422284in}{0.699294in}%
\pgfsys@useobject{currentmarker}{}%
\end{pgfscope}%
\end{pgfscope}%
\begin{pgfscope}%
\pgfsetbuttcap%
\pgfsetroundjoin%
\definecolor{currentfill}{rgb}{0.000000,0.000000,0.000000}%
\pgfsetfillcolor{currentfill}%
\pgfsetlinewidth{0.602250pt}%
\definecolor{currentstroke}{rgb}{0.000000,0.000000,0.000000}%
\pgfsetstrokecolor{currentstroke}%
\pgfsetdash{}{0pt}%
\pgfsys@defobject{currentmarker}{\pgfqpoint{-0.027778in}{0.000000in}}{\pgfqpoint{0.000000in}{0.000000in}}{%
\pgfpathmoveto{\pgfqpoint{0.000000in}{0.000000in}}%
\pgfpathlineto{\pgfqpoint{-0.027778in}{0.000000in}}%
\pgfusepath{stroke,fill}%
}%
\begin{pgfscope}%
\pgfsys@transformshift{0.422284in}{0.719795in}%
\pgfsys@useobject{currentmarker}{}%
\end{pgfscope}%
\end{pgfscope}%
\begin{pgfscope}%
\pgfsetbuttcap%
\pgfsetroundjoin%
\definecolor{currentfill}{rgb}{0.000000,0.000000,0.000000}%
\pgfsetfillcolor{currentfill}%
\pgfsetlinewidth{0.602250pt}%
\definecolor{currentstroke}{rgb}{0.000000,0.000000,0.000000}%
\pgfsetstrokecolor{currentstroke}%
\pgfsetdash{}{0pt}%
\pgfsys@defobject{currentmarker}{\pgfqpoint{-0.027778in}{0.000000in}}{\pgfqpoint{0.000000in}{0.000000in}}{%
\pgfpathmoveto{\pgfqpoint{0.000000in}{0.000000in}}%
\pgfpathlineto{\pgfqpoint{-0.027778in}{0.000000in}}%
\pgfusepath{stroke,fill}%
}%
\begin{pgfscope}%
\pgfsys@transformshift{0.422284in}{0.737878in}%
\pgfsys@useobject{currentmarker}{}%
\end{pgfscope}%
\end{pgfscope}%
\begin{pgfscope}%
\pgfsetbuttcap%
\pgfsetroundjoin%
\definecolor{currentfill}{rgb}{0.000000,0.000000,0.000000}%
\pgfsetfillcolor{currentfill}%
\pgfsetlinewidth{0.602250pt}%
\definecolor{currentstroke}{rgb}{0.000000,0.000000,0.000000}%
\pgfsetstrokecolor{currentstroke}%
\pgfsetdash{}{0pt}%
\pgfsys@defobject{currentmarker}{\pgfqpoint{-0.027778in}{0.000000in}}{\pgfqpoint{0.000000in}{0.000000in}}{%
\pgfpathmoveto{\pgfqpoint{0.000000in}{0.000000in}}%
\pgfpathlineto{\pgfqpoint{-0.027778in}{0.000000in}}%
\pgfusepath{stroke,fill}%
}%
\begin{pgfscope}%
\pgfsys@transformshift{0.422284in}{0.860472in}%
\pgfsys@useobject{currentmarker}{}%
\end{pgfscope}%
\end{pgfscope}%
\begin{pgfscope}%
\pgfsetbuttcap%
\pgfsetroundjoin%
\definecolor{currentfill}{rgb}{0.000000,0.000000,0.000000}%
\pgfsetfillcolor{currentfill}%
\pgfsetlinewidth{0.602250pt}%
\definecolor{currentstroke}{rgb}{0.000000,0.000000,0.000000}%
\pgfsetstrokecolor{currentstroke}%
\pgfsetdash{}{0pt}%
\pgfsys@defobject{currentmarker}{\pgfqpoint{-0.027778in}{0.000000in}}{\pgfqpoint{0.000000in}{0.000000in}}{%
\pgfpathmoveto{\pgfqpoint{0.000000in}{0.000000in}}%
\pgfpathlineto{\pgfqpoint{-0.027778in}{0.000000in}}%
\pgfusepath{stroke,fill}%
}%
\begin{pgfscope}%
\pgfsys@transformshift{0.422284in}{0.922722in}%
\pgfsys@useobject{currentmarker}{}%
\end{pgfscope}%
\end{pgfscope}%
\begin{pgfscope}%
\pgfsetbuttcap%
\pgfsetroundjoin%
\definecolor{currentfill}{rgb}{0.000000,0.000000,0.000000}%
\pgfsetfillcolor{currentfill}%
\pgfsetlinewidth{0.602250pt}%
\definecolor{currentstroke}{rgb}{0.000000,0.000000,0.000000}%
\pgfsetstrokecolor{currentstroke}%
\pgfsetdash{}{0pt}%
\pgfsys@defobject{currentmarker}{\pgfqpoint{-0.027778in}{0.000000in}}{\pgfqpoint{0.000000in}{0.000000in}}{%
\pgfpathmoveto{\pgfqpoint{0.000000in}{0.000000in}}%
\pgfpathlineto{\pgfqpoint{-0.027778in}{0.000000in}}%
\pgfusepath{stroke,fill}%
}%
\begin{pgfscope}%
\pgfsys@transformshift{0.422284in}{0.966890in}%
\pgfsys@useobject{currentmarker}{}%
\end{pgfscope}%
\end{pgfscope}%
\begin{pgfscope}%
\pgfsetbuttcap%
\pgfsetroundjoin%
\definecolor{currentfill}{rgb}{0.000000,0.000000,0.000000}%
\pgfsetfillcolor{currentfill}%
\pgfsetlinewidth{0.602250pt}%
\definecolor{currentstroke}{rgb}{0.000000,0.000000,0.000000}%
\pgfsetstrokecolor{currentstroke}%
\pgfsetdash{}{0pt}%
\pgfsys@defobject{currentmarker}{\pgfqpoint{-0.027778in}{0.000000in}}{\pgfqpoint{0.000000in}{0.000000in}}{%
\pgfpathmoveto{\pgfqpoint{0.000000in}{0.000000in}}%
\pgfpathlineto{\pgfqpoint{-0.027778in}{0.000000in}}%
\pgfusepath{stroke,fill}%
}%
\begin{pgfscope}%
\pgfsys@transformshift{0.422284in}{1.001149in}%
\pgfsys@useobject{currentmarker}{}%
\end{pgfscope}%
\end{pgfscope}%
\begin{pgfscope}%
\pgfsetbuttcap%
\pgfsetroundjoin%
\definecolor{currentfill}{rgb}{0.000000,0.000000,0.000000}%
\pgfsetfillcolor{currentfill}%
\pgfsetlinewidth{0.602250pt}%
\definecolor{currentstroke}{rgb}{0.000000,0.000000,0.000000}%
\pgfsetstrokecolor{currentstroke}%
\pgfsetdash{}{0pt}%
\pgfsys@defobject{currentmarker}{\pgfqpoint{-0.027778in}{0.000000in}}{\pgfqpoint{0.000000in}{0.000000in}}{%
\pgfpathmoveto{\pgfqpoint{0.000000in}{0.000000in}}%
\pgfpathlineto{\pgfqpoint{-0.027778in}{0.000000in}}%
\pgfusepath{stroke,fill}%
}%
\begin{pgfscope}%
\pgfsys@transformshift{0.422284in}{1.029140in}%
\pgfsys@useobject{currentmarker}{}%
\end{pgfscope}%
\end{pgfscope}%
\begin{pgfscope}%
\pgfsetbuttcap%
\pgfsetroundjoin%
\definecolor{currentfill}{rgb}{0.000000,0.000000,0.000000}%
\pgfsetfillcolor{currentfill}%
\pgfsetlinewidth{0.602250pt}%
\definecolor{currentstroke}{rgb}{0.000000,0.000000,0.000000}%
\pgfsetstrokecolor{currentstroke}%
\pgfsetdash{}{0pt}%
\pgfsys@defobject{currentmarker}{\pgfqpoint{-0.027778in}{0.000000in}}{\pgfqpoint{0.000000in}{0.000000in}}{%
\pgfpathmoveto{\pgfqpoint{0.000000in}{0.000000in}}%
\pgfpathlineto{\pgfqpoint{-0.027778in}{0.000000in}}%
\pgfusepath{stroke,fill}%
}%
\begin{pgfscope}%
\pgfsys@transformshift{0.422284in}{1.052807in}%
\pgfsys@useobject{currentmarker}{}%
\end{pgfscope}%
\end{pgfscope}%
\begin{pgfscope}%
\pgfsetbuttcap%
\pgfsetroundjoin%
\definecolor{currentfill}{rgb}{0.000000,0.000000,0.000000}%
\pgfsetfillcolor{currentfill}%
\pgfsetlinewidth{0.602250pt}%
\definecolor{currentstroke}{rgb}{0.000000,0.000000,0.000000}%
\pgfsetstrokecolor{currentstroke}%
\pgfsetdash{}{0pt}%
\pgfsys@defobject{currentmarker}{\pgfqpoint{-0.027778in}{0.000000in}}{\pgfqpoint{0.000000in}{0.000000in}}{%
\pgfpathmoveto{\pgfqpoint{0.000000in}{0.000000in}}%
\pgfpathlineto{\pgfqpoint{-0.027778in}{0.000000in}}%
\pgfusepath{stroke,fill}%
}%
\begin{pgfscope}%
\pgfsys@transformshift{0.422284in}{1.073308in}%
\pgfsys@useobject{currentmarker}{}%
\end{pgfscope}%
\end{pgfscope}%
\begin{pgfscope}%
\pgfsetbuttcap%
\pgfsetroundjoin%
\definecolor{currentfill}{rgb}{0.000000,0.000000,0.000000}%
\pgfsetfillcolor{currentfill}%
\pgfsetlinewidth{0.602250pt}%
\definecolor{currentstroke}{rgb}{0.000000,0.000000,0.000000}%
\pgfsetstrokecolor{currentstroke}%
\pgfsetdash{}{0pt}%
\pgfsys@defobject{currentmarker}{\pgfqpoint{-0.027778in}{0.000000in}}{\pgfqpoint{0.000000in}{0.000000in}}{%
\pgfpathmoveto{\pgfqpoint{0.000000in}{0.000000in}}%
\pgfpathlineto{\pgfqpoint{-0.027778in}{0.000000in}}%
\pgfusepath{stroke,fill}%
}%
\begin{pgfscope}%
\pgfsys@transformshift{0.422284in}{1.091391in}%
\pgfsys@useobject{currentmarker}{}%
\end{pgfscope}%
\end{pgfscope}%
\begin{pgfscope}%
\pgfsetbuttcap%
\pgfsetroundjoin%
\definecolor{currentfill}{rgb}{0.000000,0.000000,0.000000}%
\pgfsetfillcolor{currentfill}%
\pgfsetlinewidth{0.602250pt}%
\definecolor{currentstroke}{rgb}{0.000000,0.000000,0.000000}%
\pgfsetstrokecolor{currentstroke}%
\pgfsetdash{}{0pt}%
\pgfsys@defobject{currentmarker}{\pgfqpoint{-0.027778in}{0.000000in}}{\pgfqpoint{0.000000in}{0.000000in}}{%
\pgfpathmoveto{\pgfqpoint{0.000000in}{0.000000in}}%
\pgfpathlineto{\pgfqpoint{-0.027778in}{0.000000in}}%
\pgfusepath{stroke,fill}%
}%
\begin{pgfscope}%
\pgfsys@transformshift{0.422284in}{1.213985in}%
\pgfsys@useobject{currentmarker}{}%
\end{pgfscope}%
\end{pgfscope}%
\begin{pgfscope}%
\pgfsetbuttcap%
\pgfsetroundjoin%
\definecolor{currentfill}{rgb}{0.000000,0.000000,0.000000}%
\pgfsetfillcolor{currentfill}%
\pgfsetlinewidth{0.602250pt}%
\definecolor{currentstroke}{rgb}{0.000000,0.000000,0.000000}%
\pgfsetstrokecolor{currentstroke}%
\pgfsetdash{}{0pt}%
\pgfsys@defobject{currentmarker}{\pgfqpoint{-0.027778in}{0.000000in}}{\pgfqpoint{0.000000in}{0.000000in}}{%
\pgfpathmoveto{\pgfqpoint{0.000000in}{0.000000in}}%
\pgfpathlineto{\pgfqpoint{-0.027778in}{0.000000in}}%
\pgfusepath{stroke,fill}%
}%
\begin{pgfscope}%
\pgfsys@transformshift{0.422284in}{1.276235in}%
\pgfsys@useobject{currentmarker}{}%
\end{pgfscope}%
\end{pgfscope}%
\begin{pgfscope}%
\pgfsetbuttcap%
\pgfsetroundjoin%
\definecolor{currentfill}{rgb}{0.000000,0.000000,0.000000}%
\pgfsetfillcolor{currentfill}%
\pgfsetlinewidth{0.602250pt}%
\definecolor{currentstroke}{rgb}{0.000000,0.000000,0.000000}%
\pgfsetstrokecolor{currentstroke}%
\pgfsetdash{}{0pt}%
\pgfsys@defobject{currentmarker}{\pgfqpoint{-0.027778in}{0.000000in}}{\pgfqpoint{0.000000in}{0.000000in}}{%
\pgfpathmoveto{\pgfqpoint{0.000000in}{0.000000in}}%
\pgfpathlineto{\pgfqpoint{-0.027778in}{0.000000in}}%
\pgfusepath{stroke,fill}%
}%
\begin{pgfscope}%
\pgfsys@transformshift{0.422284in}{1.320403in}%
\pgfsys@useobject{currentmarker}{}%
\end{pgfscope}%
\end{pgfscope}%
\begin{pgfscope}%
\pgfsetbuttcap%
\pgfsetroundjoin%
\definecolor{currentfill}{rgb}{0.000000,0.000000,0.000000}%
\pgfsetfillcolor{currentfill}%
\pgfsetlinewidth{0.602250pt}%
\definecolor{currentstroke}{rgb}{0.000000,0.000000,0.000000}%
\pgfsetstrokecolor{currentstroke}%
\pgfsetdash{}{0pt}%
\pgfsys@defobject{currentmarker}{\pgfqpoint{-0.027778in}{0.000000in}}{\pgfqpoint{0.000000in}{0.000000in}}{%
\pgfpathmoveto{\pgfqpoint{0.000000in}{0.000000in}}%
\pgfpathlineto{\pgfqpoint{-0.027778in}{0.000000in}}%
\pgfusepath{stroke,fill}%
}%
\begin{pgfscope}%
\pgfsys@transformshift{0.422284in}{1.354661in}%
\pgfsys@useobject{currentmarker}{}%
\end{pgfscope}%
\end{pgfscope}%
\begin{pgfscope}%
\pgfsetbuttcap%
\pgfsetroundjoin%
\definecolor{currentfill}{rgb}{0.000000,0.000000,0.000000}%
\pgfsetfillcolor{currentfill}%
\pgfsetlinewidth{0.602250pt}%
\definecolor{currentstroke}{rgb}{0.000000,0.000000,0.000000}%
\pgfsetstrokecolor{currentstroke}%
\pgfsetdash{}{0pt}%
\pgfsys@defobject{currentmarker}{\pgfqpoint{-0.027778in}{0.000000in}}{\pgfqpoint{0.000000in}{0.000000in}}{%
\pgfpathmoveto{\pgfqpoint{0.000000in}{0.000000in}}%
\pgfpathlineto{\pgfqpoint{-0.027778in}{0.000000in}}%
\pgfusepath{stroke,fill}%
}%
\begin{pgfscope}%
\pgfsys@transformshift{0.422284in}{1.382653in}%
\pgfsys@useobject{currentmarker}{}%
\end{pgfscope}%
\end{pgfscope}%
\begin{pgfscope}%
\pgfsetbuttcap%
\pgfsetroundjoin%
\definecolor{currentfill}{rgb}{0.000000,0.000000,0.000000}%
\pgfsetfillcolor{currentfill}%
\pgfsetlinewidth{0.602250pt}%
\definecolor{currentstroke}{rgb}{0.000000,0.000000,0.000000}%
\pgfsetstrokecolor{currentstroke}%
\pgfsetdash{}{0pt}%
\pgfsys@defobject{currentmarker}{\pgfqpoint{-0.027778in}{0.000000in}}{\pgfqpoint{0.000000in}{0.000000in}}{%
\pgfpathmoveto{\pgfqpoint{0.000000in}{0.000000in}}%
\pgfpathlineto{\pgfqpoint{-0.027778in}{0.000000in}}%
\pgfusepath{stroke,fill}%
}%
\begin{pgfscope}%
\pgfsys@transformshift{0.422284in}{1.406320in}%
\pgfsys@useobject{currentmarker}{}%
\end{pgfscope}%
\end{pgfscope}%
\begin{pgfscope}%
\pgfsetbuttcap%
\pgfsetroundjoin%
\definecolor{currentfill}{rgb}{0.000000,0.000000,0.000000}%
\pgfsetfillcolor{currentfill}%
\pgfsetlinewidth{0.602250pt}%
\definecolor{currentstroke}{rgb}{0.000000,0.000000,0.000000}%
\pgfsetstrokecolor{currentstroke}%
\pgfsetdash{}{0pt}%
\pgfsys@defobject{currentmarker}{\pgfqpoint{-0.027778in}{0.000000in}}{\pgfqpoint{0.000000in}{0.000000in}}{%
\pgfpathmoveto{\pgfqpoint{0.000000in}{0.000000in}}%
\pgfpathlineto{\pgfqpoint{-0.027778in}{0.000000in}}%
\pgfusepath{stroke,fill}%
}%
\begin{pgfscope}%
\pgfsys@transformshift{0.422284in}{1.426821in}%
\pgfsys@useobject{currentmarker}{}%
\end{pgfscope}%
\end{pgfscope}%
\begin{pgfscope}%
\pgfsetbuttcap%
\pgfsetroundjoin%
\definecolor{currentfill}{rgb}{0.000000,0.000000,0.000000}%
\pgfsetfillcolor{currentfill}%
\pgfsetlinewidth{0.602250pt}%
\definecolor{currentstroke}{rgb}{0.000000,0.000000,0.000000}%
\pgfsetstrokecolor{currentstroke}%
\pgfsetdash{}{0pt}%
\pgfsys@defobject{currentmarker}{\pgfqpoint{-0.027778in}{0.000000in}}{\pgfqpoint{0.000000in}{0.000000in}}{%
\pgfpathmoveto{\pgfqpoint{0.000000in}{0.000000in}}%
\pgfpathlineto{\pgfqpoint{-0.027778in}{0.000000in}}%
\pgfusepath{stroke,fill}%
}%
\begin{pgfscope}%
\pgfsys@transformshift{0.422284in}{1.444904in}%
\pgfsys@useobject{currentmarker}{}%
\end{pgfscope}%
\end{pgfscope}%
\begin{pgfscope}%
\definecolor{textcolor}{rgb}{0.000000,0.000000,0.000000}%
\pgfsetstrokecolor{textcolor}%
\pgfsetfillcolor{textcolor}%
\pgftext[x=0.162642in,y=0.891597in,,bottom,rotate=90.000000]{\color{textcolor}\rmfamily\fontsize{9.000000}{10.800000}\selectfont time (ms)}%
\end{pgfscope}%
\begin{pgfscope}%
\pgfpathrectangle{\pgfqpoint{0.422284in}{0.259864in}}{\pgfqpoint{1.301046in}{1.263466in}}%
\pgfusepath{clip}%
\pgfsetbuttcap%
\pgfsetroundjoin%
\pgfsetlinewidth{1.505625pt}%
\definecolor{currentstroke}{rgb}{0.376471,0.376471,0.376471}%
\pgfsetstrokecolor{currentstroke}%
\pgfsetdash{}{0pt}%
\pgfpathmoveto{\pgfqpoint{0.535185in}{0.383460in}}%
\pgfpathlineto{\pgfqpoint{0.535185in}{0.395343in}}%
\pgfusepath{stroke}%
\end{pgfscope}%
\begin{pgfscope}%
\pgfpathrectangle{\pgfqpoint{0.422284in}{0.259864in}}{\pgfqpoint{1.301046in}{1.263466in}}%
\pgfusepath{clip}%
\pgfsetbuttcap%
\pgfsetroundjoin%
\pgfsetlinewidth{1.505625pt}%
\definecolor{currentstroke}{rgb}{0.376471,0.376471,0.376471}%
\pgfsetstrokecolor{currentstroke}%
\pgfsetdash{}{0pt}%
\pgfpathmoveto{\pgfqpoint{0.965283in}{0.489981in}}%
\pgfpathlineto{\pgfqpoint{0.965283in}{0.499562in}}%
\pgfusepath{stroke}%
\end{pgfscope}%
\begin{pgfscope}%
\pgfpathrectangle{\pgfqpoint{0.422284in}{0.259864in}}{\pgfqpoint{1.301046in}{1.263466in}}%
\pgfusepath{clip}%
\pgfsetbuttcap%
\pgfsetroundjoin%
\pgfsetlinewidth{1.505625pt}%
\definecolor{currentstroke}{rgb}{0.376471,0.376471,0.376471}%
\pgfsetstrokecolor{currentstroke}%
\pgfsetdash{}{0pt}%
\pgfpathmoveto{\pgfqpoint{1.395380in}{0.630655in}}%
\pgfpathlineto{\pgfqpoint{1.395380in}{0.776380in}}%
\pgfusepath{stroke}%
\end{pgfscope}%
\begin{pgfscope}%
\pgfpathrectangle{\pgfqpoint{0.422284in}{0.259864in}}{\pgfqpoint{1.301046in}{1.263466in}}%
\pgfusepath{clip}%
\pgfsetbuttcap%
\pgfsetroundjoin%
\pgfsetlinewidth{1.505625pt}%
\definecolor{currentstroke}{rgb}{0.376471,0.376471,0.376471}%
\pgfsetstrokecolor{currentstroke}%
\pgfsetdash{}{0pt}%
\pgfpathmoveto{\pgfqpoint{0.642709in}{0.398658in}}%
\pgfpathlineto{\pgfqpoint{0.642709in}{0.410673in}}%
\pgfusepath{stroke}%
\end{pgfscope}%
\begin{pgfscope}%
\pgfpathrectangle{\pgfqpoint{0.422284in}{0.259864in}}{\pgfqpoint{1.301046in}{1.263466in}}%
\pgfusepath{clip}%
\pgfsetbuttcap%
\pgfsetroundjoin%
\pgfsetlinewidth{1.505625pt}%
\definecolor{currentstroke}{rgb}{0.376471,0.376471,0.376471}%
\pgfsetstrokecolor{currentstroke}%
\pgfsetdash{}{0pt}%
\pgfpathmoveto{\pgfqpoint{1.072807in}{0.498536in}}%
\pgfpathlineto{\pgfqpoint{1.072807in}{0.505735in}}%
\pgfusepath{stroke}%
\end{pgfscope}%
\begin{pgfscope}%
\pgfpathrectangle{\pgfqpoint{0.422284in}{0.259864in}}{\pgfqpoint{1.301046in}{1.263466in}}%
\pgfusepath{clip}%
\pgfsetbuttcap%
\pgfsetroundjoin%
\pgfsetlinewidth{1.505625pt}%
\definecolor{currentstroke}{rgb}{0.376471,0.376471,0.376471}%
\pgfsetstrokecolor{currentstroke}%
\pgfsetdash{}{0pt}%
\pgfpathmoveto{\pgfqpoint{1.502905in}{0.802057in}}%
\pgfpathlineto{\pgfqpoint{1.502905in}{0.950391in}}%
\pgfusepath{stroke}%
\end{pgfscope}%
\begin{pgfscope}%
\pgfpathrectangle{\pgfqpoint{0.422284in}{0.259864in}}{\pgfqpoint{1.301046in}{1.263466in}}%
\pgfusepath{clip}%
\pgfsetbuttcap%
\pgfsetroundjoin%
\pgfsetlinewidth{1.505625pt}%
\definecolor{currentstroke}{rgb}{0.376471,0.376471,0.376471}%
\pgfsetstrokecolor{currentstroke}%
\pgfsetdash{}{0pt}%
\pgfpathmoveto{\pgfqpoint{0.750234in}{0.456920in}}%
\pgfpathlineto{\pgfqpoint{0.750234in}{0.486451in}}%
\pgfusepath{stroke}%
\end{pgfscope}%
\begin{pgfscope}%
\pgfpathrectangle{\pgfqpoint{0.422284in}{0.259864in}}{\pgfqpoint{1.301046in}{1.263466in}}%
\pgfusepath{clip}%
\pgfsetbuttcap%
\pgfsetroundjoin%
\pgfsetlinewidth{1.505625pt}%
\definecolor{currentstroke}{rgb}{0.376471,0.376471,0.376471}%
\pgfsetstrokecolor{currentstroke}%
\pgfsetdash{}{0pt}%
\pgfpathmoveto{\pgfqpoint{1.180332in}{0.554061in}}%
\pgfpathlineto{\pgfqpoint{1.180332in}{0.561148in}}%
\pgfusepath{stroke}%
\end{pgfscope}%
\begin{pgfscope}%
\pgfpathrectangle{\pgfqpoint{0.422284in}{0.259864in}}{\pgfqpoint{1.301046in}{1.263466in}}%
\pgfusepath{clip}%
\pgfsetbuttcap%
\pgfsetroundjoin%
\pgfsetlinewidth{1.505625pt}%
\definecolor{currentstroke}{rgb}{0.376471,0.376471,0.376471}%
\pgfsetstrokecolor{currentstroke}%
\pgfsetdash{}{0pt}%
\pgfpathmoveto{\pgfqpoint{1.610429in}{1.110948in}}%
\pgfpathlineto{\pgfqpoint{1.610429in}{1.386140in}}%
\pgfusepath{stroke}%
\end{pgfscope}%
\begin{pgfscope}%
\pgfpathrectangle{\pgfqpoint{0.422284in}{0.259864in}}{\pgfqpoint{1.301046in}{1.263466in}}%
\pgfusepath{clip}%
\pgfsetbuttcap%
\pgfsetroundjoin%
\definecolor{currentfill}{rgb}{0.376471,0.376471,0.376471}%
\pgfsetfillcolor{currentfill}%
\pgfsetlinewidth{0.803000pt}%
\definecolor{currentstroke}{rgb}{0.376471,0.376471,0.376471}%
\pgfsetstrokecolor{currentstroke}%
\pgfsetdash{}{0pt}%
\pgfsys@defobject{currentmarker}{\pgfqpoint{-0.027778in}{-0.000000in}}{\pgfqpoint{0.027778in}{0.000000in}}{%
\pgfpathmoveto{\pgfqpoint{0.027778in}{-0.000000in}}%
\pgfpathlineto{\pgfqpoint{-0.027778in}{0.000000in}}%
\pgfusepath{stroke,fill}%
}%
\begin{pgfscope}%
\pgfsys@transformshift{0.535185in}{0.383460in}%
\pgfsys@useobject{currentmarker}{}%
\end{pgfscope}%
\begin{pgfscope}%
\pgfsys@transformshift{0.965283in}{0.489981in}%
\pgfsys@useobject{currentmarker}{}%
\end{pgfscope}%
\begin{pgfscope}%
\pgfsys@transformshift{1.395380in}{0.630655in}%
\pgfsys@useobject{currentmarker}{}%
\end{pgfscope}%
\end{pgfscope}%
\begin{pgfscope}%
\pgfpathrectangle{\pgfqpoint{0.422284in}{0.259864in}}{\pgfqpoint{1.301046in}{1.263466in}}%
\pgfusepath{clip}%
\pgfsetbuttcap%
\pgfsetroundjoin%
\definecolor{currentfill}{rgb}{0.376471,0.376471,0.376471}%
\pgfsetfillcolor{currentfill}%
\pgfsetlinewidth{0.803000pt}%
\definecolor{currentstroke}{rgb}{0.376471,0.376471,0.376471}%
\pgfsetstrokecolor{currentstroke}%
\pgfsetdash{}{0pt}%
\pgfsys@defobject{currentmarker}{\pgfqpoint{-0.027778in}{-0.000000in}}{\pgfqpoint{0.027778in}{0.000000in}}{%
\pgfpathmoveto{\pgfqpoint{0.027778in}{-0.000000in}}%
\pgfpathlineto{\pgfqpoint{-0.027778in}{0.000000in}}%
\pgfusepath{stroke,fill}%
}%
\begin{pgfscope}%
\pgfsys@transformshift{0.535185in}{0.395343in}%
\pgfsys@useobject{currentmarker}{}%
\end{pgfscope}%
\begin{pgfscope}%
\pgfsys@transformshift{0.965283in}{0.499562in}%
\pgfsys@useobject{currentmarker}{}%
\end{pgfscope}%
\begin{pgfscope}%
\pgfsys@transformshift{1.395380in}{0.776380in}%
\pgfsys@useobject{currentmarker}{}%
\end{pgfscope}%
\end{pgfscope}%
\begin{pgfscope}%
\pgfpathrectangle{\pgfqpoint{0.422284in}{0.259864in}}{\pgfqpoint{1.301046in}{1.263466in}}%
\pgfusepath{clip}%
\pgfsetbuttcap%
\pgfsetroundjoin%
\definecolor{currentfill}{rgb}{0.376471,0.376471,0.376471}%
\pgfsetfillcolor{currentfill}%
\pgfsetlinewidth{0.803000pt}%
\definecolor{currentstroke}{rgb}{0.376471,0.376471,0.376471}%
\pgfsetstrokecolor{currentstroke}%
\pgfsetdash{}{0pt}%
\pgfsys@defobject{currentmarker}{\pgfqpoint{-0.027778in}{-0.000000in}}{\pgfqpoint{0.027778in}{0.000000in}}{%
\pgfpathmoveto{\pgfqpoint{0.027778in}{-0.000000in}}%
\pgfpathlineto{\pgfqpoint{-0.027778in}{0.000000in}}%
\pgfusepath{stroke,fill}%
}%
\begin{pgfscope}%
\pgfsys@transformshift{0.642709in}{0.398658in}%
\pgfsys@useobject{currentmarker}{}%
\end{pgfscope}%
\begin{pgfscope}%
\pgfsys@transformshift{1.072807in}{0.498536in}%
\pgfsys@useobject{currentmarker}{}%
\end{pgfscope}%
\begin{pgfscope}%
\pgfsys@transformshift{1.502905in}{0.802057in}%
\pgfsys@useobject{currentmarker}{}%
\end{pgfscope}%
\end{pgfscope}%
\begin{pgfscope}%
\pgfpathrectangle{\pgfqpoint{0.422284in}{0.259864in}}{\pgfqpoint{1.301046in}{1.263466in}}%
\pgfusepath{clip}%
\pgfsetbuttcap%
\pgfsetroundjoin%
\definecolor{currentfill}{rgb}{0.376471,0.376471,0.376471}%
\pgfsetfillcolor{currentfill}%
\pgfsetlinewidth{0.803000pt}%
\definecolor{currentstroke}{rgb}{0.376471,0.376471,0.376471}%
\pgfsetstrokecolor{currentstroke}%
\pgfsetdash{}{0pt}%
\pgfsys@defobject{currentmarker}{\pgfqpoint{-0.027778in}{-0.000000in}}{\pgfqpoint{0.027778in}{0.000000in}}{%
\pgfpathmoveto{\pgfqpoint{0.027778in}{-0.000000in}}%
\pgfpathlineto{\pgfqpoint{-0.027778in}{0.000000in}}%
\pgfusepath{stroke,fill}%
}%
\begin{pgfscope}%
\pgfsys@transformshift{0.642709in}{0.410673in}%
\pgfsys@useobject{currentmarker}{}%
\end{pgfscope}%
\begin{pgfscope}%
\pgfsys@transformshift{1.072807in}{0.505735in}%
\pgfsys@useobject{currentmarker}{}%
\end{pgfscope}%
\begin{pgfscope}%
\pgfsys@transformshift{1.502905in}{0.950391in}%
\pgfsys@useobject{currentmarker}{}%
\end{pgfscope}%
\end{pgfscope}%
\begin{pgfscope}%
\pgfpathrectangle{\pgfqpoint{0.422284in}{0.259864in}}{\pgfqpoint{1.301046in}{1.263466in}}%
\pgfusepath{clip}%
\pgfsetbuttcap%
\pgfsetroundjoin%
\definecolor{currentfill}{rgb}{0.376471,0.376471,0.376471}%
\pgfsetfillcolor{currentfill}%
\pgfsetlinewidth{0.803000pt}%
\definecolor{currentstroke}{rgb}{0.376471,0.376471,0.376471}%
\pgfsetstrokecolor{currentstroke}%
\pgfsetdash{}{0pt}%
\pgfsys@defobject{currentmarker}{\pgfqpoint{-0.027778in}{-0.000000in}}{\pgfqpoint{0.027778in}{0.000000in}}{%
\pgfpathmoveto{\pgfqpoint{0.027778in}{-0.000000in}}%
\pgfpathlineto{\pgfqpoint{-0.027778in}{0.000000in}}%
\pgfusepath{stroke,fill}%
}%
\begin{pgfscope}%
\pgfsys@transformshift{0.750234in}{0.456920in}%
\pgfsys@useobject{currentmarker}{}%
\end{pgfscope}%
\begin{pgfscope}%
\pgfsys@transformshift{1.180332in}{0.554061in}%
\pgfsys@useobject{currentmarker}{}%
\end{pgfscope}%
\begin{pgfscope}%
\pgfsys@transformshift{1.610429in}{1.110948in}%
\pgfsys@useobject{currentmarker}{}%
\end{pgfscope}%
\end{pgfscope}%
\begin{pgfscope}%
\pgfpathrectangle{\pgfqpoint{0.422284in}{0.259864in}}{\pgfqpoint{1.301046in}{1.263466in}}%
\pgfusepath{clip}%
\pgfsetbuttcap%
\pgfsetroundjoin%
\definecolor{currentfill}{rgb}{0.376471,0.376471,0.376471}%
\pgfsetfillcolor{currentfill}%
\pgfsetlinewidth{0.803000pt}%
\definecolor{currentstroke}{rgb}{0.376471,0.376471,0.376471}%
\pgfsetstrokecolor{currentstroke}%
\pgfsetdash{}{0pt}%
\pgfsys@defobject{currentmarker}{\pgfqpoint{-0.027778in}{-0.000000in}}{\pgfqpoint{0.027778in}{0.000000in}}{%
\pgfpathmoveto{\pgfqpoint{0.027778in}{-0.000000in}}%
\pgfpathlineto{\pgfqpoint{-0.027778in}{0.000000in}}%
\pgfusepath{stroke,fill}%
}%
\begin{pgfscope}%
\pgfsys@transformshift{0.750234in}{0.486451in}%
\pgfsys@useobject{currentmarker}{}%
\end{pgfscope}%
\begin{pgfscope}%
\pgfsys@transformshift{1.180332in}{0.561148in}%
\pgfsys@useobject{currentmarker}{}%
\end{pgfscope}%
\begin{pgfscope}%
\pgfsys@transformshift{1.610429in}{1.386140in}%
\pgfsys@useobject{currentmarker}{}%
\end{pgfscope}%
\end{pgfscope}%
\begin{pgfscope}%
\pgfsetrectcap%
\pgfsetmiterjoin%
\pgfsetlinewidth{0.803000pt}%
\definecolor{currentstroke}{rgb}{0.000000,0.000000,0.000000}%
\pgfsetstrokecolor{currentstroke}%
\pgfsetdash{}{0pt}%
\pgfpathmoveto{\pgfqpoint{0.422284in}{0.259864in}}%
\pgfpathlineto{\pgfqpoint{0.422284in}{1.523330in}}%
\pgfusepath{stroke}%
\end{pgfscope}%
\begin{pgfscope}%
\pgfsetrectcap%
\pgfsetmiterjoin%
\pgfsetlinewidth{0.803000pt}%
\definecolor{currentstroke}{rgb}{0.000000,0.000000,0.000000}%
\pgfsetstrokecolor{currentstroke}%
\pgfsetdash{}{0pt}%
\pgfpathmoveto{\pgfqpoint{1.723330in}{0.259864in}}%
\pgfpathlineto{\pgfqpoint{1.723330in}{1.523330in}}%
\pgfusepath{stroke}%
\end{pgfscope}%
\begin{pgfscope}%
\pgfsetrectcap%
\pgfsetmiterjoin%
\pgfsetlinewidth{0.803000pt}%
\definecolor{currentstroke}{rgb}{0.000000,0.000000,0.000000}%
\pgfsetstrokecolor{currentstroke}%
\pgfsetdash{}{0pt}%
\pgfpathmoveto{\pgfqpoint{0.422284in}{0.259864in}}%
\pgfpathlineto{\pgfqpoint{1.723330in}{0.259864in}}%
\pgfusepath{stroke}%
\end{pgfscope}%
\begin{pgfscope}%
\pgfsetrectcap%
\pgfsetmiterjoin%
\pgfsetlinewidth{0.803000pt}%
\definecolor{currentstroke}{rgb}{0.000000,0.000000,0.000000}%
\pgfsetstrokecolor{currentstroke}%
\pgfsetdash{}{0pt}%
\pgfpathmoveto{\pgfqpoint{0.422284in}{1.523330in}}%
\pgfpathlineto{\pgfqpoint{1.723330in}{1.523330in}}%
\pgfusepath{stroke}%
\end{pgfscope}%
\begin{pgfscope}%
\pgfsetbuttcap%
\pgfsetmiterjoin%
\definecolor{currentfill}{rgb}{0.121569,0.466667,0.705882}%
\pgfsetfillcolor{currentfill}%
\pgfsetfillopacity{0.500000}%
\pgfsetlinewidth{0.501875pt}%
\definecolor{currentstroke}{rgb}{0.376471,0.376471,0.376471}%
\pgfsetstrokecolor{currentstroke}%
\pgfsetdash{}{0pt}%
\pgfpathmoveto{\pgfqpoint{0.484784in}{1.365859in}}%
\pgfpathlineto{\pgfqpoint{0.584784in}{1.365859in}}%
\pgfpathlineto{\pgfqpoint{0.584784in}{1.453359in}}%
\pgfpathlineto{\pgfqpoint{0.484784in}{1.453359in}}%
\pgfpathclose%
\pgfusepath{stroke,fill}%
\end{pgfscope}%
\begin{pgfscope}%
\definecolor{textcolor}{rgb}{0.000000,0.000000,0.000000}%
\pgfsetstrokecolor{textcolor}%
\pgfsetfillcolor{textcolor}%
\pgftext[x=0.634784in,y=1.365859in,left,base]{\color{textcolor}\rmfamily\fontsize{9.000000}{10.800000}\selectfont \(\displaystyle d=10\)}%
\end{pgfscope}%
\begin{pgfscope}%
\pgfsetbuttcap%
\pgfsetmiterjoin%
\definecolor{currentfill}{rgb}{1.000000,0.498039,0.054902}%
\pgfsetfillcolor{currentfill}%
\pgfsetfillopacity{0.500000}%
\pgfsetlinewidth{0.501875pt}%
\definecolor{currentstroke}{rgb}{0.376471,0.376471,0.376471}%
\pgfsetstrokecolor{currentstroke}%
\pgfsetdash{}{0pt}%
\pgfpathmoveto{\pgfqpoint{0.484784in}{1.182388in}}%
\pgfpathlineto{\pgfqpoint{0.584784in}{1.182388in}}%
\pgfpathlineto{\pgfqpoint{0.584784in}{1.269888in}}%
\pgfpathlineto{\pgfqpoint{0.484784in}{1.269888in}}%
\pgfpathclose%
\pgfusepath{stroke,fill}%
\end{pgfscope}%
\begin{pgfscope}%
\definecolor{textcolor}{rgb}{0.000000,0.000000,0.000000}%
\pgfsetstrokecolor{textcolor}%
\pgfsetfillcolor{textcolor}%
\pgftext[x=0.634784in,y=1.182388in,left,base]{\color{textcolor}\rmfamily\fontsize{9.000000}{10.800000}\selectfont \(\displaystyle d=100\)}%
\end{pgfscope}%
\begin{pgfscope}%
\pgfsetbuttcap%
\pgfsetmiterjoin%
\definecolor{currentfill}{rgb}{0.839216,0.152941,0.156863}%
\pgfsetfillcolor{currentfill}%
\pgfsetfillopacity{0.500000}%
\pgfsetlinewidth{0.501875pt}%
\definecolor{currentstroke}{rgb}{0.376471,0.376471,0.376471}%
\pgfsetstrokecolor{currentstroke}%
\pgfsetdash{}{0pt}%
\pgfpathmoveto{\pgfqpoint{0.484784in}{0.998916in}}%
\pgfpathlineto{\pgfqpoint{0.584784in}{0.998916in}}%
\pgfpathlineto{\pgfqpoint{0.584784in}{1.086416in}}%
\pgfpathlineto{\pgfqpoint{0.484784in}{1.086416in}}%
\pgfpathclose%
\pgfusepath{stroke,fill}%
\end{pgfscope}%
\begin{pgfscope}%
\definecolor{textcolor}{rgb}{0.000000,0.000000,0.000000}%
\pgfsetstrokecolor{textcolor}%
\pgfsetfillcolor{textcolor}%
\pgftext[x=0.634784in,y=0.998916in,left,base]{\color{textcolor}\rmfamily\fontsize{9.000000}{10.800000}\selectfont \(\displaystyle d=1000\)}%
\end{pgfscope}%
\end{pgfpicture}%
\makeatother%
\endgroup%

%% file: figures/exp_label_proportions.pgf
%% Creator: Matplotlib, PGF backend
%%
%% To include the figure in your LaTeX document, write
%%   \input{<filename>.pgf}
%%
%% Make sure the required packages are loaded in your preamble
%%   \usepackage{pgf}
%%
%% Figures using additional raster images can only be included by \input if
%% they are in the same directory as the main LaTeX file. For loading figures
%% from other directories you can use the `import` package
%% and then include the figures with
%%   \import{<path to file>}{<filename>.pgf}
%%
%% Matplotlib used the following preamble
%%   \usepackage{times}
%%   \usepackage{amsmath}
%%   \usepackage{amssymb}
%%   \usepackage{bm}
%%   \usepackage{fontspec}
%%   \setmainfont{DejaVuSerif.ttf}[Path=/home/vlad/conda/envs/main/lib/python3.7/site-packages/matplotlib/mpl-data/fonts/ttf/]
%%   \setsansfont{DejaVuSans.ttf}[Path=/home/vlad/conda/envs/main/lib/python3.7/site-packages/matplotlib/mpl-data/fonts/ttf/]
%%   \setmonofont{DejaVuSansMono.ttf}[Path=/home/vlad/conda/envs/main/lib/python3.7/site-packages/matplotlib/mpl-data/fonts/ttf/]
%%
\begingroup%
\makeatletter%
\begin{pgfpicture}%
\pgfpathrectangle{\pgfpointorigin}{\pgfqpoint{7.000000in}{2.300000in}}%
\pgfusepath{use as bounding box, clip}%
\begin{pgfscope}%
\pgfsetbuttcap%
\pgfsetmiterjoin%
\definecolor{currentfill}{rgb}{1.000000,1.000000,1.000000}%
\pgfsetfillcolor{currentfill}%
\pgfsetlinewidth{0.000000pt}%
\definecolor{currentstroke}{rgb}{1.000000,1.000000,1.000000}%
\pgfsetstrokecolor{currentstroke}%
\pgfsetdash{}{0pt}%
\pgfpathmoveto{\pgfqpoint{0.000000in}{0.000000in}}%
\pgfpathlineto{\pgfqpoint{7.000000in}{0.000000in}}%
\pgfpathlineto{\pgfqpoint{7.000000in}{2.300000in}}%
\pgfpathlineto{\pgfqpoint{0.000000in}{2.300000in}}%
\pgfpathclose%
\pgfusepath{fill}%
\end{pgfscope}%
\begin{pgfscope}%
\pgfsetbuttcap%
\pgfsetmiterjoin%
\definecolor{currentfill}{rgb}{1.000000,1.000000,1.000000}%
\pgfsetfillcolor{currentfill}%
\pgfsetlinewidth{0.000000pt}%
\definecolor{currentstroke}{rgb}{0.000000,0.000000,0.000000}%
\pgfsetstrokecolor{currentstroke}%
\pgfsetstrokeopacity{0.000000}%
\pgfsetdash{}{0pt}%
\pgfpathmoveto{\pgfqpoint{0.875000in}{0.391000in}}%
\pgfpathlineto{\pgfqpoint{3.340909in}{0.391000in}}%
\pgfpathlineto{\pgfqpoint{3.340909in}{2.024000in}}%
\pgfpathlineto{\pgfqpoint{0.875000in}{2.024000in}}%
\pgfpathclose%
\pgfusepath{fill}%
\end{pgfscope}%
\begin{pgfscope}%
\pgfsetbuttcap%
\pgfsetroundjoin%
\definecolor{currentfill}{rgb}{0.000000,0.000000,0.000000}%
\pgfsetfillcolor{currentfill}%
\pgfsetlinewidth{0.803000pt}%
\definecolor{currentstroke}{rgb}{0.000000,0.000000,0.000000}%
\pgfsetstrokecolor{currentstroke}%
\pgfsetdash{}{0pt}%
\pgfsys@defobject{currentmarker}{\pgfqpoint{0.000000in}{-0.048611in}}{\pgfqpoint{0.000000in}{0.000000in}}{%
\pgfpathmoveto{\pgfqpoint{0.000000in}{0.000000in}}%
\pgfpathlineto{\pgfqpoint{0.000000in}{-0.048611in}}%
\pgfusepath{stroke,fill}%
}%
\begin{pgfscope}%
\pgfsys@transformshift{1.360709in}{0.391000in}%
\pgfsys@useobject{currentmarker}{}%
\end{pgfscope}%
\end{pgfscope}%
\begin{pgfscope}%
\definecolor{textcolor}{rgb}{0.000000,0.000000,0.000000}%
\pgfsetstrokecolor{textcolor}%
\pgfsetfillcolor{textcolor}%
\pgftext[x=1.360709in,y=0.293778in,,top]{\color{textcolor}\rmfamily\fontsize{9.000000}{10.800000}\selectfont \(\displaystyle 500\)}%
\end{pgfscope}%
\begin{pgfscope}%
\pgfsetbuttcap%
\pgfsetroundjoin%
\definecolor{currentfill}{rgb}{0.000000,0.000000,0.000000}%
\pgfsetfillcolor{currentfill}%
\pgfsetlinewidth{0.803000pt}%
\definecolor{currentstroke}{rgb}{0.000000,0.000000,0.000000}%
\pgfsetstrokecolor{currentstroke}%
\pgfsetdash{}{0pt}%
\pgfsys@defobject{currentmarker}{\pgfqpoint{0.000000in}{-0.048611in}}{\pgfqpoint{0.000000in}{0.000000in}}{%
\pgfpathmoveto{\pgfqpoint{0.000000in}{0.000000in}}%
\pgfpathlineto{\pgfqpoint{0.000000in}{-0.048611in}}%
\pgfusepath{stroke,fill}%
}%
\begin{pgfscope}%
\pgfsys@transformshift{1.983414in}{0.391000in}%
\pgfsys@useobject{currentmarker}{}%
\end{pgfscope}%
\end{pgfscope}%
\begin{pgfscope}%
\definecolor{textcolor}{rgb}{0.000000,0.000000,0.000000}%
\pgfsetstrokecolor{textcolor}%
\pgfsetfillcolor{textcolor}%
\pgftext[x=1.983414in,y=0.293778in,,top]{\color{textcolor}\rmfamily\fontsize{9.000000}{10.800000}\selectfont \(\displaystyle 1000\)}%
\end{pgfscope}%
\begin{pgfscope}%
\pgfsetbuttcap%
\pgfsetroundjoin%
\definecolor{currentfill}{rgb}{0.000000,0.000000,0.000000}%
\pgfsetfillcolor{currentfill}%
\pgfsetlinewidth{0.803000pt}%
\definecolor{currentstroke}{rgb}{0.000000,0.000000,0.000000}%
\pgfsetstrokecolor{currentstroke}%
\pgfsetdash{}{0pt}%
\pgfsys@defobject{currentmarker}{\pgfqpoint{0.000000in}{-0.048611in}}{\pgfqpoint{0.000000in}{0.000000in}}{%
\pgfpathmoveto{\pgfqpoint{0.000000in}{0.000000in}}%
\pgfpathlineto{\pgfqpoint{0.000000in}{-0.048611in}}%
\pgfusepath{stroke,fill}%
}%
\begin{pgfscope}%
\pgfsys@transformshift{2.606118in}{0.391000in}%
\pgfsys@useobject{currentmarker}{}%
\end{pgfscope}%
\end{pgfscope}%
\begin{pgfscope}%
\definecolor{textcolor}{rgb}{0.000000,0.000000,0.000000}%
\pgfsetstrokecolor{textcolor}%
\pgfsetfillcolor{textcolor}%
\pgftext[x=2.606118in,y=0.293778in,,top]{\color{textcolor}\rmfamily\fontsize{9.000000}{10.800000}\selectfont \(\displaystyle 1500\)}%
\end{pgfscope}%
\begin{pgfscope}%
\pgfsetbuttcap%
\pgfsetroundjoin%
\definecolor{currentfill}{rgb}{0.000000,0.000000,0.000000}%
\pgfsetfillcolor{currentfill}%
\pgfsetlinewidth{0.803000pt}%
\definecolor{currentstroke}{rgb}{0.000000,0.000000,0.000000}%
\pgfsetstrokecolor{currentstroke}%
\pgfsetdash{}{0pt}%
\pgfsys@defobject{currentmarker}{\pgfqpoint{0.000000in}{-0.048611in}}{\pgfqpoint{0.000000in}{0.000000in}}{%
\pgfpathmoveto{\pgfqpoint{0.000000in}{0.000000in}}%
\pgfpathlineto{\pgfqpoint{0.000000in}{-0.048611in}}%
\pgfusepath{stroke,fill}%
}%
\begin{pgfscope}%
\pgfsys@transformshift{3.228822in}{0.391000in}%
\pgfsys@useobject{currentmarker}{}%
\end{pgfscope}%
\end{pgfscope}%
\begin{pgfscope}%
\definecolor{textcolor}{rgb}{0.000000,0.000000,0.000000}%
\pgfsetstrokecolor{textcolor}%
\pgfsetfillcolor{textcolor}%
\pgftext[x=3.228822in,y=0.293778in,,top]{\color{textcolor}\rmfamily\fontsize{9.000000}{10.800000}\selectfont \(\displaystyle 2000\)}%
\end{pgfscope}%
\begin{pgfscope}%
\definecolor{textcolor}{rgb}{0.000000,0.000000,0.000000}%
\pgfsetstrokecolor{textcolor}%
\pgfsetfillcolor{textcolor}%
\pgftext[x=2.107955in,y=0.117251in,,top]{\color{textcolor}\rmfamily\fontsize{9.000000}{10.800000}\selectfont Document length}%
\end{pgfscope}%
\begin{pgfscope}%
\pgfsetbuttcap%
\pgfsetroundjoin%
\definecolor{currentfill}{rgb}{0.000000,0.000000,0.000000}%
\pgfsetfillcolor{currentfill}%
\pgfsetlinewidth{0.803000pt}%
\definecolor{currentstroke}{rgb}{0.000000,0.000000,0.000000}%
\pgfsetstrokecolor{currentstroke}%
\pgfsetdash{}{0pt}%
\pgfsys@defobject{currentmarker}{\pgfqpoint{-0.048611in}{0.000000in}}{\pgfqpoint{0.000000in}{0.000000in}}{%
\pgfpathmoveto{\pgfqpoint{0.000000in}{0.000000in}}%
\pgfpathlineto{\pgfqpoint{-0.048611in}{0.000000in}}%
\pgfusepath{stroke,fill}%
}%
\begin{pgfscope}%
\pgfsys@transformshift{0.875000in}{0.709922in}%
\pgfsys@useobject{currentmarker}{}%
\end{pgfscope}%
\end{pgfscope}%
\begin{pgfscope}%
\definecolor{textcolor}{rgb}{0.000000,0.000000,0.000000}%
\pgfsetstrokecolor{textcolor}%
\pgfsetfillcolor{textcolor}%
\pgftext[x=0.549384in,y=0.662437in,left,base]{\color{textcolor}\rmfamily\fontsize{9.000000}{10.800000}\selectfont \(\displaystyle 0.02\)}%
\end{pgfscope}%
\begin{pgfscope}%
\pgfsetbuttcap%
\pgfsetroundjoin%
\definecolor{currentfill}{rgb}{0.000000,0.000000,0.000000}%
\pgfsetfillcolor{currentfill}%
\pgfsetlinewidth{0.803000pt}%
\definecolor{currentstroke}{rgb}{0.000000,0.000000,0.000000}%
\pgfsetstrokecolor{currentstroke}%
\pgfsetdash{}{0pt}%
\pgfsys@defobject{currentmarker}{\pgfqpoint{-0.048611in}{0.000000in}}{\pgfqpoint{0.000000in}{0.000000in}}{%
\pgfpathmoveto{\pgfqpoint{0.000000in}{0.000000in}}%
\pgfpathlineto{\pgfqpoint{-0.048611in}{0.000000in}}%
\pgfusepath{stroke,fill}%
}%
\begin{pgfscope}%
\pgfsys@transformshift{0.875000in}{1.086244in}%
\pgfsys@useobject{currentmarker}{}%
\end{pgfscope}%
\end{pgfscope}%
\begin{pgfscope}%
\definecolor{textcolor}{rgb}{0.000000,0.000000,0.000000}%
\pgfsetstrokecolor{textcolor}%
\pgfsetfillcolor{textcolor}%
\pgftext[x=0.549384in,y=1.038759in,left,base]{\color{textcolor}\rmfamily\fontsize{9.000000}{10.800000}\selectfont \(\displaystyle 0.03\)}%
\end{pgfscope}%
\begin{pgfscope}%
\pgfsetbuttcap%
\pgfsetroundjoin%
\definecolor{currentfill}{rgb}{0.000000,0.000000,0.000000}%
\pgfsetfillcolor{currentfill}%
\pgfsetlinewidth{0.803000pt}%
\definecolor{currentstroke}{rgb}{0.000000,0.000000,0.000000}%
\pgfsetstrokecolor{currentstroke}%
\pgfsetdash{}{0pt}%
\pgfsys@defobject{currentmarker}{\pgfqpoint{-0.048611in}{0.000000in}}{\pgfqpoint{0.000000in}{0.000000in}}{%
\pgfpathmoveto{\pgfqpoint{0.000000in}{0.000000in}}%
\pgfpathlineto{\pgfqpoint{-0.048611in}{0.000000in}}%
\pgfusepath{stroke,fill}%
}%
\begin{pgfscope}%
\pgfsys@transformshift{0.875000in}{1.462567in}%
\pgfsys@useobject{currentmarker}{}%
\end{pgfscope}%
\end{pgfscope}%
\begin{pgfscope}%
\definecolor{textcolor}{rgb}{0.000000,0.000000,0.000000}%
\pgfsetstrokecolor{textcolor}%
\pgfsetfillcolor{textcolor}%
\pgftext[x=0.549384in,y=1.415081in,left,base]{\color{textcolor}\rmfamily\fontsize{9.000000}{10.800000}\selectfont \(\displaystyle 0.04\)}%
\end{pgfscope}%
\begin{pgfscope}%
\pgfsetbuttcap%
\pgfsetroundjoin%
\definecolor{currentfill}{rgb}{0.000000,0.000000,0.000000}%
\pgfsetfillcolor{currentfill}%
\pgfsetlinewidth{0.803000pt}%
\definecolor{currentstroke}{rgb}{0.000000,0.000000,0.000000}%
\pgfsetstrokecolor{currentstroke}%
\pgfsetdash{}{0pt}%
\pgfsys@defobject{currentmarker}{\pgfqpoint{-0.048611in}{0.000000in}}{\pgfqpoint{0.000000in}{0.000000in}}{%
\pgfpathmoveto{\pgfqpoint{0.000000in}{0.000000in}}%
\pgfpathlineto{\pgfqpoint{-0.048611in}{0.000000in}}%
\pgfusepath{stroke,fill}%
}%
\begin{pgfscope}%
\pgfsys@transformshift{0.875000in}{1.838889in}%
\pgfsys@useobject{currentmarker}{}%
\end{pgfscope}%
\end{pgfscope}%
\begin{pgfscope}%
\definecolor{textcolor}{rgb}{0.000000,0.000000,0.000000}%
\pgfsetstrokecolor{textcolor}%
\pgfsetfillcolor{textcolor}%
\pgftext[x=0.549384in,y=1.791404in,left,base]{\color{textcolor}\rmfamily\fontsize{9.000000}{10.800000}\selectfont \(\displaystyle 0.05\)}%
\end{pgfscope}%
\begin{pgfscope}%
\definecolor{textcolor}{rgb}{0.000000,0.000000,0.000000}%
\pgfsetstrokecolor{textcolor}%
\pgfsetfillcolor{textcolor}%
\pgftext[x=0.493829in,y=1.207500in,,bottom,rotate=90.000000]{\color{textcolor}\rmfamily\fontsize{9.000000}{10.800000}\selectfont Jensen-Shannon divergence}%
\end{pgfscope}%
\begin{pgfscope}%
\pgfpathrectangle{\pgfqpoint{0.875000in}{0.391000in}}{\pgfqpoint{2.465909in}{1.633000in}}%
\pgfusepath{clip}%
\pgfsetbuttcap%
\pgfsetroundjoin%
\pgfsetlinewidth{2.007500pt}%
\definecolor{currentstroke}{rgb}{0.172549,0.627451,0.172549}%
\pgfsetstrokecolor{currentstroke}%
\pgfsetdash{{2.000000pt}{0.000000pt}}{0.000000pt}%
\pgfpathmoveto{\pgfqpoint{0.987087in}{1.949773in}}%
\pgfpathlineto{\pgfqpoint{1.236169in}{1.475761in}}%
\pgfpathlineto{\pgfqpoint{1.485250in}{1.362412in}}%
\pgfpathlineto{\pgfqpoint{1.734332in}{1.249957in}}%
\pgfpathlineto{\pgfqpoint{1.983414in}{1.208192in}}%
\pgfpathlineto{\pgfqpoint{2.232495in}{1.092942in}}%
\pgfpathlineto{\pgfqpoint{2.481577in}{1.016323in}}%
\pgfpathlineto{\pgfqpoint{2.730659in}{1.084629in}}%
\pgfpathlineto{\pgfqpoint{2.979741in}{1.075282in}}%
\pgfpathlineto{\pgfqpoint{3.228822in}{1.043446in}}%
\pgfusepath{stroke}%
\end{pgfscope}%
\begin{pgfscope}%
\pgfpathrectangle{\pgfqpoint{0.875000in}{0.391000in}}{\pgfqpoint{2.465909in}{1.633000in}}%
\pgfusepath{clip}%
\pgfsetbuttcap%
\pgfsetroundjoin%
\pgfsetlinewidth{3.011250pt}%
\definecolor{currentstroke}{rgb}{0.890196,0.466667,0.760784}%
\pgfsetstrokecolor{currentstroke}%
\pgfsetdash{{3.000000pt}{0.000000pt}}{0.000000pt}%
\pgfpathmoveto{\pgfqpoint{0.987087in}{1.871439in}}%
\pgfpathlineto{\pgfqpoint{1.236169in}{1.117148in}}%
\pgfpathlineto{\pgfqpoint{1.485250in}{0.940459in}}%
\pgfpathlineto{\pgfqpoint{1.734332in}{0.844887in}}%
\pgfpathlineto{\pgfqpoint{1.983414in}{0.758993in}}%
\pgfpathlineto{\pgfqpoint{2.232495in}{0.637207in}}%
\pgfpathlineto{\pgfqpoint{2.481577in}{0.526600in}}%
\pgfpathlineto{\pgfqpoint{2.730659in}{0.600381in}}%
\pgfpathlineto{\pgfqpoint{2.979741in}{0.546320in}}%
\pgfpathlineto{\pgfqpoint{3.228822in}{0.465227in}}%
\pgfusepath{stroke}%
\end{pgfscope}%
\begin{pgfscope}%
\pgfpathrectangle{\pgfqpoint{0.875000in}{0.391000in}}{\pgfqpoint{2.465909in}{1.633000in}}%
\pgfusepath{clip}%
\pgfsetbuttcap%
\pgfsetroundjoin%
\pgfsetlinewidth{3.011250pt}%
\definecolor{currentstroke}{rgb}{0.737255,0.741176,0.133333}%
\pgfsetstrokecolor{currentstroke}%
\pgfsetdash{{6.000000pt}{6.000000pt}}{0.000000pt}%
\pgfpathmoveto{\pgfqpoint{0.987087in}{1.871524in}}%
\pgfpathlineto{\pgfqpoint{1.236169in}{1.113921in}}%
\pgfpathlineto{\pgfqpoint{1.485250in}{0.940954in}}%
\pgfpathlineto{\pgfqpoint{1.734332in}{0.846611in}}%
\pgfpathlineto{\pgfqpoint{1.983414in}{0.755437in}}%
\pgfpathlineto{\pgfqpoint{2.232495in}{0.635461in}}%
\pgfpathlineto{\pgfqpoint{2.481577in}{0.524289in}}%
\pgfpathlineto{\pgfqpoint{2.730659in}{0.595685in}}%
\pgfpathlineto{\pgfqpoint{2.979741in}{0.542503in}}%
\pgfpathlineto{\pgfqpoint{3.228822in}{0.467041in}}%
\pgfusepath{stroke}%
\end{pgfscope}%
\begin{pgfscope}%
\pgfpathrectangle{\pgfqpoint{0.875000in}{0.391000in}}{\pgfqpoint{2.465909in}{1.633000in}}%
\pgfusepath{clip}%
\pgfsetbuttcap%
\pgfsetroundjoin%
\pgfsetlinewidth{2.007500pt}%
\definecolor{currentstroke}{rgb}{0.000000,0.000000,0.000000}%
\pgfsetstrokecolor{currentstroke}%
\pgfsetdash{{2.000000pt}{6.000000pt}}{0.000000pt}%
\pgfpathmoveto{\pgfqpoint{0.987087in}{1.914989in}}%
\pgfpathlineto{\pgfqpoint{1.236169in}{1.113921in}}%
\pgfpathlineto{\pgfqpoint{1.485250in}{0.940954in}}%
\pgfpathlineto{\pgfqpoint{1.734332in}{0.846611in}}%
\pgfpathlineto{\pgfqpoint{1.983414in}{0.755437in}}%
\pgfpathlineto{\pgfqpoint{2.232495in}{0.635461in}}%
\pgfpathlineto{\pgfqpoint{2.481577in}{0.524289in}}%
\pgfpathlineto{\pgfqpoint{2.730659in}{0.595685in}}%
\pgfpathlineto{\pgfqpoint{2.979741in}{0.542503in}}%
\pgfpathlineto{\pgfqpoint{3.228822in}{0.467041in}}%
\pgfusepath{stroke}%
\end{pgfscope}%
\begin{pgfscope}%
\pgfsetrectcap%
\pgfsetmiterjoin%
\pgfsetlinewidth{0.803000pt}%
\definecolor{currentstroke}{rgb}{0.000000,0.000000,0.000000}%
\pgfsetstrokecolor{currentstroke}%
\pgfsetdash{}{0pt}%
\pgfpathmoveto{\pgfqpoint{0.875000in}{0.391000in}}%
\pgfpathlineto{\pgfqpoint{0.875000in}{2.024000in}}%
\pgfusepath{stroke}%
\end{pgfscope}%
\begin{pgfscope}%
\pgfsetrectcap%
\pgfsetmiterjoin%
\pgfsetlinewidth{0.803000pt}%
\definecolor{currentstroke}{rgb}{0.000000,0.000000,0.000000}%
\pgfsetstrokecolor{currentstroke}%
\pgfsetdash{}{0pt}%
\pgfpathmoveto{\pgfqpoint{3.340909in}{0.391000in}}%
\pgfpathlineto{\pgfqpoint{3.340909in}{2.024000in}}%
\pgfusepath{stroke}%
\end{pgfscope}%
\begin{pgfscope}%
\pgfsetrectcap%
\pgfsetmiterjoin%
\pgfsetlinewidth{0.803000pt}%
\definecolor{currentstroke}{rgb}{0.000000,0.000000,0.000000}%
\pgfsetstrokecolor{currentstroke}%
\pgfsetdash{}{0pt}%
\pgfpathmoveto{\pgfqpoint{0.875000in}{0.391000in}}%
\pgfpathlineto{\pgfqpoint{3.340909in}{0.391000in}}%
\pgfusepath{stroke}%
\end{pgfscope}%
\begin{pgfscope}%
\pgfsetrectcap%
\pgfsetmiterjoin%
\pgfsetlinewidth{0.803000pt}%
\definecolor{currentstroke}{rgb}{0.000000,0.000000,0.000000}%
\pgfsetstrokecolor{currentstroke}%
\pgfsetdash{}{0pt}%
\pgfpathmoveto{\pgfqpoint{0.875000in}{2.024000in}}%
\pgfpathlineto{\pgfqpoint{3.340909in}{2.024000in}}%
\pgfusepath{stroke}%
\end{pgfscope}%
\begin{pgfscope}%
\definecolor{textcolor}{rgb}{0.000000,0.000000,0.000000}%
\pgfsetstrokecolor{textcolor}%
\pgfsetfillcolor{textcolor}%
\pgftext[x=2.107955in,y=2.107333in,,base]{\color{textcolor}\rmfamily\fontsize{10.800000}{12.960000}\selectfont 10 classes}%
\end{pgfscope}%
\begin{pgfscope}%
\pgfsetbuttcap%
\pgfsetroundjoin%
\pgfsetlinewidth{2.007500pt}%
\definecolor{currentstroke}{rgb}{0.172549,0.627451,0.172549}%
\pgfsetstrokecolor{currentstroke}%
\pgfsetdash{{2.000000pt}{0.000000pt}}{0.000000pt}%
\pgfpathmoveto{\pgfqpoint{1.891742in}{1.922779in}}%
\pgfpathlineto{\pgfqpoint{2.266742in}{1.922779in}}%
\pgfusepath{stroke}%
\end{pgfscope}%
\begin{pgfscope}%
\definecolor{textcolor}{rgb}{0.000000,0.000000,0.000000}%
\pgfsetstrokecolor{textcolor}%
\pgfsetfillcolor{textcolor}%
\pgftext[x=2.366742in,y=1.879029in,left,base]{\color{textcolor}\rmfamily\fontsize{9.000000}{10.800000}\selectfont Logistic}%
\end{pgfscope}%
\begin{pgfscope}%
\pgfsetbuttcap%
\pgfsetroundjoin%
\pgfsetlinewidth{3.011250pt}%
\definecolor{currentstroke}{rgb}{0.890196,0.466667,0.760784}%
\pgfsetstrokecolor{currentstroke}%
\pgfsetdash{{3.000000pt}{0.000000pt}}{0.000000pt}%
\pgfpathmoveto{\pgfqpoint{1.891742in}{1.737538in}}%
\pgfpathlineto{\pgfqpoint{2.266742in}{1.737538in}}%
\pgfusepath{stroke}%
\end{pgfscope}%
\begin{pgfscope}%
\definecolor{textcolor}{rgb}{0.000000,0.000000,0.000000}%
\pgfsetstrokecolor{textcolor}%
\pgfsetfillcolor{textcolor}%
\pgftext[x=2.366742in,y=1.693788in,left,base]{\color{textcolor}\rmfamily\fontsize{9.000000}{10.800000}\selectfont Sparsemax}%
\end{pgfscope}%
\begin{pgfscope}%
\pgfsetbuttcap%
\pgfsetroundjoin%
\pgfsetlinewidth{3.011250pt}%
\definecolor{currentstroke}{rgb}{0.737255,0.741176,0.133333}%
\pgfsetstrokecolor{currentstroke}%
\pgfsetdash{{6.000000pt}{6.000000pt}}{0.000000pt}%
\pgfpathmoveto{\pgfqpoint{1.891742in}{1.554066in}}%
\pgfpathlineto{\pgfqpoint{2.266742in}{1.554066in}}%
\pgfusepath{stroke}%
\end{pgfscope}%
\begin{pgfscope}%
\definecolor{textcolor}{rgb}{0.000000,0.000000,0.000000}%
\pgfsetstrokecolor{textcolor}%
\pgfsetfillcolor{textcolor}%
\pgftext[x=2.366742in,y=1.510316in,left,base]{\color{textcolor}\rmfamily\fontsize{9.000000}{10.800000}\selectfont Tsallis \(\displaystyle \alpha=1.5\)}%
\end{pgfscope}%
\begin{pgfscope}%
\pgfsetbuttcap%
\pgfsetroundjoin%
\pgfsetlinewidth{2.007500pt}%
\definecolor{currentstroke}{rgb}{0.000000,0.000000,0.000000}%
\pgfsetstrokecolor{currentstroke}%
\pgfsetdash{{2.000000pt}{6.000000pt}}{0.000000pt}%
\pgfpathmoveto{\pgfqpoint{1.891742in}{1.370595in}}%
\pgfpathlineto{\pgfqpoint{2.266742in}{1.370595in}}%
\pgfusepath{stroke}%
\end{pgfscope}%
\begin{pgfscope}%
\definecolor{textcolor}{rgb}{0.000000,0.000000,0.000000}%
\pgfsetstrokecolor{textcolor}%
\pgfsetfillcolor{textcolor}%
\pgftext[x=2.366742in,y=1.326845in,left,base]{\color{textcolor}\rmfamily\fontsize{9.000000}{10.800000}\selectfont Tsallis \(\displaystyle \alpha\) tuned}%
\end{pgfscope}%
\begin{pgfscope}%
\pgfsetbuttcap%
\pgfsetmiterjoin%
\definecolor{currentfill}{rgb}{1.000000,1.000000,1.000000}%
\pgfsetfillcolor{currentfill}%
\pgfsetlinewidth{0.000000pt}%
\definecolor{currentstroke}{rgb}{0.000000,0.000000,0.000000}%
\pgfsetstrokecolor{currentstroke}%
\pgfsetstrokeopacity{0.000000}%
\pgfsetdash{}{0pt}%
\pgfpathmoveto{\pgfqpoint{3.834091in}{0.391000in}}%
\pgfpathlineto{\pgfqpoint{6.300000in}{0.391000in}}%
\pgfpathlineto{\pgfqpoint{6.300000in}{2.024000in}}%
\pgfpathlineto{\pgfqpoint{3.834091in}{2.024000in}}%
\pgfpathclose%
\pgfusepath{fill}%
\end{pgfscope}%
\begin{pgfscope}%
\pgfsetbuttcap%
\pgfsetroundjoin%
\definecolor{currentfill}{rgb}{0.000000,0.000000,0.000000}%
\pgfsetfillcolor{currentfill}%
\pgfsetlinewidth{0.803000pt}%
\definecolor{currentstroke}{rgb}{0.000000,0.000000,0.000000}%
\pgfsetstrokecolor{currentstroke}%
\pgfsetdash{}{0pt}%
\pgfsys@defobject{currentmarker}{\pgfqpoint{0.000000in}{-0.048611in}}{\pgfqpoint{0.000000in}{0.000000in}}{%
\pgfpathmoveto{\pgfqpoint{0.000000in}{0.000000in}}%
\pgfpathlineto{\pgfqpoint{0.000000in}{-0.048611in}}%
\pgfusepath{stroke,fill}%
}%
\begin{pgfscope}%
\pgfsys@transformshift{4.319800in}{0.391000in}%
\pgfsys@useobject{currentmarker}{}%
\end{pgfscope}%
\end{pgfscope}%
\begin{pgfscope}%
\definecolor{textcolor}{rgb}{0.000000,0.000000,0.000000}%
\pgfsetstrokecolor{textcolor}%
\pgfsetfillcolor{textcolor}%
\pgftext[x=4.319800in,y=0.293778in,,top]{\color{textcolor}\rmfamily\fontsize{9.000000}{10.800000}\selectfont \(\displaystyle 500\)}%
\end{pgfscope}%
\begin{pgfscope}%
\pgfsetbuttcap%
\pgfsetroundjoin%
\definecolor{currentfill}{rgb}{0.000000,0.000000,0.000000}%
\pgfsetfillcolor{currentfill}%
\pgfsetlinewidth{0.803000pt}%
\definecolor{currentstroke}{rgb}{0.000000,0.000000,0.000000}%
\pgfsetstrokecolor{currentstroke}%
\pgfsetdash{}{0pt}%
\pgfsys@defobject{currentmarker}{\pgfqpoint{0.000000in}{-0.048611in}}{\pgfqpoint{0.000000in}{0.000000in}}{%
\pgfpathmoveto{\pgfqpoint{0.000000in}{0.000000in}}%
\pgfpathlineto{\pgfqpoint{0.000000in}{-0.048611in}}%
\pgfusepath{stroke,fill}%
}%
\begin{pgfscope}%
\pgfsys@transformshift{4.942505in}{0.391000in}%
\pgfsys@useobject{currentmarker}{}%
\end{pgfscope}%
\end{pgfscope}%
\begin{pgfscope}%
\definecolor{textcolor}{rgb}{0.000000,0.000000,0.000000}%
\pgfsetstrokecolor{textcolor}%
\pgfsetfillcolor{textcolor}%
\pgftext[x=4.942505in,y=0.293778in,,top]{\color{textcolor}\rmfamily\fontsize{9.000000}{10.800000}\selectfont \(\displaystyle 1000\)}%
\end{pgfscope}%
\begin{pgfscope}%
\pgfsetbuttcap%
\pgfsetroundjoin%
\definecolor{currentfill}{rgb}{0.000000,0.000000,0.000000}%
\pgfsetfillcolor{currentfill}%
\pgfsetlinewidth{0.803000pt}%
\definecolor{currentstroke}{rgb}{0.000000,0.000000,0.000000}%
\pgfsetstrokecolor{currentstroke}%
\pgfsetdash{}{0pt}%
\pgfsys@defobject{currentmarker}{\pgfqpoint{0.000000in}{-0.048611in}}{\pgfqpoint{0.000000in}{0.000000in}}{%
\pgfpathmoveto{\pgfqpoint{0.000000in}{0.000000in}}%
\pgfpathlineto{\pgfqpoint{0.000000in}{-0.048611in}}%
\pgfusepath{stroke,fill}%
}%
\begin{pgfscope}%
\pgfsys@transformshift{5.565209in}{0.391000in}%
\pgfsys@useobject{currentmarker}{}%
\end{pgfscope}%
\end{pgfscope}%
\begin{pgfscope}%
\definecolor{textcolor}{rgb}{0.000000,0.000000,0.000000}%
\pgfsetstrokecolor{textcolor}%
\pgfsetfillcolor{textcolor}%
\pgftext[x=5.565209in,y=0.293778in,,top]{\color{textcolor}\rmfamily\fontsize{9.000000}{10.800000}\selectfont \(\displaystyle 1500\)}%
\end{pgfscope}%
\begin{pgfscope}%
\pgfsetbuttcap%
\pgfsetroundjoin%
\definecolor{currentfill}{rgb}{0.000000,0.000000,0.000000}%
\pgfsetfillcolor{currentfill}%
\pgfsetlinewidth{0.803000pt}%
\definecolor{currentstroke}{rgb}{0.000000,0.000000,0.000000}%
\pgfsetstrokecolor{currentstroke}%
\pgfsetdash{}{0pt}%
\pgfsys@defobject{currentmarker}{\pgfqpoint{0.000000in}{-0.048611in}}{\pgfqpoint{0.000000in}{0.000000in}}{%
\pgfpathmoveto{\pgfqpoint{0.000000in}{0.000000in}}%
\pgfpathlineto{\pgfqpoint{0.000000in}{-0.048611in}}%
\pgfusepath{stroke,fill}%
}%
\begin{pgfscope}%
\pgfsys@transformshift{6.187913in}{0.391000in}%
\pgfsys@useobject{currentmarker}{}%
\end{pgfscope}%
\end{pgfscope}%
\begin{pgfscope}%
\definecolor{textcolor}{rgb}{0.000000,0.000000,0.000000}%
\pgfsetstrokecolor{textcolor}%
\pgfsetfillcolor{textcolor}%
\pgftext[x=6.187913in,y=0.293778in,,top]{\color{textcolor}\rmfamily\fontsize{9.000000}{10.800000}\selectfont \(\displaystyle 2000\)}%
\end{pgfscope}%
\begin{pgfscope}%
\definecolor{textcolor}{rgb}{0.000000,0.000000,0.000000}%
\pgfsetstrokecolor{textcolor}%
\pgfsetfillcolor{textcolor}%
\pgftext[x=5.067045in,y=0.117251in,,top]{\color{textcolor}\rmfamily\fontsize{9.000000}{10.800000}\selectfont Document length}%
\end{pgfscope}%
\begin{pgfscope}%
\pgfsetbuttcap%
\pgfsetroundjoin%
\definecolor{currentfill}{rgb}{0.000000,0.000000,0.000000}%
\pgfsetfillcolor{currentfill}%
\pgfsetlinewidth{0.803000pt}%
\definecolor{currentstroke}{rgb}{0.000000,0.000000,0.000000}%
\pgfsetstrokecolor{currentstroke}%
\pgfsetdash{}{0pt}%
\pgfsys@defobject{currentmarker}{\pgfqpoint{-0.048611in}{0.000000in}}{\pgfqpoint{0.000000in}{0.000000in}}{%
\pgfpathmoveto{\pgfqpoint{0.000000in}{0.000000in}}%
\pgfpathlineto{\pgfqpoint{-0.048611in}{0.000000in}}%
\pgfusepath{stroke,fill}%
}%
\begin{pgfscope}%
\pgfsys@transformshift{3.834091in}{0.762200in}%
\pgfsys@useobject{currentmarker}{}%
\end{pgfscope}%
\end{pgfscope}%
\begin{pgfscope}%
\definecolor{textcolor}{rgb}{0.000000,0.000000,0.000000}%
\pgfsetstrokecolor{textcolor}%
\pgfsetfillcolor{textcolor}%
\pgftext[x=3.508475in,y=0.714715in,left,base]{\color{textcolor}\rmfamily\fontsize{9.000000}{10.800000}\selectfont \(\displaystyle 0.04\)}%
\end{pgfscope}%
\begin{pgfscope}%
\pgfsetbuttcap%
\pgfsetroundjoin%
\definecolor{currentfill}{rgb}{0.000000,0.000000,0.000000}%
\pgfsetfillcolor{currentfill}%
\pgfsetlinewidth{0.803000pt}%
\definecolor{currentstroke}{rgb}{0.000000,0.000000,0.000000}%
\pgfsetstrokecolor{currentstroke}%
\pgfsetdash{}{0pt}%
\pgfsys@defobject{currentmarker}{\pgfqpoint{-0.048611in}{0.000000in}}{\pgfqpoint{0.000000in}{0.000000in}}{%
\pgfpathmoveto{\pgfqpoint{0.000000in}{0.000000in}}%
\pgfpathlineto{\pgfqpoint{-0.048611in}{0.000000in}}%
\pgfusepath{stroke,fill}%
}%
\begin{pgfscope}%
\pgfsys@transformshift{3.834091in}{1.151000in}%
\pgfsys@useobject{currentmarker}{}%
\end{pgfscope}%
\end{pgfscope}%
\begin{pgfscope}%
\definecolor{textcolor}{rgb}{0.000000,0.000000,0.000000}%
\pgfsetstrokecolor{textcolor}%
\pgfsetfillcolor{textcolor}%
\pgftext[x=3.508475in,y=1.103515in,left,base]{\color{textcolor}\rmfamily\fontsize{9.000000}{10.800000}\selectfont \(\displaystyle 0.06\)}%
\end{pgfscope}%
\begin{pgfscope}%
\pgfsetbuttcap%
\pgfsetroundjoin%
\definecolor{currentfill}{rgb}{0.000000,0.000000,0.000000}%
\pgfsetfillcolor{currentfill}%
\pgfsetlinewidth{0.803000pt}%
\definecolor{currentstroke}{rgb}{0.000000,0.000000,0.000000}%
\pgfsetstrokecolor{currentstroke}%
\pgfsetdash{}{0pt}%
\pgfsys@defobject{currentmarker}{\pgfqpoint{-0.048611in}{0.000000in}}{\pgfqpoint{0.000000in}{0.000000in}}{%
\pgfpathmoveto{\pgfqpoint{0.000000in}{0.000000in}}%
\pgfpathlineto{\pgfqpoint{-0.048611in}{0.000000in}}%
\pgfusepath{stroke,fill}%
}%
\begin{pgfscope}%
\pgfsys@transformshift{3.834091in}{1.539800in}%
\pgfsys@useobject{currentmarker}{}%
\end{pgfscope}%
\end{pgfscope}%
\begin{pgfscope}%
\definecolor{textcolor}{rgb}{0.000000,0.000000,0.000000}%
\pgfsetstrokecolor{textcolor}%
\pgfsetfillcolor{textcolor}%
\pgftext[x=3.508475in,y=1.492314in,left,base]{\color{textcolor}\rmfamily\fontsize{9.000000}{10.800000}\selectfont \(\displaystyle 0.08\)}%
\end{pgfscope}%
\begin{pgfscope}%
\pgfsetbuttcap%
\pgfsetroundjoin%
\definecolor{currentfill}{rgb}{0.000000,0.000000,0.000000}%
\pgfsetfillcolor{currentfill}%
\pgfsetlinewidth{0.803000pt}%
\definecolor{currentstroke}{rgb}{0.000000,0.000000,0.000000}%
\pgfsetstrokecolor{currentstroke}%
\pgfsetdash{}{0pt}%
\pgfsys@defobject{currentmarker}{\pgfqpoint{-0.048611in}{0.000000in}}{\pgfqpoint{0.000000in}{0.000000in}}{%
\pgfpathmoveto{\pgfqpoint{0.000000in}{0.000000in}}%
\pgfpathlineto{\pgfqpoint{-0.048611in}{0.000000in}}%
\pgfusepath{stroke,fill}%
}%
\begin{pgfscope}%
\pgfsys@transformshift{3.834091in}{1.928599in}%
\pgfsys@useobject{currentmarker}{}%
\end{pgfscope}%
\end{pgfscope}%
\begin{pgfscope}%
\definecolor{textcolor}{rgb}{0.000000,0.000000,0.000000}%
\pgfsetstrokecolor{textcolor}%
\pgfsetfillcolor{textcolor}%
\pgftext[x=3.508475in,y=1.881114in,left,base]{\color{textcolor}\rmfamily\fontsize{9.000000}{10.800000}\selectfont \(\displaystyle 0.10\)}%
\end{pgfscope}%
\begin{pgfscope}%
\pgfpathrectangle{\pgfqpoint{3.834091in}{0.391000in}}{\pgfqpoint{2.465909in}{1.633000in}}%
\pgfusepath{clip}%
\pgfsetbuttcap%
\pgfsetroundjoin%
\pgfsetlinewidth{2.007500pt}%
\definecolor{currentstroke}{rgb}{0.172549,0.627451,0.172549}%
\pgfsetstrokecolor{currentstroke}%
\pgfsetdash{{2.000000pt}{0.000000pt}}{0.000000pt}%
\pgfpathmoveto{\pgfqpoint{3.946178in}{1.815716in}}%
\pgfpathlineto{\pgfqpoint{4.195259in}{1.638805in}}%
\pgfpathlineto{\pgfqpoint{4.444341in}{1.518631in}}%
\pgfpathlineto{\pgfqpoint{4.693423in}{1.488123in}}%
\pgfpathlineto{\pgfqpoint{4.942505in}{1.607086in}}%
\pgfpathlineto{\pgfqpoint{5.191586in}{1.582357in}}%
\pgfpathlineto{\pgfqpoint{5.440668in}{1.463668in}}%
\pgfpathlineto{\pgfqpoint{5.689750in}{1.511205in}}%
\pgfpathlineto{\pgfqpoint{5.938831in}{1.592848in}}%
\pgfpathlineto{\pgfqpoint{6.187913in}{1.666170in}}%
\pgfusepath{stroke}%
\end{pgfscope}%
\begin{pgfscope}%
\pgfpathrectangle{\pgfqpoint{3.834091in}{0.391000in}}{\pgfqpoint{2.465909in}{1.633000in}}%
\pgfusepath{clip}%
\pgfsetbuttcap%
\pgfsetroundjoin%
\pgfsetlinewidth{3.011250pt}%
\definecolor{currentstroke}{rgb}{0.890196,0.466667,0.760784}%
\pgfsetstrokecolor{currentstroke}%
\pgfsetdash{{3.000000pt}{0.000000pt}}{0.000000pt}%
\pgfpathmoveto{\pgfqpoint{3.946178in}{1.949773in}}%
\pgfpathlineto{\pgfqpoint{4.195259in}{1.446473in}}%
\pgfpathlineto{\pgfqpoint{4.444341in}{1.122562in}}%
\pgfpathlineto{\pgfqpoint{4.693423in}{0.986451in}}%
\pgfpathlineto{\pgfqpoint{4.942505in}{1.057103in}}%
\pgfpathlineto{\pgfqpoint{5.191586in}{0.891332in}}%
\pgfpathlineto{\pgfqpoint{5.440668in}{0.763776in}}%
\pgfpathlineto{\pgfqpoint{5.689750in}{0.832462in}}%
\pgfpathlineto{\pgfqpoint{5.938831in}{0.743282in}}%
\pgfpathlineto{\pgfqpoint{6.187913in}{0.721975in}}%
\pgfusepath{stroke}%
\end{pgfscope}%
\begin{pgfscope}%
\pgfpathrectangle{\pgfqpoint{3.834091in}{0.391000in}}{\pgfqpoint{2.465909in}{1.633000in}}%
\pgfusepath{clip}%
\pgfsetbuttcap%
\pgfsetroundjoin%
\pgfsetlinewidth{3.011250pt}%
\definecolor{currentstroke}{rgb}{0.737255,0.741176,0.133333}%
\pgfsetstrokecolor{currentstroke}%
\pgfsetdash{{6.000000pt}{6.000000pt}}{0.000000pt}%
\pgfpathmoveto{\pgfqpoint{3.946178in}{1.758670in}}%
\pgfpathlineto{\pgfqpoint{4.195259in}{1.208540in}}%
\pgfpathlineto{\pgfqpoint{4.444341in}{0.929667in}}%
\pgfpathlineto{\pgfqpoint{4.693423in}{0.751309in}}%
\pgfpathlineto{\pgfqpoint{4.942505in}{0.742005in}}%
\pgfpathlineto{\pgfqpoint{5.191586in}{0.663575in}}%
\pgfpathlineto{\pgfqpoint{5.440668in}{0.529677in}}%
\pgfpathlineto{\pgfqpoint{5.689750in}{0.548809in}}%
\pgfpathlineto{\pgfqpoint{5.938831in}{0.513275in}}%
\pgfpathlineto{\pgfqpoint{6.187913in}{0.465227in}}%
\pgfusepath{stroke}%
\end{pgfscope}%
\begin{pgfscope}%
\pgfpathrectangle{\pgfqpoint{3.834091in}{0.391000in}}{\pgfqpoint{2.465909in}{1.633000in}}%
\pgfusepath{clip}%
\pgfsetbuttcap%
\pgfsetroundjoin%
\pgfsetlinewidth{2.007500pt}%
\definecolor{currentstroke}{rgb}{0.000000,0.000000,0.000000}%
\pgfsetstrokecolor{currentstroke}%
\pgfsetdash{{2.000000pt}{6.000000pt}}{0.000000pt}%
\pgfpathmoveto{\pgfqpoint{3.946178in}{1.758670in}}%
\pgfpathlineto{\pgfqpoint{4.195259in}{1.287131in}}%
\pgfpathlineto{\pgfqpoint{4.444341in}{0.929667in}}%
\pgfpathlineto{\pgfqpoint{4.693423in}{0.751309in}}%
\pgfpathlineto{\pgfqpoint{4.942505in}{0.742005in}}%
\pgfpathlineto{\pgfqpoint{5.191586in}{0.663575in}}%
\pgfpathlineto{\pgfqpoint{5.440668in}{0.529677in}}%
\pgfpathlineto{\pgfqpoint{5.689750in}{0.548809in}}%
\pgfpathlineto{\pgfqpoint{5.938831in}{0.513275in}}%
\pgfpathlineto{\pgfqpoint{6.187913in}{0.465227in}}%
\pgfusepath{stroke}%
\end{pgfscope}%
\begin{pgfscope}%
\pgfsetrectcap%
\pgfsetmiterjoin%
\pgfsetlinewidth{0.803000pt}%
\definecolor{currentstroke}{rgb}{0.000000,0.000000,0.000000}%
\pgfsetstrokecolor{currentstroke}%
\pgfsetdash{}{0pt}%
\pgfpathmoveto{\pgfqpoint{3.834091in}{0.391000in}}%
\pgfpathlineto{\pgfqpoint{3.834091in}{2.024000in}}%
\pgfusepath{stroke}%
\end{pgfscope}%
\begin{pgfscope}%
\pgfsetrectcap%
\pgfsetmiterjoin%
\pgfsetlinewidth{0.803000pt}%
\definecolor{currentstroke}{rgb}{0.000000,0.000000,0.000000}%
\pgfsetstrokecolor{currentstroke}%
\pgfsetdash{}{0pt}%
\pgfpathmoveto{\pgfqpoint{6.300000in}{0.391000in}}%
\pgfpathlineto{\pgfqpoint{6.300000in}{2.024000in}}%
\pgfusepath{stroke}%
\end{pgfscope}%
\begin{pgfscope}%
\pgfsetrectcap%
\pgfsetmiterjoin%
\pgfsetlinewidth{0.803000pt}%
\definecolor{currentstroke}{rgb}{0.000000,0.000000,0.000000}%
\pgfsetstrokecolor{currentstroke}%
\pgfsetdash{}{0pt}%
\pgfpathmoveto{\pgfqpoint{3.834091in}{0.391000in}}%
\pgfpathlineto{\pgfqpoint{6.300000in}{0.391000in}}%
\pgfusepath{stroke}%
\end{pgfscope}%
\begin{pgfscope}%
\pgfsetrectcap%
\pgfsetmiterjoin%
\pgfsetlinewidth{0.803000pt}%
\definecolor{currentstroke}{rgb}{0.000000,0.000000,0.000000}%
\pgfsetstrokecolor{currentstroke}%
\pgfsetdash{}{0pt}%
\pgfpathmoveto{\pgfqpoint{3.834091in}{2.024000in}}%
\pgfpathlineto{\pgfqpoint{6.300000in}{2.024000in}}%
\pgfusepath{stroke}%
\end{pgfscope}%
\begin{pgfscope}%
\definecolor{textcolor}{rgb}{0.000000,0.000000,0.000000}%
\pgfsetstrokecolor{textcolor}%
\pgfsetfillcolor{textcolor}%
\pgftext[x=5.067045in,y=2.107333in,,base]{\color{textcolor}\rmfamily\fontsize{10.800000}{12.960000}\selectfont 50 classes}%
\end{pgfscope}%
\end{pgfpicture}%
\makeatother%
\endgroup%

%% file: aistats.bbl
\begin{thebibliography}{50}
\providecommand{\natexlab}[1]{#1}
\providecommand{\url}[1]{\texttt{#1}}
\expandafter\ifx\csname urlstyle\endcsname\relax
  \providecommand{\doi}[1]{doi: #1}\else
  \providecommand{\doi}{doi: \begingroup \urlstyle{rm}\Url}\fi

\bibitem[Amari(2016)]{amari_2016}
{Shun-ichi} Amari.
\newblock
  \emph{\href{https://www.springer.com/us/book/9784431559771}{Information
  Geometry and Its Applications}}.
\newblock Springer, 2016.

\bibitem[Amid and Warmuth(2017)]{amid_2017}
Ehsan Amid and Manfred~K Warmuth.
\newblock \href{https://arxiv.org/abs/1705.07210}{Two-temperature logistic
  regression based on the Tsallis divergence}.
\newblock \emph{arXiv preprint arXiv:1705.07210}, 2017.

\bibitem[Ball et~al.(1994)Ball, Carlen, and Lieb]{ball_1994}
Keith Ball, Eric~A Carlen, and Elliott~H Lieb.
\newblock \href{https://link.springer.com/article/10.1007/BF01231769} {Sharp
  uniform convexity and smoothness inequalities for trace norms}.
\newblock \emph{Inventiones Mathematicae}, 115\penalty0 (1):\penalty0 463--482,
  1994.

\bibitem[Banerjee et~al.(2005)Banerjee, Merugu, Dhillon, and
  Ghosh]{bregman_clustering}
Arindam Banerjee, Srujana Merugu, Inderjit~S Dhillon, and Joydeep Ghosh.
\newblock Clustering with bregman divergences.
\newblock \emph{Journal of machine learning research}, 6:\penalty0 1705--1749,
  2005.

\bibitem[Bauschke and Combettes(2017)]{Bauschke2017}
Heinz~H Bauschke and Patrick~L Combettes.
\newblock \emph{\href{https://dx.doi.org/10.1007/978-3-319-48311-5}{Convex
  Analysis and Monotone Operator Theory in Hilbert Spaces}}.
\newblock Springer, 2nd edition, 2017.

\bibitem[Beck and Teboulle(2012)]{beck_2012}
Amir Beck and Marc Teboulle.
\newblock \href{https://epubs.siam.org/doi/abs/10.1137/100818327}{Smoothing and
  first order methods: A unified framework}.
\newblock \emph{SIAM Journal on Optimization}, 22\penalty0 (2):\penalty0
  557--580, 2012.

\bibitem[Berger and Parker(1970)]{Berger1970}
Wolfgang~H Berger and Frances~L Parker.
\newblock \href{https://www.ncbi.nlm.nih.gov/pubmed/17731043}{Diversity of
  planktonic foraminifera in deep-sea sediments}.
\newblock \emph{Science}, 168\penalty0 (3937):\penalty0 1345--1347, 1970.

\bibitem[Bertsekas(1999)]{bertsekas_book}
Dimitri~P Bertsekas.
\newblock \emph{\href{http://www.athenasc.com/nonlinbook.html}{Nonlinear
  Programming}}.
\newblock Athena Scientific Belmont, 1999.

\bibitem[Blondel et~al.(2019)Blondel, Martins, and Niculae]{journal_version}
Mathieu Blondel, Andr{\'e}~FT Martins, and Vlad Niculae.
\newblock Learning with fenchel-young losses.
\newblock \emph{arXiv preprint arXiv:1901.02324}, 2019.

\bibitem[Borwein and Lewis(2010)]{borwein_2010}
Jonathan Borwein and Adrian~S Lewis.
\newblock \emph{\href{https://dx.doi.org/10.1007/978-0-387-31256-9}{Convex
  Analysis and Nonlinear Optimization: Theory and Examples}}.
\newblock Springer Science \& Business Media, 2010.

\bibitem[Boyd and Vandenberghe(2004)]{boyd_book}
Stephen Boyd and Lieven Vandenberghe.
\newblock \emph{\href{https://web.stanford.edu/~boyd/cvxbook/}{Convex
  Optimization}}.
\newblock Cambridge University Press, 2004.

\bibitem[Bregman(1967)]{bregman_1967}
Lev~M Bregman.
\newblock
  \href{https://www.sciencedirect.com/science/article/pii/0041555367900407}{The
  relaxation method of finding the common point of convex sets and its
  application to the solution of problems in convex programming}.
\newblock \emph{USSR Computational Mathematics and Mathematical Physics},
  7\penalty0 (3):\penalty0 200--217, 1967.

\bibitem[Brier(1950)]{brier_1950}
Glenn~W Brier.
\newblock
  \href{https://journals.ametsoc.org/doi/abs/10.1175/1520-0493%281950%29078%3C0001%3AVOFEIT%3E2.0.CO%3B2}{Verification
  of forecasts expressed in terms of probability}.
\newblock \emph{Monthly Weather Review}, 78\penalty0 (1):\penalty0 1--3, 1950.

\bibitem[Brucker(1984)]{Brucker1984}
Peter Brucker.
\newblock
  \href{https://www.sciencedirect.com/science/article/pii/0167637784900105}{An
  $O(n)$ algorithm for quadratic knapsack problems}.
\newblock \emph{Operations Research Letters}, 3\penalty0 (3):\penalty0
  163--166, 1984.

\bibitem[Buja et~al.(2005)Buja, Stuetzle, and Shen]{buja_2005}
Andreas Buja, Werner Stuetzle, and Yi~Shen.
\newblock
  \href{http://www-stat.wharton.upenn.edu/~buja/PAPERS/paper-proper-scoring.pdf}{Loss
  functions for binary class probability estimation and classification:
  Structure and applications}.
\newblock Technical report, University of Pennsylvania, 2005.

\bibitem[Condat(2016)]{Condat2016}
Laurent Condat.
\newblock \href{https://hal.archives-ouvertes.fr/hal-01056171}{Fast projection
  onto the simplex and the $\ell_1$ ball}.
\newblock \emph{Mathematical Programming}, 158\penalty0 (1-2):\penalty0
  575--585, 2016.

\bibitem[Crammer and Singer(2001)]{multiclass_svm}
Koby Crammer and Yoram Singer.
\newblock \href{http://www.jmlr.org/papers/v2/crammer01a.html}{On the
  algorithmic implementation of multiclass kernel-based vector machines}.
\newblock \emph{Journal of Machine Learning Research}, 2:\penalty0 265--292,
  2001.

\bibitem[Danskin(1966)]{danskin_theorem}
John~M Danskin.
\newblock \href{https://epubs.siam.org/doi/abs/10.1137/0114053}{The theory of
  max-min, with applications}.
\newblock \emph{SIAM Journal on Applied Mathematics}, 14\penalty0 (4):\penalty0
  641--664, 1966.

\bibitem[DeGroot(1962)]{degroot_1962}
Morris~H DeGroot.
\newblock \href{https://projecteuclid.org/euclid.aoms/1177704567}{Uncertainty,
  information, and sequential experiments}.
\newblock \emph{The Annals of Mathematical Statistics}, 33\penalty0
  (2):\penalty0 404--419, 1962.

\bibitem[Duchi et~al.(2008)Duchi, Shalev-Shwartz, Singer, and Chandra]{duchi}
John~C Duchi, Shai Shalev-Shwartz, Yoram Singer, and Tushar Chandra.
\newblock \href{https://stanford.edu/~jduchi/projects/DuchiShSiCh08.pdf}
  {Efficient projections onto the $\ell_1$-ball for learning in high
  dimensions}.
\newblock In \emph{Proceedings of ICML}, 2008.

\bibitem[Duchi et~al.(2018)Duchi, Khosravi, and Ruan]{duchi_2016}
John~C Duchi, Khashayar Khosravi, and Feng Ruan.
\newblock \href{http://arxiv.org/abs/1603.00126}{Multiclass classification,
  information, divergence, and surrogate risk}.
\newblock \emph{The Annals of Statistics}, 46\penalty0 (6B):\penalty0
  3246--3275, 2018.

\bibitem[Gell-Mann and Tsallis(2004)]{GellMannTsallis2004}
Murray Gell-Mann and Constantino Tsallis.
\newblock
  \emph{\href{https://global.oup.com/academic/product/nonextensive-entropy-9780195159776}{Nonextensive
  Entropy: Interdisciplinary Applications}}.
\newblock Oxford University Press, 2004.

\bibitem[Gini(1912)]{gini_index}
Corrado Gini.
\newblock
  \href{https://books.google.com/books/about/Variabilit%C3%A0_e_mutabilit%C3%A0.html?id=fqjaBPMxB9kC}{Variabilit{\`a}
  e mutabilit{\`a}}.
\newblock \emph{Reprinted in Memorie di metodologica statistica (Ed. Pizetti E,
  Salvemini, T). Rome: Libreria Eredi Virgilio Veschi}, 1912.

\bibitem[Gneiting and Raftery(2007)]{gneiting_2007}
Tilmann Gneiting and Adrian~E Raftery.
\newblock
  \href{https://amstat.tandfonline.com/doi/abs/10.1198/016214506000001437}{Strictly
  proper scoring rules, prediction, and estimation}.
\newblock \emph{Journal of the American Statistical Association}, 102\penalty0
  (477):\penalty0 359--378, 2007.

\bibitem[Gr{\"u}nwald and Dawid(2004)]{grunwald_2004}
Peter~D Gr{\"u}nwald and A~Philip Dawid.
\newblock \href{https://arxiv.org/abs/math/0410076}{Game theory, maximum
  entropy, minimum discrepancy and robust Bayesian decision theory}.
\newblock \emph{The Annals of Statistics}, 32\penalty0 (4):\penalty0
  1367--1433, 2004.

\bibitem[Guermeur(2007)]{guermeur_2007}
Yann Guermeur.
\newblock \href{http://www.jmlr.org/papers/v8/guermeur07a.html}{VC theory of
  large margin multi-category classifiers}.
\newblock \emph{Journal of Machine Learning Research}, 8:\penalty0 2551--2594,
  2007.

\bibitem[Krichene et~al.(2015)Krichene, Krichene, and Bayen]{bregmanproj}
Walid Krichene, Syrine Krichene, and Alexandre Bayen.
\newblock \href{https://ieeexplore.ieee.org/document/7402714/}{Efficient
  Bregman projections onto the simplex}.
\newblock In \emph{Proceedings of CDC}, 2015.

\bibitem[Liu and Nocedal(1989)]{lbfgs}
Dong~C Liu and Jorge Nocedal.
\newblock \href{https://doi.org/10.1007/BF01589116}{On the limited memory BFGS
  method for large scale optimization}.
\newblock \emph{Mathematical Programming}, 45\penalty0 (1):\penalty0 503--528,
  1989.

\bibitem[Mangasarian(1965)]{Mangasarian1965}
Olvi~L Mangasarian.
\newblock \href{https://epubs.siam.org/doi/abs/10.1137/0303020}{Pseudo-convex
  functions}.
\newblock \emph{Journal of the Society for Industrial and Applied Mathematics,
  Series A: Control}, 3\penalty0 (2):\penalty0 281--290, 1965.

\bibitem[Martins and Astudillo(2016)]{sparsemax}
Andr{\'e}~FT Martins and Ram{\'o}n~Fernandez Astudillo.
\newblock \href{https://arxiv.org/abs/1602.02068} {From softmax to sparsemax: A
  sparse model of attention and multi-label classification}.
\newblock In \emph{Proceedings of ICML}, 2016.

\bibitem[Martins et~al.(2009)Martins, Figueiredo, Aguiar, Smith, and
  Xing]{Martins2009JMLR}
Andr\'{e}~FT Martins, M\'{a}rio~AT Figueiredo, Pedro~MQ Aguiar, Noah~A Smith,
  and Eric~P Xing.
\newblock \href{http://www.jmlr.org/papers/v10/martins09a.html}{Nonextensive
  information theoretic kernels on measures}.
\newblock \emph{Journal of Machine Learning Research}, 10:\penalty0 935--975,
  2009.

\bibitem[Masnadi-Shirazi(2011)]{masnadi_2011}
Hamed Masnadi-Shirazi.
\newblock \emph{\href{https://escholarship.org/uc/item/1cv1947c}{The design of
  bayes consistent loss functions for classification}}.
\newblock PhD thesis, UC San Diego, 2011.

\bibitem[{Mensch} and {Blondel}(2018)]{differentiable_dp}
Arthur {Mensch} and Mathieu {Blondel}.
\newblock \href{https://arxiv.org/abs/1802.03676}{Differentiable dynamic
  programming for structured prediction and attention}.
\newblock In \emph{Proceedings of ICML}, 2018.

\bibitem[Nelder and Baker(1972)]{glm}
John~Ashworth Nelder and R~Jacob Baker.
\newblock
  \emph{\href{https://onlinelibrary.wiley.com/doi/full/10.1002/0471667196.ess0866.pub2}{Generalized
  Linear Models}}.
\newblock Wiley Online Library, 1972.

\bibitem[Nesterov(2005)]{nesterov_smooth}
Yurii Nesterov.
\newblock \href{https://link.springer.com/article/10.1007/s10107-004-0552-5}
  {Smooth minimization of non-smooth functions}.
\newblock \emph{Mathematical Programming}, 103\penalty0 (1):\penalty0 127--152,
  2005.

\bibitem[Niculae and Blondel(2017)]{Niculae2017}
Vlad Niculae and Mathieu Blondel.
\newblock \href{https://arxiv.org/abs/1705.07704}{A regularized framework for
  sparse and structured neural attention}.
\newblock In \emph{Proceedings of NIPS}, 2017.

\bibitem[Niculae et~al.(2018)Niculae, Martins, Blondel, and Cardie]{sparsemap}
Vlad Niculae, Andr{\'e}~FT Martins, Mathieu Blondel, and Claire Cardie.
\newblock \href{https://arxiv.org/abs/1802.04223}{SparseMAP: Differentiable
  sparse structured inference}.
\newblock In \emph{Proceedings of ICML}, 2018.

\bibitem[Nock and Nielsen(2009)]{nock_2009}
Richard Nock and Frank Nielsen.
\newblock \href{https://doi.org/10.1109/TPAMI.2008.225}{Bregman divergences and
  surrogates for learning}.
\newblock \emph{IEEE Transactions on Pattern Analysis and Machine
  Intelligence}, 31\penalty0 (11):\penalty0 2048--2059, 2009.

\bibitem[Reid and Williamson(2010)]{reid_composite_binary}
Mark~D Reid and Robert~C Williamson.
\newblock \href{http://www.jmlr.org/papers/v11/reid10a.html}{Composite binary
  losses}.
\newblock \emph{Journal of Machine Learning Research}, 11:\penalty0 2387--2422,
  2010.

\bibitem[R{\'e}nyi(1961)]{Renyi1960}
Alfr{\'e}d R{\'e}nyi.
\newblock \href{https://projecteuclid.org/euclid.bsmsp/1200512181}{On measures
  of entropy and information}.
\newblock In \emph{Proceedings of the 4th Berkeley Symposium on Mathematical
  Statistics, and Probability}, volume~1. University of California Press, 1961.

\bibitem[Rockafellar(1970)]{Rockafellar1970}
R~Tyrrell Rockafellar.
\newblock \emph{\href{https://press.princeton.edu/titles/1815.html}{Convex
  Analysis}}.
\newblock Princeton University Press, 1970.

\bibitem[Rosenblatt(1958)]{perceptron}
Frank Rosenblatt.
\newblock \href{http://psycnet.apa.org/record/1959-09865-001}{The perceptron: A
  probabilistic model for information storage and organization in the brain.}
\newblock \emph{Psychological Review}, 65\penalty0 (6):\penalty0 386, 1958.

\bibitem[Savage(1971)]{savage_1971}
Leonard~J Savage.
\newblock \href{https://www.jstor.org/stable/2284229}{Elicitation of personal
  probabilities and expectations}.
\newblock \emph{Journal of the American Statistical Association}, 66\penalty0
  (336):\penalty0 783--801, 1971.

\bibitem[Sch{\"o}lkopf and Smola(2002)]{Schoelkopf2002}
Bernhard Sch{\"o}lkopf and Alexander~J Smola.
\newblock \emph{\href{https://mitpress.mit.edu/books/learning-kernels}{Learning
  with Kernels}}.
\newblock The MIT Press, Cambridge, MA, 2002.

\bibitem[Shalev-Shwartz and Zhang(2016)]{accelerated_sdca}
Shai Shalev-Shwartz and Tong Zhang.
\newblock \href{https://arxiv.org/abs/1309.2375} {Accelerated proximal
  stochastic dual coordinate ascent for regularized loss minimization}.
\newblock \emph{Mathematical Programming}, 155\penalty0 (1):\penalty0 105--145,
  2016.

\bibitem[Suyari(2004)]{Suyari2004}
Hiroki Suyari.
\newblock \href{https://arxiv.org/abs/math-ph/0205004}{Generalization of
  Shannon-Khinchin axioms to nonextensive systems and the uniqueness theorem
  for the nonextensive entropy}.
\newblock \emph{IEEE Transactions on Information Theory}, 50\penalty0
  (8):\penalty0 1783--1787, 2004.

\bibitem[Tsallis(1988)]{Tsallis1988}
Constantino Tsallis.
\newblock \href{https://link.springer.com/article/10.1007/BF01016429}{Possible
  generalization of Boltzmann-Gibbs statistics}.
\newblock \emph{Journal of Statistical Physics}, 52:\penalty0 479--487, 1988.

\bibitem[Vapnik(1998)]{Vapnik1998}
Vladimir~N Vapnik.
\newblock
  \emph{\href{https://www.wiley.com/en-us/Statistical+Learning+Theory-p-9780471030034}{Statistical
  Learning Theory}}.
\newblock Wiley, 1998.

\bibitem[Wainwright and Jordan(2008)]{wainwright_2008}
Martin~J Wainwright and Michael~I Jordan.
\newblock
  \href{https://people.eecs.berkeley.edu/~wainwrig/Papers/WaiJor08_FTML.pdf}{Graphical
  models, exponential families, and variational inference.}
\newblock \emph{Foundations and Trends{\textregistered} in Machine Learning},
  1\penalty0 (1--2):\penalty0 1--305, 2008.

\bibitem[Williamson et~al.(2016)Williamson, Vernet, and Reid]{vernet_2016}
Robert~C Williamson, Elodie Vernet, and Mark~D Reid.
\newblock \href{http://jmlr.org/papers/v17/14-294.html}{Composite multiclass
  losses}.
\newblock \emph{Journal of Machine Learning Research}, 17\penalty0
  (223):\penalty0 1--52, 2016.

\end{thebibliography}
